\documentclass[journal]{IEEEtran}
\PassOptionsToPackage{dvipsnames, table}{xcolor}
\usepackage{standalone}

\usepackage[T1]{fontenc}
\usepackage[utf8]{inputenc}%
\usepackage{microtype}
\usepackage{times}
\usepackage{cite}

\usepackage{amsmath}
\usepackage{amssymb}
\usepackage{mathtools}

\usepackage{scalerel}
\DeclareMathOperator*{\concat}{\scalerel*{\|}{\sum}}

\DeclarePairedDelimiter{\norm}{\lVert}{\rVert}
\DeclarePairedDelimiter{\abs}{\lvert}{\rvert}

\DeclarePairedDelimiterX{\infdivx}[2]{(}{)}{%
  #1\;\delimsize\|\;#2%
}
\newcommand{\kl}{\operatorname{KL}\infdivx}

\let\given\givenbase

\DeclareMathOperator*{\argmin}{arg\,min}

\let\oldtimes\times
\def\times{{\mkern1mu\oldtimes\mkern1mu}}

\usepackage{etoolbox}
\robustify\bfseries

\usepackage{xcolor}

\usepackage{graphicx}
\graphicspath{{./images/}}

\usepackage[labelformat=simple]{subcaption}
\usepackage{tikz}
\usepackage{pgfplots}
\usepackage{pgfplotstable}

\usetikzlibrary{%
  calc,
  shapes,
  arrows,
  intersections,
  positioning,
  scopes,
  fit,
  plotmarks,
  patterns,
  matrix,
}
\usepgfplotslibrary{
  external,
  colorbrewer,
  patchplots,
  groupplots,
}

\pgfplotsset{compat=newest,%
  width=\linewidth,height=0.85\linewidth,%
  every axis plot post/.append style={thick},%
  ymajorgrids,
  grid style={dashed}, %
  minimal plot grid/.style={
    y axis line style={opacity=0},
    axis x line*=bottom,
    x axis line style={black},
  },
  minimal plot grid,
  /pgfplots/legend pos/south center/.style={/pgfplots/legend style={at={(0.5,0.03)},anchor=south}},
  /pgfplots/legend pos/north center/.style={/pgfplots/legend style={at={(0.5,0.97)},anchor=north}},
  y tick label style={
    /pgf/number format/.cd,
    fixed,
    fixed zerofill,
    precision=1,
    /tikz/.cd,
    font=\footnotesize,
  },
  x tick label style={
    /pgf/number format/.cd,
    fixed,
    fixed zerofill,
    precision=1,
    /tikz/.cd,
    font=\footnotesize,
  },
  /pgfplots/legend pos/south center/.style={/pgfplots/legend style={at={(0.5,0.03)},anchor=south}},
  /pgfplots/legend pos/north center/.style={/pgfplots/legend style={at={(0.5,0.97)},anchor=north}},
  /pgfplots/legend pos/outer north center/.style={/pgfplots/legend style={at={(0.5,1.02)},anchor=south}},
  discard if not/.style 2 args={
    x filter/.code={
      \edef\tempa{\thisrow{#1}}
      \edef\tempb{#2}
      \ifx\tempa\tempb
      \else
      \def\pgfmathresult{inf}
      \fi
    }
  },
  alpha beta/.style={
    minor xtick={1.25,1.75,...,2.25},
    xtick={1,1.5,...,2.5},
    minor ytick={1,1.5,2.5,...,4.5},
    ytick={2,...,5},
    minor z tick num = 1,
    grid=none,
    every axis plot post/.append style={thin},
    mesh/rows=9,%
    mesh/ordering=y varies,%
    colormap/viridis,
    view={0}{90},
  },
  alpha beta 3d/.style={
    width=5cm,
    height=4.3125cm,
    font=\footnotesize,
    anchor=center,
    z axis line style={opacity=0},
    xmajorgrids,
    axis y line*=left,
    y axis line style={opacity=1, black},
    minor xtick={1.25,1.75,...,2.25},
    xtick={1,1.5,...,2.5},
    minor ytick={1,1.5,2.5,...,4.5},
    ytick={2,...,5},
    zmin=0.1, 
    zmax=1,
    minor z tick num = 1,
    every axis plot post/.append style={thin},
    xlabel={$\alpha$},
    ylabel={$\beta$},
    zlabel={#1},
    mesh/rows=9,%
    mesh/ordering=y varies,%
    colormap/viridis,
    view={-45}{30},
  },
}

\usepackage{multirow}
\usepackage{booktabs}
\usepackage{siunitx}
\usepackage{xspace}
\makeatletter
\DeclareRobustCommand\onedot{\futurelet\@let@token\@onedot}
\def\@onedot{\ifx\@let@token.\else.\null\fi\xspace}

\def\eg{e.g\onedot} \def\Eg{E.g\onedot}
\def\ie{i.e\onedot} 
\def\cf{cf\onedot} 
\def\etc{etc\onedot} 
 
\def\etal{et al\onedot}

\makeatother

\definecolor{orcidlogocol}{HTML}{A6CE39}
\usetikzlibrary{svg.path}
\tikzset{
  orcidlogo/.pic={
    \fill[orcidlogocol] svg{M256,128c0,70.7-57.3,128-128,128C57.3,256,0,198.7,0,128C0,57.3,57.3,0,128,0C198.7,0,256,57.3,256,128z};
    \fill[white] svg{M86.3,186.2H70.9V79.1h15.4v48.4V186.2z}
    svg{M108.9,79.1h41.6c39.6,0,57,28.3,57,53.6c0,27.5-21.5,53.6-56.8,53.6h-41.8V79.1z M124.3,172.4h24.5c34.9,0,42.9-26.5,42.9-39.7c0-21.5-13.7-39.7-43.7-39.7h-23.7V172.4z}
    svg{M88.7,56.8c0,5.5-4.5,10.1-10.1,10.1c-5.6,0-10.1-4.6-10.1-10.1c0-5.6,4.5-10.1,10.1-10.1C84.2,46.7,88.7,51.3,88.7,56.8z};
  }
}
\newcommand\orcidlogo[1]{\href{https://orcid.org/#1}{\mbox{\scalerel*{
        \begin{tikzpicture}[yscale=-1,transform shape]
        \pic{orcidlogo};
        \end{tikzpicture}
      }{1}}}}
\newcommand\orcidID[1]{$^{\orcidlogo{#1}}$}

\usepackage[pagebackref=false,breaklinks=true,letterpaper=true,colorlinks,bookmarks=false]{hyperref}

\hypersetup{
  linkcolor  = red!75!black,
  citecolor  = green!75!black,
  urlcolor   = magenta!75!black,
  colorlinks = true,
}

\begin{document}

\title{Anomaly Detection based on Zero-Shot Outlier Synthesis and Hierarchical Feature Distillation}

\author{%
  Ad\'in Ram\'irez~Rivera,\orcidID{0000-0002-4321-9075}~\IEEEmembership{Member,~IEEE,} %
  Adil Khan,~\IEEEmembership{Member,~IEEE,}  %
  Imad E.\ I.\ Bekkouch, %
  Taimoor S.\ Sheikh %
  \thanks{A.\ Ram\'irez~Rivera is with the Institute of Computing, University of Campinas, Brazil, e-mail: \texttt{adin@ic.unicamp.br}.  A. Khan, I.\ Bekkouch, and T.\ Sheikh are with the Institute of Data Science \& Artificial Intelligence, Innopolis University, Russia, e-mails: \texttt{\{a.khan, t.sheikh\}@innopolis.ru}, \texttt{i.bekkouch@innopolis.university}.}
  \thanks{This work was supported in part by the Brazilian National Council for Scientific and Technological Development (CNPq) under grant No.~307425/2017-7 and in part by the S\~ao Paulo Research Foundation (FAPESP) under grant No.~2019/07257-3.}%
  \thanks{Pre-print to appear in IEEE Trans.\ on Neural Networks and Learning Systems \href{https://doi.org/10.1109/TNNLS.2020.3027667}{10.1109/TNNLS.2020.3027667}}}%
\maketitle

\begin{abstract}
Anomaly detection suffers from unbalanced data since anomalies are quite rare.  Synthetically generated anomalies are a solution to such ill or not fully defined data.  However, synthesis requires an expressive representation to guarantee the quality of the generated data.  In this paper, we propose a two-level hierarchical latent space representation that distills inliers' feature-descriptors (through autoencoders) into more robust representations based on a variational family of distributions (through a variational autoencoder) for zero-shot anomaly generation.  From the learned latent distributions, we select those that lie on the outskirts of the training data as synthetic-outlier generators.  And, we synthesize from them, \ie, generate negative samples without seen them before, to train binary classifiers.  We found that the use of the proposed hierarchical structure for feature distillation and fusion creates robust and general representations that allow us to synthesize pseudo outlier samples.  And in turn, train robust binary classifiers for true outlier detection (without the need for actual outliers during training).  We demonstrate the performance of our proposal on several benchmarks for anomaly detection.
\end{abstract}

\section{Introduction}
\label{sec:intro}

Outlier, novelty, or anomaly detection is the process of identifying new data samples as part of the learned class (inliers) or not (outliers).  This problem is relevant since most problems do not have a fully characterized set of interest data.  During testing, the classifier must adapt to unseen data.  There is a vast literature on these topics~\cite{Ahmed2016, Chandola2009, Hodge2004, Pimentel2014} since they have broad applications, with a particular interest in vision~\cite{Buades2005, Cong2011, Li2014, Sabokrou2017, Sabokrou2018, Xia2015, You2017}.

We can classify these methods depending on the type of modeling used, such as probabilistic-based~\cite{Eskin2000, Kim2012, Lerman2015, Markou2003, Yamanishi2004}, which model the distribution of inliers and the outliers are those points with low probability; distance-based~\cite{Bodesheim2013, Breunig2000, Liu2017}, which identify outliers by their distance to inliers; self-representation~\cite{Cong2011, Rahmani2017, Sabokrou2016, You2017}, which detect outliers as non-sparse representations on the given inliers; and deep-learning-based~\cite{Hasan2016, Kimura2018, Pidhorskyi2018, Ravanbakhsh2017, Sabokrou2015, Sabokrou2018, Schlegl2017, Xia2015, Wang2018}, which use reconstruction-based networks to detect outliers according to their reconstruction error.  

Some previous methods rely on true outliers and could suffer from unbalanced data problems since anomalies are quite rare.  A solution to these problems is the creation of synthetic anomalies from the anomaly class.  In contrast to using true outliers for anomaly creation, we propose a zero-shot synthesis, by generating outliers from the boundaries of the inlier data, to train binary classifiers.  However, using the data on the feature space is ineffective due to uncertainties on the original data and errors introduced by the feature extraction process.  Instead, we propose to model the data using probability distributions through a hierarchical encoding process that models the uncertainty too.  Then, we use the outskirt distributions as sources for our outlier synthesizing process.  This process proves to be effective in learning a robust boundary that correctly classifies (never seen before) outliers. 

\begin{figure}[t]
  \centering
  \resizebox{.9\linewidth}{!}{\def\go{0/0/Dark2-A,1/1.05/Dark2-B,.2/.95/Dark2-C,.95/.1/Dark2-D,0.5/0.5/Dark2-E}
\makeatletter
\begin{tikzpicture}[
rounded corners,
>=latex,
font=\footnotesize,
declare function={
  gaussian(\x,\mu,\sigma) = 1/(\sigma*sqrt(2*pi))*exp(-((\x-\mu)^2)/(2*\sigma^2));
  gamma(\z)= (2.506628274631*sqrt(1/\z) + 0.20888568*(1/\z)^(1.5) + 0.00870357*(1/\z)^(2.5) - (174.2106599*(1/\z)^(3.5))/25920 - (715.6423511*(1/\z)^(4.5))/1244160)*exp((-ln(1/\z)-1)*\z);
  beta(\x,\alpha,\beta) = gamma(\alpha+\beta)*\x^(\alpha-1)*(1-\x)^(\beta-1) / (gamma(\alpha)*gamma(\beta));
  gammapdf(\x,\k,\theta) = \x^(\k-1)*exp(-\x/\theta) / (\theta^\k*gamma(\k));
},
/densities/.cd,
function/.store in=\@density,
domain/.store in=\@density@domain,
domain/.default={0:1},
width/.store in=\@density@width,
width/.default=1.5cm,
height/.store in=\@density@height,
height/.default=1.5cm,
density style/.store in=\@density@style,
axis style/.store in=\@density@axis@style,
density hidden/.style={width, height, domain, density style={}, axis style={}},
density/.style={density hidden/.append style={#1}},
/tikz/.cd,
pics/density/.style={code={%
    \tikzset{/densities/.cd, density hidden, #1}%
    \node[inner sep=0pt, outer sep=0pt] (\tikz@fig@name) {%
      \resizebox{\@density@width}{!}{%
        \begin{tikzpicture}[]%
        \begin{axis}[
        samples=100,
        y axis line style={draw=none},
        tick style={draw=none},
        axis x line*=bottom,
        ticks=none,
        grid=none,
        enlargelimits=false,
        \@density@axis@style
        ]
        \edef\tmp{\noexpand\addplot [smooth, domain=\@density@domain, \@density@style] {\@density};}
        \tmp
        \end{axis}
        \end{tikzpicture}%
      }%
    };%
}},
pics/gaussian/.style args={mu #1, sigma #2 [#3]}{ density={function={gaussian(x, #1, #2)},#3} },
]
\makeatother

\foreach \x/\y/\col [count=\i] in \go{
\node[mark size=3pt,color=\col] (p-\i) at (\x,\y) {\pgfuseplotmark{pentagon*}};
\node[mark size=3pt,color=\col, xshift=-2cm] (t-\i) at (\x,\y) {\pgfuseplotmark{triangle*}};
\node[mark size=3pt,color=\col, xshift=-4cm] (s-\i) at (\x,\y) {\pgfuseplotmark{square*}};
}
\coordinate (g1) at (p-5);
\coordinate (g2) at (t-5);
\coordinate (g3) at (s-5);
\draw[gray] (g1) circle(.9);
\draw[gray] (g2) circle(.9);
\draw[gray] (g3) circle(.9);
\draw[dashed, gray] (-4.5,1.5) rectangle (1.5,-.5);

\node[anchor=south west, above=12pt of s-2] (x) {$x$};
\node[anchor=east, below=1.2cm of g2] (g) {$g_i(x)$};

\draw[gray] (g.north) -| ($(g1)+(0,-1)$);
\draw[gray] (g.north) -- ($(g2)+(0,-1)$);
\draw[gray] (g.north) -| ($(g3)+(0,-1)$);

\draw[gray] (x.south) -- (s-2.north);

\begin{scope}[xshift=3cm, yshift=0cm]
\foreach \x/\y/\col in \go{
\draw pic at (\x,\y) {gaussian={mu .5, sigma .2 [density style={fill=\col, draw=\col, fill opacity=.5, very thick}, width=.5cm]}};
\node[mark size=1pt,color=\col, yshift=-6] at (\x,\y) {\pgfuseplotmark{*}};
}
\draw pic at (0,0) {gaussian={mu .5, sigma .2 [density style={draw=Dark2-A, very thick, pattern=north east lines, pattern color=Dark2-A}, width=.5cm, axis style={x axis line style={draw=none}}]}};
\draw pic at (1,1.05) {gaussian={mu .5, sigma .2 [density style={draw=Dark2-B, fill opacity=.5, very thick, pattern=north east lines, pattern color=Dark2-B}, width=.5cm, axis style={x axis line style={draw=none}}]}};

\def\out{-0.21,-0.1,-0.15}
\foreach \x in \out{
\node[mark size=1pt,color=Dark2-F] at (\x,-.205) {\pgfuseplotmark{diamond*}};
}
\node[mark size=1pt,color=Dark2-F] at (1.2,.84) {\pgfuseplotmark{diamond*}};
\node[mark size=1pt,color=Dark2-F] at (1.15,.84) {\pgfuseplotmark{diamond*}};

\draw[gray] (.5,.5) circle(1.1);

\node at (-.5,1.5) {$\mathcal{Q}$};
\end{scope}

\begin{scope}[xshift=5.5cm,yshift=0cm]
\foreach \col [count=\xi, evaluate=\xi as \x using \xi*.25] in {Dark2-A, Dark2-B, Dark2-C, Dark2-D, Dark2-E}{
  \node[mark size=3pt,color=\col] at (\x,.75) {\pgfuseplotmark{*}};
  \node[mark size=3pt,color=Dark2-F] at (\x,.25) {\pgfuseplotmark{diamond*}};
}
\draw[black!75] (0,0) rectangle (1.5,1);
\draw[dashed, Dark2-G] (0.05,.45) -- ++(1.45,0.05);

\node at (0.1,1.2) {$C$};
\end{scope}

\coordinate (z-left) at (2.3,0.5);
\draw[->] (1.5, .5) -- (z-left);
\draw[->] (4.7, .5) -- ++(.7,0);

\node[text width=1.cm, font=\scriptsize, text=black!75, align=center, anchor=north west] (df-lbl) at (1.5,-.25) {Distillation\\\& Fusion};
\node[text width=1.cm, font=\scriptsize, text=black!75, align=center, anchor=north west, right=1.5cm of df-lbl] {Synthetic\\Outlier\\Generation};

\node[] (od-lbl) at (4,1.8) {$\mathcal{O}$};
\draw[gray] (od-lbl.south) -- (4.,1.3);
\draw[gray] (od-lbl.south) -- (3,.25);

\node[right=.35cm of od-lbl] (o-lbl) {$y$};
\draw[gray] (o-lbl.south) -- (4.25,.85);
\draw[gray] (o-lbl.south) -- (5.65,.25);

\end{tikzpicture}  }
  \caption{Given a set of feature descriptors $\{g_i(x)\}_{i=0}^k$ (denoted by different shapes) for the set of inputs $x \in \mathcal{X}$ (denoted by different colors), we distill the features and fuse them into a distribution space $\mathcal{Q}$.  From them, we select a set of outskirt distributions, $\mathcal{O}$, as those farther from the other set of distributions (marked with diagonal lines).  They represent samples that are on the boundary of the latent space.  Then, we sample, from these outskirt distributions, our synthesized-outliers $y \in \mathcal{Y}$ (denoted as diamonds).  Finally, we train a classifier $C$ using the fused inputs (inliers) and the synthesized outliers as positive and negative samples, respectively.  This classifier is then used to detect true outliers, never seen before on this process.%
  }
  \label{fig:diagram}
\end{figure}
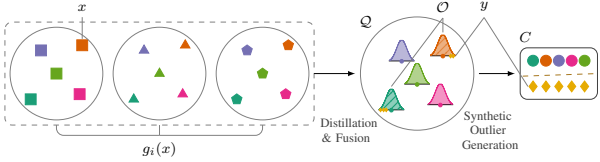

Our proposed method lies at the intersection of probabilistic-, distance-, and deep-learning-based models.  On one hand, we model our data (inliers) with distributions that represent the distilled information of a set of features.  In contrast to existing probabilistic models, we learn a variational (autoencoder) family that is conditioned on the inlier samples alone.  Thus, we generate a latent space of distributions that represent our inliers.  Then, similarly to distance-based methods, we select the distributions that are on the outskirts of our learned distribution-space to generate synthetic outliers by perturbing the drawings from these distributions.  Conversely to distance-based methods that work on the original feature space, our transformation creates a robust distribution-space to work with.  And, instead of directly using these outskirt distributions for classification, as probabilistic or distance methods do, we synthesize outlier samples to train a binary classifier.  We found that the mixture of these techniques yields better results than their traditional counterparts, with a simpler representation.  We show an illustration of the overall approach in Fig.~\ref{fig:diagram}.  Additionally, our method is trained without seeing any true outlier (thanks to our synthesis process).  In a way, our synthesized outliers are distorted versions of already ``strange'' samples. 

Our main contributions are: (1)~A simple, yet powerful, zero-shot outlier synthesis framework for anomaly detection, \ie, we do not use actual outlier data to train our classifier or feature distillation but rather synthetically generated outliers from inlier data.  (2)~A two-level hierarchical scheme for feature distillation and creation of inlier's distributions that allows synthesizing pseudo outliers which are sufficient to train a robust binary classifier.  (3)~We demonstrate that the use of probabilistic modeling paired with distance-based methods for boundary definition produces robust-enough synthetic negative samples to train binary classifiers with robust performance on novel data, in particular, from an outlier class.  (4)~We show that loosely coupled methods have advantages over tightly coupled ones (\ie, commonly end-to-end trained methods) on a set of outlier detection benchmarks.

\section{Related Works}
\label{sec:related}

\textbf{Data Representation.}
Existing methods rely on low level features, \eg, Histogram of Oriented Gradients~\cite{Dalal2005}, local patterns~\cite{Ojala1996, Huang2011, RamirezRivera2013, RamirezRivera2015a, RamirezRivera2015}; high level features, \eg, bag of words~\cite{Venhuizen2015}, trajectories~\cite{Morris2011}; or deep-learned features~\cite{Erfani2016, Kimura2018, Pidhorskyi2018, Sabokrou2015, Sabokrou2017, Sabokrou2018, Schlegl2017, Xu2015}.  The latter have gained traction in the past years since they tend to outperform low or high-level features designed for general purposes since the deep-features are tuned for each particular task.  However, due to this same property, they are not generalizable and not generally used outside of their own proposed architecture.  When a similar architecture is adapted into another problem, the previous weights need to be fine-tuned for the new tasks.  On the contrary, we propose a distillation method from feature descriptors (\eg, low or high-level features, or from deep-features from other tasks) that can generalize them without adapting their original architectures.  Our proposed model takes advantage of compressing the features through autoencoders (AEs).  
On the other hand, we also propose a zero-shot synthetic outlier generation scheme.  Some methods tried to synthesize outliers while training by adding noise to the input samples~\cite{Sabokrou2018}.  Nonetheless, by perturbing the original data, they can miss the original data variations.  On the contrary, we propose to synthesize data on a constructed feature-distribution space that already incorporated the variations from the inliers.

\textbf{Self-Representation.} 
By assuming that outliers cannot be reconstructed sparsely from a set of inlier points, self-representation methods~\cite{Cong2011, Sabokrou2016, You2017} achieve outlier detection.  Others exploit the correlations between the inliers that must lie on lower-dimensional spaces, and assume that incoherent samples must be outliers~\cite{Rahmani2017}. Another similar stream of research~\cite{Lerman2015, Xu2010} solves a problem with variations of the data points to find lower-dimensional spaces that define the inliers.  The idea of sparse representation can be extended as the ability to reconstruct the samples too~\cite{Cong2011, Xu2015}.  In this sense, they are related to deep learning generative methods.  

\textbf{Probabilistic-based Methods.}  
Conventional probabilistic methods~\cite{Eskin2000, Kim2012, Lerman2015, Markou2003, Yamanishi2004, Park2018, Atli2018} model the inliers' features' distributions, where the outliers are detected as samples with low probability.  Several approximations to the true distribution of the data have been proposed, for instance, Genetic programming~\cite{Cao2016} to estimate a kernel density function, minimum-volume-sets to estimate a particular level set of the unknown nominal multivariate density~\cite{Scott2006}, or constructing minimal graphs covering a $K$-point subset to estimate the critical region~\cite{Hero2007, Sricharan2011}. Maximum-entropy-discrimination based models were proposed with several variants, and a recent method mixes this framework with hinge loss style discriminant functions and latent variables to discriminate the outliers~\cite{Hou2018}. More recent methods~\cite{Pidhorskyi2018} mix the distribution's generation with manifolds to better model the inliers.  Similarly, our proposed method is a mixture of ideas, and cannot be classified in a single category.  On the contrary to existing probabilistic methods, our proposed method does not use the probabilities as a novelty score, but rather as a synthesizer of new data.  

\textbf{Distance-based Methods.} 
The core idea of distance-based methods is the assumption that inliers are close to each other, while outliers are far from them.  Commonly, $k$-nearest neighbors are used with density estimation~\cite{Breunig2000}, as well as more advanced kernel transforms~\cite{Bodesheim2013, Liu2017} where the distance between the projections is used for the novelty score. Conversely, we use the notion of distance to select outlier distributions that will be used later on, instead of assigning a novelty score from the neighbors.  The notion of distance between points can be posed as a classification problem.  One breakthrough was the kernel-based novelty detection scheme that relies on unsupervised support vector machine~\cite{Schoelkopf2000}.

\textbf{Deep Learning.}  
The state of the art on outlier detection is based on deep neural networks.  In particular, they continue the ideas of self-representation methods based on reconstruction.  The main architecture is based on autoencoders (AEs)~\cite{Hasan2016, Xu2015}.  Others~\cite{Pidhorskyi2018, Ravanbakhsh2017, Sabokrou2018, Xia2015, Wang2018, Lim2018, Abdulhammed2019} add a generative adversarial loss~\cite{Goodfellow2014} to enhance the reconstruction of the AEs.  The main idea is to train a discriminator and a generator (\ie, the AE) on a min-max fashion.  This optimization scheme yields better results for the generation of the original samples.  However, it is hard to train, and it is unstable due to the different learning rates of its components.  Similarly to existing approaches, we use the unsupervised nature of AEs to learn a compact space of the original features (akin to  self-representation).  However, we do not work on this space, and, yet again, we produce a higher-level feature space based on distributions through variational autoencoders (VAEs), which have shown excellent capabilities for unsupervised clustering~\cite{AriasFigueroa2017, AriasFigueroa2017a, Kingma2014, Baur2018, Xu2018a}.  And in contrast to common deep learning approaches, we detect outliers with vanilla classifiers.

\textbf{Novelty.}  We use AEs and VAEs similar to previous approaches~\cite{Kingma2014, Zhai2016}.  (1)~However, our proposal's novelty lies in the construction of a two-level latent space hierarchy that (i)~encodes the inliers into a low-dimensional latent space that supports a reliable estimate on inliers' distributions (through AEs); and (ii)~provides a dense representation of the points in a distribution-space where inlier distributions can be compacted to find a reliable boundary (through VAEs).  (2)~Unlike previous tightly coupled outlier detectors~\cite{Nguyen2018}, we separate the feature modeling, synthetic-outlier generation, and classification, which enables our learning spaces to be used on a diverse set of tasks.  (3)~Our zero-shot outlier generation from near-boundary inlier-distributions is different from previous generation methods~\cite{Landgrebe2006} that commonly assume a uniform maximum entropy distributed outlier class.  Our method is superior in the sense that we do not assume any distribution over the outliers.  Instead, we rely only on the inliers' distributions (available at training time), and synthesize outliers from them (at training time).  Conversely to probabilistic methods that rely on matching different inlier and outlier distributions, we use a distance-based mechanism (training a classifier on synthetic outliers sampled from the distributions and the true inliers) to classify outliers and inliers (since the distributions and the outliers themselves are unknown at training time).

\section{Proposed Method}
\label{sec:proposal}

We define outlier detection as a binary classification problem.  Such that given any sample from the inlier set $x \in \mathcal{X}$, we classify it as inlier, while any other data $y \notin \mathcal{X}$ is classified as an outlier.  However, we assume that the outliers are unknown at training time, which allows us to tackle broader problems.  This setup poses a challenge since we need negative samples to train the binary classifier.  Although a one-class classifier may be an option, we show that learning a good boundary outperforms it (\cf Section~\ref{sec:results}).

To train a binary classifier using only the inlier data, we need to synthesize outliers to act as negative samples.  We find the boundary of the training data, and then synthesize the outlier samples from it (detailed on Section~\ref{sec:outlier}).  Instead of working on the original data or feature spaces, we learn a distribution-space that represents the data changes and uncertainties (see Section~\ref{sec:distr}).  This feature distillation process helps to find better boundaries between inliers and outliers.  We detail our proposed hierarchical encoding scheme in Section~\ref{sec:features}.

\subsection{Probabilistic Representation}
\label{sec:distr}

We are interested in modeling the inliers as distributions to consider their uncertainty, in addition to compressing their representation.  Thus, we create a conditioned Gaussian distribution that represents each data point by learning a variational autoencoder~(\ref{eq:hf}), see Section~\ref{sec:features} for details.  Given an inlier $x \in \mathcal{X}$ and its latent vector $w(x)$~(\ref{eq:w}), we learn a distribution conditioned on the latent representations of the given data point $q \left( z \given w(x) \right) \equiv (\mu, \sigma)$.  In practice, we use standard Gaussians to represent the distributions, and we are interested in obtaining one distribution per data point in our training set.  In other words, we construct the set of distributions
\begin{equation}
\label{eq:q}
  \mathcal{Q} = \{(\mu, \sigma) = q \left(z \given w(x) \right) : x \in \mathcal{X}\}.
\end{equation}
That is, each distribution $q \in \mathcal{Q}$ is a Normal distribution with parameters equal to the latent representation of our VAE that corresponds to a given inlier~$x$ in the training set.

\subsection{Outlier Classification}
\label{sec:outlier}

Our synthesis process consists of perturbing inliers that are in the boundary of the training data, and using them with the inliers to train our classifier.  We perturb the inliers by drawing from the distributions that correspond to the borderline inliers.  In other words, our goal is to find the distributions from $\mathcal{Q}$~(\ref{eq:q}) that are on the outskirts of the space (as shown in Fig.~\ref{fig:diagram}).  Our first task is to find the distribution-space's boundary given the training inliers.  We propose two methods to define this boundary and its corresponding distributions.

\textbf{Ellipsoid-based.} The first boundary-defining method is based on selecting the distributions that lie on the boundary of the distribution of distributions (\ie, a meta-distribution).  To find them, we compute the center, $\bar{\mu}$, and the standard deviation, $\bar{\sigma}$, of the meta-distribution of $\mathcal{Q}$ through
\begin{align}
  \bar{\mu} &= \frac{1}{|\mathcal{Q}|} \sum_{i=1}^{|\mathcal{Q}|} \mu_i,\\
  \bar{\sigma}^2 &= \frac{1}{|\mathcal{Q}|-1} \sum_{i=1}^{|\mathcal{Q}|} (\mu_i - \bar{\mu})^2,
\end{align}
where $|\mathcal{Q}|$ is the size of the set that represents the amount of samples.  Then, our outskirt distributions are within the set
\begin{equation}
\label{eq:ellipsoid-outlier}
\mathcal{O}_e = \left\{(\mu, \sigma) : \sum_i \frac{(\mu_i - \bar{\mu}_i)^2}{\alpha \bar{\sigma}_i^2} \geq 1, (\mu, \sigma) \in \mathcal{Q} \right\},
\end{equation}
where the subscripted values represent the components of the vectors, and $\alpha$ is a scaling hyperparameter.  In other words, we select the outskirt distributions as those that lie outside of the ellipsoid defined by the scaled (by $\alpha$) mean and variance of the meta-distribution.  Note that we constructed a diagonal covariance matrix through the above procedure.

\textbf{Distance-based.} On the other hand, since we have high dimensional distributions, we would be computing their distributions on a very sparse space.  Instead, we propose to analyze the $\ell_2$ norm of their means.  Consequently, we collapse the high dimensional space into a lower one, and then we proceed in a similar fashion as above.  We compute the mean, $\bar{\mu}_\ell$, and the standard deviation, $\bar{\sigma}_\ell$, of the $\ell_2$ norms ($\norm{\cdot}_2$) of all the distributions' means on $\mathcal{Q}$, such that
\begin{align}
\bar{\mu}_\ell &= \frac{1}{|\mathcal{Q}|} \sum_{i=1}^{|\mathcal{Q}|} \norm{\mu_i}_2,\\
\bar{\sigma}^2_\ell &= \frac{1}{|\mathcal{Q}|-1} \sum_{i=1}^{|\mathcal{Q}|} (\norm{\mu_i}_2 - \bar{\mu}_\ell)^2.
\end{align}
And the outskirts distributions are within the set
\begin{equation}
\label{eq:l2-outlier}
\mathcal{O}_\ell = \left\{(\mu, \sigma) : \abs[\big]{ \norm{\mu}_2 - \bar{\mu}_\ell} \geq \alpha \bar{\sigma}_\ell, (\mu, \sigma) \in \mathcal{Q} \right\}.
\end{equation}
Similarly, the $\alpha$ parameter scales the boundary of which distributions will be selected.

\textbf{Outlier Synthesis.} Finally, given any set of distributions that lie on the outskirts of the space, $\mathcal{O}$, we can synthesize a set of outlier points that will be our negative class for training.  Hence, let the set of synthetic outlier samples be
\begin{equation}
\label{eq:outliers}
\mathcal{Y} = \left\{ y = \mu + \beta \sigma \epsilon : (\mu, \sigma) \in \mathcal{O}, \epsilon \sim \mathcal{E} \right\},
\end{equation}
where each new sample is drawn from the distributions defined in $\mathcal{O}$, $\epsilon$ is generated noise from a distribution~$\mathcal{E}$, and we use a scalar factor $\beta$ to draw from the tails of the distribution.  Geometrically, we can draw from any direction from the mean.  However, we are interested in the samples that lie on the outside part of the boundary.  \Eg, in Fig.~\ref{fig:diagram}, we draw samples on the side closer to the boundary of each of the outskirt distributions instead of sampling from any side.  Hence, we construct $\beta$ such that for each dimension its value is positive or negative depending on whether the mean of this distribution is greater or less than the mean of the meta-distribution.  We uniformly sample from all of the distributions to create enough outliers to match the amount of inliers used for training.  

\textbf{Classification.} Once we have defined our training sets, $\mathcal{X}$ and $\mathcal{Y}$, we can proceed to train a classifier $C$ by considering the projected $\mathcal{X}$ in our distribution space (in practice, we use the mean of the corresponding distribution since they represent the expected value of each inlier) as our positive examples, and $\mathcal{Y}$ as our negative ones.  During testing, we introduce new (never seen) samples to the classifier to evaluate its capabilities for outlier detection.  We present more details on Section~\ref{sec:results} regarding our experimental setup.  Note that we defined a loosely coupled process that allows us to switch between classifiers, without the need to relearn the feature distillation process.

\subsection{Feature Encoding}
\label{sec:features}

\begin{figure}
  \centering
  \resizebox{.75\linewidth}{!}{\makeatletter
\tikzset{
  shift to anchor/.code={
    \tikz@scan@one@point\pgfutil@firstofone(-\tikz@anchor)\relax
    \pgfkeysalso{shift={(-\pgf@x,-\pgf@y)}}
  }
}

\newif\if@variational
\begin{tikzpicture}[
autoencoder fill/.store in=\@ae@fill,
autoencoder fill/.default=gray!25,
autoencoder fill mid/.store in=\@ae@fill@mid,
autoencoder fill mid/.default=gray!25,
variational/.is if=@variational,
autoencoder hidden/.style={autoencoder fill, autoencoder fill mid, variational=false},
autoencoder/.style={autoencoder hidden/.append style={#1}},
autoencoder/.pic={%
  \tikzset{autoencoder hidden}
  \node[inner sep=0pt, outer sep=0pt] (\tikz@fig@name) {%
  \begin{tikzpicture}%
  \def\sep{0.08}%
  \def\width{.2}%
  \def\height{1.5}%
  \def\finalHeight{0.35}%
  \def\amount{4}%
  \def\start{0}%
  \pgfmathsetmacro{\xsep}{\sep + \width};%
  \pgfmathsetmacro{\xmiddle}{(\amount-1)*(\xsep)+\width/2};%
  \pgfmathsetmacro{\fullWidth}{(\amount*2-2)*(\xsep)+\width};%
  \pgfmathsetmacro{\ymiddle}{\height/2};%
  \pgfmathsetmacro{\step}{(\height-\finalHeight)/\amount}%
  \foreach \i in {1,...,\amount}{%
    \pgfmathsetmacro{\x}{\height - (\i-1)*\step}%
    \pgfmathsetmacro{\xoffset}{(\i-1)*\xsep}%
    \pgfmathsetmacro{\yoffset}{(\i-1)*\step/2}%
    \pgfmathparse{\i == \amount ? int(1) : int(0)}
    \ifnum\pgfmathresult=1%
      \if@variational
        \def\ymid{.5*\x+\yoffset}
        \def\varoffset{0.1}%
        \def\scalefactor{.75}
        \draw[fill=\@ae@fill@mid] (\start+\xoffset, \ymid-\varoffset) rectangle (\start+\width+\xoffset, \ymid-\scalefactor*\x-\varoffset);%
        \draw[fill=\@ae@fill@mid] (\start+\xoffset, \ymid+ \varoffset) rectangle (\start+\width+\xoffset, \ymiddle+\scalefactor*\x+\varoffset);%
      \else
        \draw[fill=\@ae@fill@mid] (\start+\xoffset,0+\yoffset) rectangle (\start+\width+\xoffset,\x+\yoffset);%
      \fi
    \else%
      \draw[fill=\@ae@fill] (\start+\xoffset,0+\yoffset) rectangle (\start+\width+\xoffset,\x+\yoffset);%
    \fi%
  }%
  \pgfmathsetmacro{\start}{(\amount-1)*(\xsep)}%
  \foreach \i in {2,...,\amount}{%
    \pgfmathsetmacro{\x}{\height - (\amount-\i)*\step}%
    \pgfmathsetmacro{\xoffset}{(\i-1)*\xsep}%
    \pgfmathsetmacro{\yoffset}{(\amount-\i)*\step/2}%
    \draw[fill=\@ae@fill] (\start+\xoffset,0+\yoffset) rectangle (\start+\width+\xoffset,\x+\yoffset);%
  }%
  \pgfmathsetmacro{\sx}{(\amount-1)*(\xsep)+\width/2}%
  \pgfmathsetmacro{\sy}{\height/2}%
  \node[anchor=center, inner sep=3pt, rounded corners, fill opacity=.85, text opacity=1] at (\xmiddle,\height){#1};%
  \end{tikzpicture}%
  };
},
vector fill/.store in=\@vector@fill,
vector fill/.default=white,
vector hidden/.style={vector fill},
vector/.style={vector hidden/.append style={#1}},
pics/vector/.style args={#1/#2/#3}{code={%
  \tikzset{vector hidden}%
  \node[fit={(0,0) (#1, #2)}, draw, inner sep=0pt, outer sep=0pt, fill=\@vector@fill] (\tikz@fig@name) at (0,0) {};%
  \node[below=2pt of \tikz@fig@name] {#3};%
}},
node distance=0.5cm,
>=latex,
semithick,
font=\footnotesize,
]%

\def\cola{Dark2-A!70}
\def\colb{Dark2-B!70}
\def\colc{Dark2-C!80}

\draw pic (v) at (0,0) {vector=.25/1.5/$x$};

\draw pic[above right=0.75cm and .75cm of v] (v1) {vector=.25/1./{$g_1(x)$}};
\draw pic[below right=0.75cm and .75cm of v] (vk) {vector=.25/1./$g_k(x)$};

\draw pic[right=of v1, autoencoder={autoencoder fill mid={\cola}}] (h1) {autoencoder=$h_1$};
\draw pic[right=of vk, autoencoder={autoencoder fill mid={\colb}}] (hk) {autoencoder=$h_k$};

\node[inner sep=0pt] at ($(v1)!0.46!(vk)$) {$\vdots$};
\node[inner sep=0pt] at ($(h1)!0.46!(hk)$) {$\vdots$};

\draw pic[right=of h1] (rv1) {vector=.25/1./{$\hat{g}_1(x)$}};
\draw pic[right=of hk] (rvk) {vector=.25/1./{$\hat{g}_k(x)$}};

\draw pic[right=.75cm of $(rv1.south east)!0.5!(rvk.north east)$t] (g) {vector=.25/.5/{}};
\draw pic[above=.25cm of g, vector={vector fill={\cola}}] (g1) {vector=.25/.5/{}};
\draw pic[below=.25cm of g, vector={vector fill={\colb}}] (g2) {vector=.25/.5/{$w = \smashoperator[lr]{\concat\limits_{i=1}^k} \tilde{g}_i$}};

\draw pic[right=of g, autoencoder={autoencoder fill mid={\colc}, variational}] (hf) {autoencoder=$h_f$};

\draw pic[right=of hf] (rec-x) {vector=.25/1.5/$\hat{w}$};

\draw[->, rounded corners] (v.east) -- ($(v)!.4!(v -| v1)$) |- (v1);
\draw[->, rounded corners] (v.east) -- ($(v)!.4!(v -| vk)$) |- (vk);

\node[draw, rectangle, rounded corners] (cls) at ($(hf |- vk)$) {$C(\tilde{g}_f)$};

\draw[->] (v1) -- (h1);
\draw[->] (vk) -- (hk);

\draw[->] (h1) -- (rv1);
\draw[->] (hk) -- (rvk);

\draw[->, rounded corners] ($(h1.south)+(0,-1.5pt)$) |- ($(g1.south west)$);
\draw[->, rounded corners] ($(hk.north)+(0,1.5pt)$) |- ($(g2.north west)$);

\draw[->] (g) -- (hf);
\draw[->] (hf) -- (rec-x);

\draw[->] ($(hf.south)+(0,10pt)$) -- (cls);

\end{tikzpicture}
\makeatother
\end{document}}
  \caption{Our feature distillation architecture uses $k$ features $\{g_i(x)\}_{i=0}^k$ from the original data point~$x$, and process them through an AE, $h_i$, for it to learn an approximated version of them, $\hat{g}_i(x)$.  Simultaneously, we concatenate the encoded features of each AE, $h_i$, namely~$\tilde{g}_i(x)$, into a vector~$w$.  Then, we pass it through a learned fusion function, $h_f$, that is a VAE to recover the input, $\hat{w}$.  The VAE encodes the parameters of the distribution of $w$, $(\mu, \sigma) \equiv q( \cdot \given w)$.  The set of these parameters for all data points is $\mathcal{Q}$.  We use samples from the encoder of~$h_f$, namely~$\tilde{g}_f$, as the positive inputs for a classifier~$C$, while the negative samples come from the synthesized outliers from the distributions at the outskirts (\cf Section~\ref{sec:outlier}).}
  \label{fig:encoding}
\end{figure}
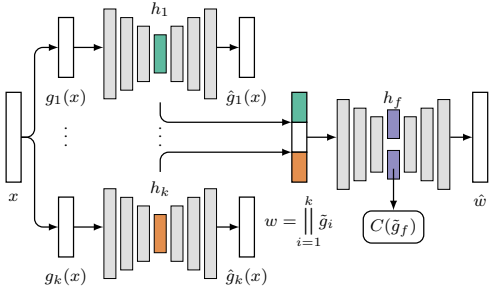

Our model for anomaly detection relies on a hierarchical feature extraction and fusion scheme based on AEs (as shown in Fig.~\ref{fig:diagram}).  Our proposal uses a two-level contraction and fusion method, as shown in Fig.~\ref{fig:encoding}.  Lets consider the raw data $x \in \mathcal{X}$, and a extracted set of $k$ features $G = \{g_i(x)\}_{i=1}^k$.  Our goal is to create a mixture of the first level features $G$ that may not have all the characteristics of $x$.  For instance, we may be using feature descriptors that are easy to compute but not expressive enough, or features learned from different tasks that need to be fused to have a more powerful descriptor, among other cases.  Hence, if we convert this data into a more suitable representation that has particular distributions or characteristics, it will be easier to classify.  Several methods can be used to achieve such space, among them, one can do dimensionality reduction or clustering followed by a mapping function.  Nevertheless, AEs have demonstrated these capabilities within a single method.  They learn an identity function with the property of embedding the input vectors into a lower-dimensional space that distills the relevant features.  Due to this property, they are the backbone of our proposed architecture.

The first level of our model consists of a set of $k$ AEs, $H = \{h_i^{\theta_i} \}_{i=1}^k$, parameterized by $\theta_i \in \Theta$, where $\Theta$ is the set of parameters for the whole model.  (In the following, we suppress the parameters from the notation of each AE for brevity.)  The job of each AE is to compact the input $g_i$ into a more expressive space, that is,
\begin{equation}
\hat{g}_i(x) = h_i \left( g_i \left(x \right) \right),
\end{equation}
the $i$th reconstructed feature $\hat{g}_i(x)$ is the output of the $i$th AE $h_i$ that operates on the input feature $g_i(x)$.  During training, one of our objectives is to minimize the error between the reconstructed and the original features, that is our feature loss
\begin{equation}
\label{eq:loss-f}
\mathcal{L}_f = \sum_{i=1}^k \norm[\big]{\hat{g}_i(x) - g_i(x)}^2.
\end{equation}

Moreover, we are interested in the compact representation learned within the AEs, as they define a space that clusters and separates the input data.  The $i$th AE, $h_i$, is the composition of encoder, $E_i$, and decoder, $D_i$, functions such that
\begin{equation}
h_i ( z ) = D_i \left( E_i \left( z \right)  \right).
\end{equation}
Lets denote the inner compact representation of the $i$th AE (the output of the encoder) as 
\begin{equation}
\tilde{g}_i (x) = E_i \left( g_i (x) \right).
\end{equation}
Then, we are interested in the learned representation
\begin{equation}
\label{eq:w}
w(x) = \concat_{i=1}^k \tilde{g}_i (x),
\end{equation}
where $\concat$ is the concatenation operator.  Our objective is to learn a transformation function, $h_f$, that converts this distilled representation into a single compact space.  Since having a sparse vectorial representation, through the AEs, is cumbersome, we plan to learn a continuous space by modeling each vector as a distribution.  Hence, we use variational inference to learn a variational family approximated through a VAE, defined by
\begin{equation}
\label{eq:hf}
\hat{w} = h_f(w),
\end{equation}
where $h_f$ comprises the encoding of $w$ into a distribution $q(z\given w) = E_f(w)$, and the decoding through a sampling process $\hat{w} \sim q(z\given w)$.  (We omit the argument $x$ from $w$ for brevity.)  And, we define the loss for the high level features as
\begin{equation}
\label{eq:loss-r}
\mathcal{L}_h = \norm{\hat{w} - w}^2 + \kl{q(z\given w)}{\mathcal{N}(0,I)},
\end{equation}
where KL is the Kullback-Leibler divergence to measure the distance between the prior and the learned distribution.  We assume that our proposed distribution, $q(z\given w)$, is distributed as a Gaussian and try to minimize its distance towards a zero-centered unitary Gaussian, $\mathcal{N}(0,I)$.

Finally, our objective is to find the set of parameters, $\Theta$, of the neural networks (the set $\left\{ H, h_f \right\}$) that minimizes the sum of the loss functions
\begin{equation}
\Theta^* = \underset{\Theta}{\argmin} \left\{\mathcal{L}_f^\Theta + \mathcal{L}_h^\Theta \right\},
\end{equation}
where the loss functions, (\ref{eq:loss-f}) and~(\ref{eq:loss-r}), are parameterized by $\Theta$ which corresponds to the parameters of AEs and VAE, $\{ H^\Theta, h_f^\Theta \}$.  These parameters are learned through back-propagation on the training phase.

\section{Results and Evaluation}
\label{sec:results}

For our experiments we are evaluating our method for outlier generation based on two different ways of selecting the outlier-generation distributions, \ie,~(\ref{eq:ellipsoid-outlier}) and~(\ref{eq:l2-outlier}), that we named $\text{OG}_e$ and $\text{OG}_\ell$, respectively.  We selected three features, namely Histogram of Oriented Gradients (HoG)~\cite{Dalal2005}, Local Binary Patterns (LBP)~\cite{Ahonen2006}, and raw pixel information as our set of features descriptors for our hierarchical learning.  For the classifier $C$, we are using a Support Vector Machine, due to training easiness.  Nevertheless, any amount of features and other classifiers can be used.  We used the area under the curve (AUC) of the Receiver Operating Characteristic curve, and $F_1$-score, which is the weighted average of the precision and recall, as our evaluation metrics for all experiments.  In both cases, the higher the score the better the methods.  The results reported in this section are from our implementation using Python~3.6.5 executed on an NVIDIA GeForce GTX 1080 Ti with the TensorFlow framework.  We detail the neural network architecture and the feature extractors hyperparameters on the supplementary material.

\subsection{Experimental Setup}
We evaluated the performance of our proposed method using MNIST~\cite{Lecun1998,Lecun2010}, Caltech~\cite{Griffin2007}, Coil-100~\cite{Nene1996}, Extended Yale B~\cite{Georghiades2001, Kclee05}, UCSD Pedestrian~\cite{Chan2008}, ISIC 2018 Challenge (Task~3)~\cite{Tschandl2018, Codella2018}, and CIFAR-10~\cite{Krizhevsky2009} datasets.

We followed the same experimental setup as previous works~\cite{Sabokrou2018, You2017} for MNIST, Caltech, Coil-100 and Extended Yale~B\@.  For UCSD Pedestrian and ISIC 2018 Challenge (Task~3) datasets, we followed Chaker \etal's~\cite{Chaker2017} and Lu and Xu's~\cite{Lu2018} setups, respectively.  For CIFAR-10, we followed previous methods~\cite{Golan2018, Hofer2019, Ruff2018} setups.  %

In general, we selected inliers from a set of classes of the dataset, and the evaluation outliers were randomly sampled from the remaining classes.  We describe the setup of each experiment per dataset on the supplementary material.  For each of our methods and per experiment, we perform a train, validation, and test (inlier) data split.  To find the best hyperparameters of the model, we used the train and validation split.  And, we evaluated in the test partition.   We report the average metrics on the test partition of $5$-fold cross-validation for the inlier data split.  Note that our training data contains no true outliers.  We train the classifier using our synthesized outliers as negative samples, as previously described, and then evaluated, during testing, on different percentages of true outliers (according to the setups of each experiment).

\subsection{Comparison against Baselines}
\label{sec:baselines}

For our experiments on MNIST, Caltech-256, Coil-100, and Yale~B, we compared our proposed methods against a set of methods, namely LOF~\cite{Breunig2000}, DRAE~\cite{Xia2015}, R and RD~\cite{Sabokrou2018}, GPND~\cite{Pidhorskyi2018}, Coherent Pursuit (CoP)~\cite{Rahmani2017},  REAPER~\cite{Lerman2015}, Outlier Pursuit (OP)~\cite{Xu2010}, LRR~\cite{Liu2010},  DPCP~\cite{Tsakiris2015}, $\ell_1$-thresh.~\cite{Soltanolkotabi2012}, R-Graph~\cite{You2017}, AnoGAN~\cite{Schlegl2017}, and  AGAN~\cite{Kimura2018}.%

In the MNIST and Caltech-256 experiments, our proposed methods obtained the highest results while maintaining a stable outlier detection rate regardless of the number of outliers, as shown in Tables~\ref{tab:mnist} and~\ref{tab:caltech}.  Moreover, our proposed method learns representations that are robust to the type of classes, as demonstrated by the results when showing them more classes in the Caltech-256 experiment (\cf Table~\ref{tab:caltech}).  
In contrast, for the Coil-100 and Yale~B experiments, which present variations, such as rotation and illumination, respectively, our proposed methods do not outperform R-Graph according to the AUC, as shown in Tables~\ref{tab:coil} and~\ref{tab:yale-b}.  Interestingly, we consistently obtain higher $F_1$ scores.  Moreover, we also noted that the variants introduced in these databases pose a great challenge for the descriptors we are using since HoG and LBP are not invariant to the point of view and intense intensity transforms of these two datasets.  Nevertheless, our method still achieves a comparable performance with existing methods.

In Table~\ref{tab:UCSD}, we reported the AUC performance of our proposed method on UCSD Pedestrian benchmark with respect to other methods, namely, Adam~\cite{Adam2008}, SF~\cite{Mehran2009}, MPPCA~\cite{Kim2009}, DTM~\cite{Mahadevan2010}, MPPCA-SF~\cite{Mahadevan2010}, Sparce~\cite{Cong2011}, SNM~\cite{Chaker2017}, LSR-VAE~\cite{Sun2018}, and GMFC-VAE~\cite{Fan2018}.  $\text{OG}_e$ outperforms the other methods in Ped2 and detects different kinds of anomalous events, such as bicycling, skateboarding, wheel-chair, \etc.  Whereas in Ped1 our methods achieve comparable results to LSR-VAE~\cite{Sun2018} but $3\%$ to $4\%$ lower than those reported by Fan \etal~\cite{Fan2018}.  We noticed that in Ped1 are several illumination variants when groups of people walk towards and away from the camera, and some amount of perspective distortion which affects the selected features.  

The AUC of our experiments on the ISIC dataset are summarized in Table~\ref{tab:ISIC2018}.  The highest AUC and more stable results were obtained by $\ell_2$-based selection method as compared to ellipsoid with an overall AUC score of $87.2\%$.  In general, our proposal outperforms other methods~\cite{Lu2018}, except on the MEL disease.  We assume that the problems are due to variations on the images that are not picked up, like in previous errors.

\begin{table}[t]
  \centering
  \sisetup{table-format=1.2}
  \setlength\tabcolsep{5pt} %
  \scriptsize
  \caption{Comparison of metrics on the MNIST dataset when varying the percentage ($\%$) of the evaluation data comprised by outliers.}
  \label{tab:mnist}
  \begin{tabular}{llSSSSSS[table-format=1.4]S[table-format=1.4]}
    \toprule
    &  {\%}  &  {LOF}  &  {DRAE}  &  {D}  &  {RD}  &  {GPND}  &  {$\text{OG}_e$}  & {$\text{OG}_\ell$}\\
    \midrule
    {AUC}  & 10 &  {\text{--}}  &  {\text{--}}  &  {\text{--}}  &  {\text{--}}  &  {\text{--}}  & 0.99181 &  \bfseries 0.99202 \\
    $F_1$  &     & 0.92 & 0.95 & 0.93 & 0.97 & 0.96 & 0.99193 & \bfseries 0.99214  \\ 
    \midrule
    {AUC}  & 20 &  {\text{--}}  &  {\text{--}}  &  {\text{--}}  &  {\text{--}}  &  {\text{--}}  & 0.99122 & \bfseries 0.99152  \\
    $F_1$  &      & 0.83 & 0.91 & 0.9 & 0.92 & 0.95 & 0.99134 & \bfseries 0.99164 \\
    \midrule
    {AUC}  & 30 &  {\text{--}}  &  {\text{--}}  &  {\text{--}}  &  {\text{--}}  &  {\text{--}}  & 0.99117 & \bfseries 0.99141  \\
    $F_1$  &      & 0.72 & 0.88 & 0.87 & 0.92 & 0.94 & 0.99129 & \bfseries 0.99154  \\
    \midrule
    {AUC}  & 40 &  {\text{--}}  &  {\text{--}}  &  {\text{--}}  &  {\text{--}}  &  {\text{--}}  & 0.99112 & \bfseries 0.99128 \\
    $F_1$  &      & 0.65 & 0.82 & 0.84 & 0.91 & 0.93 & 0.99124 & \bfseries 0.99140 \\
    \midrule
    {AUC}  & 50 &  {\text{--}}  &  {\text{--}}  &  {\text{--}}  &  {\text{--}}  &  {\text{--}}  & 0.99110 & \bfseries 0.99141 \\
    $F_1$  &      & 0.55 & 0.73 & 0.82 & 0.88 & 0.93 & 0.99122 & \bfseries 0.99154 \\
    \bottomrule
  \end{tabular}
\end{table}

\begin{table}[t]
  \centering
  \sisetup{table-format=1.3,detect-weight}
  \setlength\tabcolsep{1pt} %
  \scriptsize
  \caption{Results on the Caltech-256 dataset.  Each pair of rows shows the results when inliers were taken from 1, 3, and 5 categories.  For evaluation, $50\%$ of the data were outliers sampled from the category `clutter.'}
  \label{tab:caltech}
  \resizebox{\columnwidth}{!}{
    \begin{tabular}{lSSSSSSSSSSSSS}
      \toprule
      & {CoP} & {REAPER} & {OP} & {LRR} & {DPCP} & {$\ell_1$-thres.} & {R-Graph} & {D} & {RD}& {AGAN} & {AnoGAN} & {$\text{OG}_e$}  & {$\text{OG}_\ell$}\\
      \midrule 
      {AUC} & 0.905 & 0.816 & 0.837 & 0.907 & 0.783 & 0.772 & 0.948 & 0.932 & 0.942 & \text{--} & \text{--} & \bfseries 0.99090 & 0.98856 \\
      $F_1$ & 0.880 & 0.808 & 0.823 & 0.893 & 0.785 & 0.772 & 0.914 & 0.916 & 0.928 &  0.977 & 0.956 & \bfseries 0.99080 & 0.98842 \\
      \midrule
      {AUC} & 0.676& 0.796  & 0.788  & 0.479 & 0.798 & 0.810  & 0.929 & 0.930 &  0.938 & \text{--}  & \text{--} & 0.99562 & \bfseries 0.99717 \\
      $F_1$ &  0.718 & 0.784 &  0.779 &  0.671 &  0.777 &  0.782 &  0.880 &  0.902 & 0.913 &  0.963 & 0.915 & 0.99560 & \bfseries 0.99717 \\
      \midrule
      {AUC}&  0.487 &  0.657 &  0.629 &  0.337 &  0.676 &  0.774 &  0.913 &  0.913 & 0.923 & \text{--} & \text{--} & \bfseries 0.99776 & 0.99731 \\
      $F_1$& 0.672 & 0.716 & 0.711 & 0.667 & 0.715 & 0.762 & 0.858 & 0.890 & 0.905 & 0.945 & 0.887 & \bfseries 0.99776 & 0.99730 \\ 
      \bottomrule
  \end{tabular}}
\end{table}

\begin{table}[t]
  \centering
  \sisetup{table-format=1.3}
  \setlength\tabcolsep{1.4pt} %
  \scriptsize
  \caption{Results on the Coil-100 database.  Each pair of rows shows the results when inliers were taken from 1, 4, and 7 categories with $50\%$, $25\%$, and $15\%$ of outlier samples on the evaluation set.  Inliers categories were randomly chosen, and outliers were randomly sampled, in the aforementioned proportions, from the remaining categories (at most one from each category).}
  \label{tab:coil}
  \begin{tabular}{lSSSSSSSSSS}
    \toprule
    & {CoP} & {REAPER} & {OP} & {LRR} & {DPCP} & {$\ell_1$-thres.} & {R-Graph} & {GPND} & {$\text{OG}_e$}  & {$\text{OG}_\ell$}\\
    \midrule
    {AUC} & 0.843 & 0.9 & 0.908 & 0.847 & 0.9 & 0.991 & \bfseries 0.997 & 0.968 & 0.97241	& 0.97914\\
    $F_1$ & 0.866 & 0.892 & 0.902 & 0.872 & 0.882 & 0.978 & \bfseries 0.990 & 0.979 & 0.97155	& 0.97864 \\
    \midrule 
    {AUC} & 0.628 & 0.877 & 0.837 & 0.687 & 0.859 & 0.992 & \bfseries 0.996 & 0.945 & 0.98232	& 0.98759 \\
    $F_1$ & 0.5 & 0.703 & 0.686 & 0.541 & 0.684 & 0.941 & 0.97 & 0.96 & 0.98199 & \bfseries0.98739 \\
    \midrule  
    {AUC} & 0.58 & 0.824 & 0.822 & 0.628 & 0.804 & 0.991 & \bfseries 0.996 & 0.919 & 0.97869 & 0.97534  \\
    $F_1$ & 0.346 & 0.541 & 0.528 & 0.366 & 0.511 & 0.897 & 0.955 & 0.941 & \bfseries0.97822 & 0.97468  \\
    \bottomrule
  \end{tabular}%
\end{table}

\begin{table}[t]
  \centering
  \sisetup{table-format=1.3}
  \setlength\tabcolsep{2pt} %
  \scriptsize  
  \caption{Results on the Extended Yale B database.  Each pair of rows shows the results when inliers were taken from 1 and 3 randomly chosen subjects, and outliers were randomly chosen from the other subjects (at most one from each subject).}
  \label{tab:yale-b}
  \begin{tabular}{lSSSSSSSSS}
    \toprule
    & {CoP} & {REAPER} & {OP} & {LRR} & {DPCP} & {$\ell_1$-thres.} & {R-Graph} &  {$\text{OG}_e$}  & {$\text{OG}_\ell$}\\
    \midrule
    {AUC}  & 0.556 & 0.964 & 0.972 & 0.857 & 0.952 & 0.844 &  \bfseries 0.986  & 0.97571 & 0.95686  \\
    $F_1$  & 0.563 & 0.911 & 0.918 & 0.797 & 0.885 & 0.763 & 0.951 & \bfseries 0.97511 & 0.95449  \\
    \midrule
    {AUC}  & 0.529 & 0.932 & 0.968 & 0.807 & 0.888 & 0.848 &  \bfseries 0.985  & 0.93853 & 0.94316  \\
    $F_1$  & 0.292 & 0.758 & 0.856 & 0.509 & 0.653 & 0.545 & 0.878 & 0.93318 & \bfseries 0.94274  \\
    \bottomrule
  \end{tabular}%
\end{table}

\begin{table}[t]
  \centering
  \sisetup{table-format=1.3,detect-weight}
  \setlength\tabcolsep{1pt} %
  \scriptsize
  \caption{Results on the UCSD Pedestrian~1 and~2 datasets with the comparison of AUC performance where inliers were pedestrians and outliers were other objects.}
  \label{tab:UCSD}
  \resizebox{\columnwidth}{!}{
    \begin{tabular}{lSSSSSSSSSSS}
      \toprule
      {} & {Adam} & {SF} & {MPPCA} & {DTM} & {MPPCA-SF} & {Sparce} & {SNM} & {LSR-VAE} & {GMFC-VAE} & {$\text{OG}_e$} & {$\text{OG}_\ell$} \\ 
      \midrule
      {Ped1} & {\text{--}} & 0.675 & 0.59 & 0.818  & 0.668 & 0.86 & 0.855 & 0.902 & \bfseries{0.949} & 0.912 & 0.908 \\
      {Ped2} & 0.634 & 0.623 & 0.774 & 0.848 & 0.710 & 0.861  & 0.879 & 0.891 & 0.922  & \bfseries{0.955} & 0.900 \\
      \bottomrule
  \end{tabular}}
\end{table}

\begin{table}[t]
  \centering
  \sisetup{table-format=1.3}
  \setlength\tabcolsep{2pt} %
  \scriptsize
  \caption{Results on the ISIC 2018 Challenge (Task~3) dataset.  Each row shows the AUC when inliers were taken from NV disease, and row's category was used as an outlier.}
  \label{tab:ISIC2018}
  \begin{tabular}{lSSSSSSSS}
  \toprule
    {Disease} & $s_{\text{vae}}^{\text{reconst}}$ & ${s}_{\text{iwae}}^{\text{reconst}}$ & {${s}_{\text{vae}}^{\text{kl}}$} & ${s}_{\text{iwae}}^{\text{kl}}$ & ${s}_{\text{vae}}$ & ${s}_{\text{iwae}}$ & {$\text{OG}_e$} & {$\text{OG}_\ell$} \\
    \midrule
    {AKIEC} & 0.872 & 0.871 & 0.441 & 0.406 & 0.864 & 0.864 & 0.72702 & \bfseries{0.943} \\
    {BCC} & 0.803 & 0.802 & 0.454 & 0.431 & 0.795 & 0.795 & 0.70771 & \bfseries{0.872} \\
    {BKL} & 0.792 & 0.793 & 0.472 & 0.441 & 0.783 & 0.784 & 0.7717 & \bfseries{0.934} \\
    {DF} & 0.682 & 0.678 & 0.398 & 0.383 & 0.671 & 0.670 & 0.74139 & \bfseries{0.916} \\
    {MEL} & 0.862 & \bfseries{0.864} & 0.690 & 0.677 & 0.861 & 0.861 & 0.73641 & 0.84453 \\
    {VASC} & 0.662 & 0.657 & 0.487 & 0.477 & 0.651 & 0.648 & 0.71649 & \bfseries{0.840} \\
    {All Disease} & 0.779 & 0.777 & 0.491 & 0.469 & 0.771 & 0.771 & 0.7251 & \bfseries{0.872} \\
  \bottomrule
  \end{tabular}
\end{table}

\subsection{Ablation Study}
\label{sec:ablation}

\subsubsection{Outlier Synthesis' Parameters}  

\begin{figure*}[t]
  \centering
  \resizebox{\linewidth}{!}
  {\input{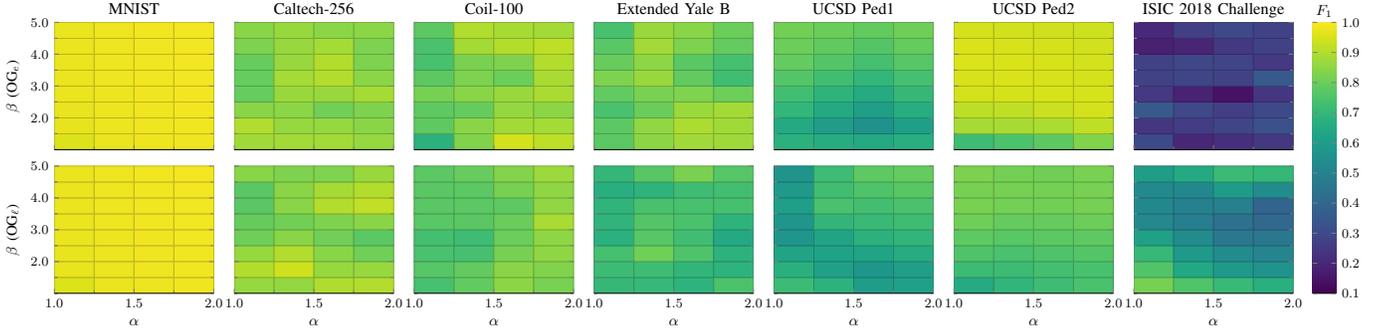}}
 \caption{%
  Comparisons of $F_1$ scores on MNIST, Caltech-256, Coil-100, Extended Yale~B, UCSD Ped1 and Ped2, and ISIC 2018 Challenge datasets by varying $\alpha$ and $\beta$ parameters when inliers come from one class and true outliers from remaining classes are used for testing (selected randomly from the remaining classes).%
}
  \label{fig:alpha-beta}
  \vspace{-10pt}
\end{figure*}

For the parameters that define the selection of distributions at the outskirts of the space,  $\alpha$,~(\ref{eq:ellipsoid-outlier}) and~(\ref{eq:l2-outlier}), and $\beta$~(\ref{eq:outliers}), we evaluated the ranges of $\alpha$ from $1$ to $2$ with step $0.25$, and $\beta$ from $1$ to $5$ with step $0.5$, on the best set of hyperparameters found on previous experiments, \cf Section~\ref{sec:baselines}.  We show in Fig.~\ref{fig:alpha-beta} the results for our metrics on all our evaluation datasets when using true outliers from the remaining classes on testing.  We show a detailed evaluation with other setups for outliers on all the datasets on the supplementary material.

We observed that selecting outskirt distributions that are more than one standard deviation away from the center of the meta-distribution yields better results, \ie, having higher values for $\alpha$.  This trend can be seen in all datasets.  However, it is clearer in Caltech, Coil and UCSD Pedestrian datasets, and a dip can be seen on MNIST when both parameters are one.  On the other hand, sampling the synthetic outliers farther from the center of the distributions, as indicated by higher $\beta$ values, shows an improvement.  However, different from $\alpha$, the variations in $\beta$ are not consistent as the former, as $\beta$'s metrics have falls on the different $\alpha$ values.  We hypothesize that, in general, these parameters will present an inverted-u behavior since they control how different the synthesized samples could be.  By increasing them, we will increase the performance up to a certain point, and then we will get diminishing returns.  Note that high values of $\alpha$ introduce outliers within the training inliers, while high values on $\beta$ may introduce outliers that are too different from any expected anomaly at prediction time.  Hence, a classifier trained on this data will learn a decision boundary that is not close enough to the normal samples.

We note the particular behavior of the metrics on Coil-100, Extended Yale-B, and ISIC datasets.  On these datasets, there are several dips and changes on the $F_1$ scores. Moreover, no patterns seem to arise from the experiments. We notice, however, that the selection of outskirts distributions varies considerably. We attribute this behavior to two factors: the variations on the datasets, and the amount of data.  On one hand, the feature descriptors we used are sensitive to the variations of these two datasets, which may be reflected in the changes in the selection of the outliers (\cf ablation results on Section~\ref{sec:other-dimensions}).  On the other hand, these datasets have considerably fewer samples, in comparison with MNIST, Caltech-256, and UCSD Pedestrian.  For example, MNIST varies between \num{5500}--\num{5900} samples, Caltech has \num{750} (when $n=5$), in comparison Coil-100 has \num{504} samples (when $n=7$) and Yale~B has only \num{192} samples (when $n=3$).  Note the variation in the classes and their relation to the number of samples of each class.

\subsubsection{Evaluation of Other Dimensions}
\label{sec:other-dimensions}

\begin{table}[t]
  \centering
  \sisetup{table-format=1.3}  
  \scriptsize
  \caption{Ablation study on the Extended Yale B database, where inliers are taken from one and three ($n$) subjects with $50\%$ and $15\%$ outlier samples from the other subjects (at most one from each subject) all randomly chosen.  We used different combinations of features, \ie, HoG~(H), LBP~(L), and raw intensity~(R); way of applying the AE, such as individually to each feature~(I), or to the concatenated features~(C), or without AE and passing the features to the next step~(WO); existence of VAE, such as with~(W) and without~(WO) it; and different outskirt distribution selection~(OS), \ie, ellipsoid~($e$), $\ell_2$-based~($\ell$), and without it~(--).  The hat OS denote experiments when the outliers were generated deterministically.  The last section show results with AE instead of VAE with outliers generated with jitter and without them~(--).  Blanks in a cell denote that the previous row's setup was used.}
\label{tab:ablation}
  \rowcolors{2}{gray!10}{white}
  \begin{tabular}{lll>{$}l<{$}SSSS}
    \toprule
    \rowcolor{white}
    \multirow{2}{*}{Feats.} & \multirow{2}{*}{AE} & \multirow{2}{*}{VAE} & \multirow{2}{*}{OS} & \multicolumn{2}{c}{$n=1$} & \multicolumn{2}{c}{$n=3$} \\
    \cmidrule{5-8}
    \rowcolor{white}
    & & & & {AUC} & {$F_1$} & {AUC} & {$F_1$} \\
    \midrule
    HLR & I & W & e & 0.976 & 0.975 & 0.939 & 0.933 \\
    & & & \ell & 0.957 & 0.954 & 0.943 & 0.943 \\
    HL &  &  & e & 0.883 & 0.884 & 0.897 & 0.902 \\
    & & & \ell & 0.914 & 0.905 & 0.903 & 0.904 \\
    HR &  &  & e & 0.950 & 0.946 & 0.938 & 0.934\\
    & & & \ell & 0.957 & 0.955 & 0.908 & 0.897\\
    LR &  &  & e & 0.957 & 0.955 & 0.884 & 0.884 \\
    & & & \ell & 0.930 & 0.923 & 0.898 & 0.891 \\
    
    HLR & C & W & e & 0.912 & 0.901 & 0.912 & 0.904 \\
    & & & \ell & 0.908 & 0.897 & 0.927 & 0.923 \\
    HL &  &  & e & 0.922 & 0.915 & 0.902 & 0.919 \\
    & & & \ell & 0.828 & 0.833 & 0.886 & 0.878 \\
    HR &  &  & e & 0.908 & 0.898 & 0.904 & 0.893 \\
    & & & \ell & 0.943 & 0.940 & 0.886 & 0.871 \\
    LR &  &  & e & 0.943 & 0.939 & 0.912 & 0.903 \\
    & & & \ell & 0.927 & 0.920 & 0.911 & 0.901 \\
    
    HLR & WO & W & e & 0.925 & 0.918 & 0.923 & 0.930 \\
    & & & \ell & 0.905 & 0.892 & 0.911 & 0.919 \\
    HL &  &  & e & 0.810 & 0.817 & 0.802 & 0.795 \\
    & & & \ell & 0.877 & 0.890 & 0.833 & 0.825 \\
    HR &  &  & e & 0.891 & 0.891 & 0.907 & 0.910 \\
    & & & \ell & 0.939 & 0.934 & 0.908 & 0.904 \\
    LR &  &  & e & 0.931 & 0.925 & 0.901 & 0.899 \\
    & & & \ell & 0.890 & 0.8931 & 0.927 & 0.921 \\
    
    HLR & C & W & \text{--} & 0.621 & 0.610 & 0.611 & 0.621 \\
    HL & & &  & 0.66222 & 0.76366 & 0.61375 & 0.44967 \\
    HR & & &  & 0.63214 & 0.72527 & 0.57596 & 0.40745 \\
    LR & & &  & 0.59250 & 0.65509 & 0.61000 & 0.44737 \\

    HLR & I & WO & \text{--} & 0.489 & 0.616 & 0.619 & 0.277 \\
    HL &  &  &  & 0.550 & 0.560 & 0.481 & 0.171 \\
    HR &  &  &  & 0.633 & 0.676 & 0.578 & 0.247 \\
    LR &  &  &  & 0.551 & 0.567 & 0.527 & 0.210 \\
    
    HLR & C & WO & \text{--} & 0.500 & 0.505 & 0.568 & 0.245 \\
    HL &  &  &  & 0.580 & 0.601 & 0.544 & 0.219 \\
    HR &  &  &  & 0.509 & 0.489 & 0.590 & 0.249 \\
    LR &  &  &  & 0.586 & 0.623 & 0.580 & 0.254 \\
    
    HLR & WO & WO & \text{--} & 0.413 & 0.394 & 0.492 & 0.183 \\
    HL &  &  &  & 0.413 & 0.394 & 0.450 & 0.177 \\
    HR &  &  &  & 0.514 & 0.563 & 0.414 & 0.179 \\
    LR &  &  &  & 0.413 & 0.394 & 0.507 & 0.195 \\
    
    \midrule
    HLR & I & W & \hat{e} & 0.601 & 0.621 & 0.611 & 0.575 \\
    & & & \hat{\ell} & 0.6 & 0.622 & 0.611 & 0.575 \\
    
    \midrule
    HLR & I & AE & \text{--} & 0.542 & 0.643 & 0.676 & 0.391 \\
    HL &  &  &  & 0.604 & 0.631 & 0.517 & 0.284 \\
    HR &  &  &  & 0.641 & 0.683 & 0.609 & 0.314 \\
    LR &  &  &  & 0.589 & 0.614 & 0.592 & 0.284 \\
    HLR & C & AE & \text{--} & 0.523 & 0.535 & 0.591 & 0.308 \\
    HL &  &  &  & 0.584 & 0.656 & 0.572 & 0.294 \\
    HR &  &  &  & 0.557 & 0.54 & 0.635 & 0.314 \\
    LR &  &  &  & 0.619 & 0.647 & 0.603 & 0.346 \\
    HLR & I & AE & \text{jitter} & 0.849 & 0.834 & 0.825 & 0.811 \\
    HL &  &  &  & 0.768 & 0.759 & 0.796 & 0.814 \\
    HR &  &  &  & 0.831 & 0.793 & 0.826 & 0.807 \\
    LR &  &  &  & 0.812 & 0.828 & 0.771 & 0.793 \\
    HLR & C & AE & \text{jitter} & 0.727 & 0.749 & 0.741 & 0.784 \\
    HL &  &  &  & 0.704 & 0.697 & 0.752 & 0.71 \\
    HR &  &  &  & 0.674 & 0.691 & 0.735 & 0.724 \\
    LR &  &  &  & 0.765 & 0.751 & 0.718 & 0.682 \\
    \bottomrule
  \end{tabular}
  \vspace*{-15pt}
\end{table}

We may think of our proposed method in terms of four different dimensions: features used, way of applying the AEs on them, the fusion through the VAE, and the outskirt-distribution selection method.  To assess our method we performed a set of experiments by varying these dimensions on the Extended Yale~B dataset, and we report the results in Table~\ref{tab:ablation}.  We selected this dataset for two reasons: size, which allows us to run several experiments on our restricted hardware, and challenging variations, which gives us a lower bound on the expected results.

For the features, we evaluate the combination of HoG~(H), LBP~(L), and raw intensity~(R) as pairs and triplets.  For the AEs, we evaluate the way of applying them to the set of inputs.  We can use different AEs per feature, \ie, individually to each feature~(I), use a single AE on the concatenation of the inputs~(C), or use no AE at all~(WO).  Moreover, we consider the setup with~(W) and without~(WO) VAE for fusion.  In the latter setup, the classification is performed over the concatenation of the extracted representations, instead of the fusion.  Lastly, we consider the use of our two proposed outskirt selection schemes, \ie, ellipsoid~($e$) and $\ell_2$-based~($\ell$).  In the case where no distributions are generated (\ie, the case without VAE for fusion), we cannot use the selection method and the concatenated features are passed to the classifier.  The cases with no synthesis are denoted with an n-dash~(--).  In them, we use a one-class SVM as a classifier due to the lack of negative samples to train a binary one.

In general, we observe that fusion is a critical component of the proposed method.  Removing the VAE (WO in the VAE column) has a significant impact on the different results regardless of the combination of the AE and features.  We, also, compared the use of an AE instead of VAE with and without synthetic outlier generation by randomly jitter the outskirt features in the AE space with noise drawn from $\mathcal{N}(0, 1.5)$, see the end of Table~\ref{tab:ablation}.  These results show a significant decrease in performance.  On the other hand, the second dimension that affects the results is the way we apply the AEs.  By encoding each feature individually we obtain consistently higher metrics.  These results align with the objective of encoding and compressing each feature individually to later fuse them.  By having an early fusion (the concatenation at the AE level), the VAE does not have richer spaces to work with.  

To evaluate the contribution of the generation of the synthetic outliers, we trained a one-class SVM using only inlier data with different feature combinations (denoted by the n-dash on the OS column).  We notice a significant drop in the metrics concerning their counterparts that use synthetic outliers for training.  Moreover, we also evaluate the case in which the distribution-space is used as features for classification (denoted by `[H][L][R] C W --' rows).  Similarly, we can see that the performance comes from the synthetic outliers that help the classifier to learn a robust boundary, in contrast to the use of more expressive features.

Interestingly, the results of HR or LR features show higher metrics in comparison to HL features.  These results indicate, again, that fusion can do so much to increase the invariants of the given inputs' features.  We can draw similar conclusions from the results of our proposed method and more complex fusion (\cf first two rows of Table~\ref{tab:ablation} and Section~\ref{sec:baselines}).  However, our proposed method can boost the metrics significantly on this task, in comparison to other methods (including concatenation and learning on the original space).

We also evaluated the case where the outliers are generated deterministically instead of using our probabilistic sampling.  In this case, we set $\epsilon=1$ in the generation~(\ref{eq:outliers}).  In that way, we will always generate the same outlier per outskirt distribution.  The corresponding deterministic counterparts of our outskirt selectors,~(\ref{eq:ellipsoid-outlier}) and~(\ref{eq:l2-outlier}), are denoted as $\hat{e}$ and $\hat{\ell}$, respectively.  The results, in the last rows of the table, show a significant reduction in the different metrics.  This demonstrates the importance of creating robust outliers for training when lacking negative samples, independently of the robust feature spaces.

In summary, the way of fusing the feature spaces is a key component in the final representation.  And our probabilistic approach seems to yield excellent results in comparison to naive fusion methods, and existing approaches that do not fuse.  Secondly, we need to represent spaces individually to take advantage of each feature space's characteristics.  Moreover, a probabilistic representation has the advantage of synthesizing robust outliers, which are necessary to learn robust and general classifiers.  Regarding the outlier selection proposals, we could not draw a line of which one is best for the task since no pattern emerged from the results.

\subsubsection{Comparison against Gold Standard}
\label{sec:gold-standard}
We also compared our proposal against what can be considered an upper bound of performance of the method.  That is, we compared against a VAE and an AE with the same architecture as the proposal, except that they were trained on true outliers.  We show the results on Table~\ref{tab:real-synth}.  We can observe that our proposal is close to the upper bound that represent the VAE and the AE results.  It is expected that the proposal achieves a lower performance due to the lack of true outliers.  Moreover, we note that the AE achieves a lower performance than its variational counterpart which also supports our decision to use a VAE as the main feature descriptor.

\begin{table}
  \sisetup{%
    table-format=1.3(3),
  }
  \caption{Comparison of the proposed method against VAE and AE when trained on true outliers (upper bound performance).  We report the AUC scores of all the methods on $5$-fold cross-validation with $20\%$ outliers.  Our proposal was trained with synthetic data, but cannot be evaluated as a one-class due to lack of outliers.  VAE and AE were trained on both binary (first four rows) and one-class (last two rows) setups. We report max-min values in parenthesis.}
  \label{tab:real-synth}
  \resizebox{\linewidth}{!}{
    \begin{tabular}{l@{ }S@{ (}S[table-format=1.3]@{, }S[table-format=1.3]@{)\ \ }S@{ (}S[table-format=1.3]@{, }S[table-format=1.3]@{)\ \ }S@{ (}S[table-format=1.3]@{, }S[table-format=1.3]@{) }}
      \toprule
      & \multicolumn{9}{c}{Dataset} \\
      \cmidrule{2-10}
      & \multicolumn{3}{c}{CIFAR 10} & \multicolumn{3}{c}{Extended Yale B} & \multicolumn{3}{c}{Coil-100} \\
      \midrule
      VAE & 0.937 (012) & 0.957 & 0.919 & 0.987 (019) & 0.998 & 0.965 & 0.997 (022) & 0.999 & 0.952 \\
      AE & 0.929 (9) & 0.941 & 0.914 & 0.965 (14) & 0.984 & 0.937 & 0.989 (7) & 0.997 & 0.974 \\
      $\text{OG}_e$ & 0.883 (31) & 0.925 & 0.832 & 0.971 (26) & 0.998 & 0.92 & 0.985 (32) & 0.998 & 0.916 \\
      $\text{OG}_\ell$ & 0.915 (23) & 0.963 & 0.878 & 0.983 (16) & 0.998 & 0.953 & 0.991 (12) & 0.998 & 0.959 \\
      \midrule
      VAE (OC) & 0.504 (2) & 0.538 & 0.48 & 0.671 (22) & 0.706 & 0.634 & 0.728 (19) & 0.74 & 0.687 \\
      AE (OC) & 0.502 (2) & 0.535 & 0.462 & 0.636 (16) & 0.661 & 0.591 & 0.711 (13) & 0.749 & 0.684 \\
      \bottomrule
  \end{tabular}}
\end{table}

\subsection{Evaluation of Other Classifiers}

\begin{table}[t]
  \centering
  \sisetup{%
    table-format=1.3(3),
  } 
  \caption{AUC scores of ellipsoid~($e$), $\ell_2$-based~($\ell$) methods on CIFAR-10 dataset where each model was trained on a single class and tested against all other classes.  We report averaged results overall classes with $5$-fold cross-validation, and different classifiers: Support Vector Machine~(SVM), Multilayer Perceptron~(MLP), Random Forest~(RF), and Naive Bayes~(NB).  We compared our AUC against other existing methods.}
  \label{tab:cifar}
  \vspace{2.5pt plus 2.5pt minus 2.5pt}
  \setlength\tabcolsep{2pt} %
  \scriptsize
  \resizebox{\columnwidth}{!}{%
    \begin{tabular}{llSSSS}
      \toprule
      {Feats.} & {OS} & {SVM} & {MLP} & {RF} & {NB} \\
      \midrule
      
      HLR & $\ell$ &  0.917(26) &  0.813(26) &  0.958(29) &  0.886(13) \\
      & e & 0.858(22) & 0.772(45) & 0.836(14) & 0.856(18) \\
      
      \cmidrule{1-6}
      
      \multicolumn{2}{l}{OC-SVM (CAE) \cite{Golan2018}} & 0.624 & \multicolumn{2}{l}{OC-SVM (E2E) \cite{Golan2018}} & 0.648\\
      \multicolumn{2}{l}{OC-SVM (RAW) \cite{Golan2018}} & 0.620 & \multicolumn{2}{l}{DAGMM \cite{Golan2018}} & 0.531 \\
      \multicolumn{2}{l}{DSEBM \cite{Golan2018}}  & 0.610 & \multicolumn{2}{l}{ADT \cite{Golan2018}} & 0.860\\
      \multicolumn{2}{l}{ADT-120 \cite{Hofer2019}} & 0.690 & \multicolumn{2}{l}{ADT-500 \cite{Hofer2019}} & 0.730\\
      \multicolumn{2}{l}{ADT-1000 \cite{Hofer2019}} & 0.750 & \multicolumn{2}{l}{COREL-120 \cite{Hofer2019}} & 0.760\\
      \multicolumn{2}{l}{AnoGAN \cite{Ruff2018}} & 0.615(27) & \multicolumn{2}{l}{Deep-SVDD \cite{Ruff2018}} & 0.648(20) \\
      \bottomrule
    \end{tabular}
  }%
\end{table}

Our previous experiments used SVM as the classifier, in this section, we explore the impact of other classifiers on our method's performance.  Namely, we additionally use Multilayer Perceptron~(MLP), Random Forest~(RF), and Naive Bayes~(NB).  Simultaneously, we evaluate our method against a set of algorithms that report results under a different dataset and metric.  We compare against One Class SVM~(OC-SVM), Deep Autoencoding Gaussian Mixture Model~(DAGMM), Deep structured energy-based models~(DSEBM), and Anomaly Detection Transformations~(ADT) reported by Golan and Ran~\cite{Golan2018}; three ADT-based models, and the COREL method reported by Hofer \etal~\cite{Hofer2019}; and AnoGAN and Deep-SVDD reported by Ruff \etal~\cite{Ruff2018}.  In this case, we evaluate our results using the area under the curve of the ROC~(AUC) on the CIFAR-10 dataset.  This change in metric and dataset is to compare against these other sets of methods.  There are more challenging datasets and versions of the CIFAR dataset.  However, not all methods report on them.  Hence, it is difficult to have a comparison against more recent methods based on those other datasets.

Our results (\cf Table~\ref{tab:cifar}) show that SVM and RF are the best classifiers on the CIFAR-10 dataset.  There is a swap on the first ranked classifier while using our proposed selection methods.  However, the results are comparable.  Moreover, our proposed methods outperform the state of the art methods regardless of the classifier used.

\section{Conclusion}
\label{sec:conclusion}

We introduced an approach that learns representations of feature descriptors, and then fuse them into a distribution-space.  This probabilistic space allows us to learn a representation of the inputs' characteristics class, and, consequently, approximate what outliers may look like.  Then, we synthesize outliers as drawings from the outskirts-distributions of the training samples.  Finally, we demonstrated that using the given inputs and the synthesized outliers yields classifiers capable of detecting outliers consistently on several datasets with varying types and amounts of outliers.  Moreover, we showed the importance of the different parts of our proposal: individual representation through compression, variational inference for fusion, and sampling for synthesis, through an ablation study.

As future work, we envision applications of our proposed idea beyond images, \ie, other domains like text, audio, etc..  We intend to explore other generative models for outlier synthesis to increase the robustness of our outliers.  Another branch to be explored is to train the synthesis and the classifiers in a tightly coupled way, \ie, an end-to-end model from data representation, distillation, outlier synthesis, and classification, to compare if this approach yields better results than a loosely coupled one, like our proposal.

\appendices
\counterwithin{figure}{section}
\counterwithin{table}{section}

\section{Experimental Setup}
\label{sec:setup}

\textbf{MNIST}~\cite{Lecun1998, Lecun2010} contains \SI{60000}{} handwritten digits from `0' to `9.'  Each of the ten digit categories is taken as the target class (\ie, inliers) and we simulate outliers (during testing) by randomly sampling images from other categories with a proportion of $10\%$ to $50\%$. This experiment is repeated for all of the ten digit categories with the images being of size $28 \times 28$.  

\textbf{Caltech-256}~\cite{Griffin2007} dataset contains 256 different object categories with a total number of \SI{30607}{} images.  Each category has at least 80.  For this dataset the images' size is $28 \times 28$ and we repeat same procedure three times by using images from $n \in \{1,3,5\}$ randomly chosen categories as inliers.  From each category the first 150 images are used only if that category has more than 150 images.  We then randomly select $50\%$ outliers from the `clutter' category which contains 827 images of different varieties in each experiment. 

\textbf{Coil-100}~\cite{Nene1996} dataset contains 100 different object categories with a total number of
\SI{7200}{} images.  Each object category has 72 images taken from different pose intervals of $5$~degrees and image's size is $32 \times 32$.  We performed three different experiments by randomly selecting $n \in \{1, 4, 7 \}$ categories images as inliers, and outliers were randomly chosen from other categories (at most one from each category) by varying percentage of outliers among $50\%$, $25\%$, and $15\%$.

\textbf{Extended Yale~B}~\cite{Georghiades2001, Kclee05} dataset contains frontal face images of 38 persons as categories.  Each category has 64 different illumination conditions.  The actual size of face images is $192 \times 168$, and for our experiment we down-sample them to $48 \times 42$.  We performed two different experiments by randomly selecting $n \in \{1, 3 \}$ categories images as inliers, and outliers are randomly chosen from other categories (at most one from each category) by varying the outliers percentage between $35\%$ and $15\%$.

\textbf{UCSD Pedestrian}~\cite{Chan2008} dataset contains two subsets of Ped1 and Ped2 with a different street scenes recorded with a stationary camera at \SI{10}{fps} and resolutions $158 \times 234$ and $240 \times 360$, respectively.  The density of the pedestrians varies from low to high.  The normal objects in all scenes are pedestrians (\ie, inliers).  All other objects (\eg, cars, skateboarders, wheelchairs, \etc) are  considered outliers.  For the experiments the images size is down-sampled to $28 \times 28$.  We performed different experiments on Ped1 scenes by randomly selecting normal object images from 34 categories as inliers, outlier images were randomly chosen from the remaining categories and each category contains 200 frames.  For Ped2 similar experiments were performed but with $16$ and $12$ object categories as inliers and outlier, respectively, where the number of frames of each clip varies in outliers.

\textbf{ISIC 2018 Challenge (Task~3)}~\cite{Tschandl2018, Codella2018} dataset contains seven different diseases as categories with a total number of 10015 images.  The actual size of disease images is $450 \times 600$, and for our experiment we down sample them to $32 \times 32$  and try to implement deep architecture with limited resources.  We performed six different experiments by considering NV disease category images as inliers and we tested outliers by select the 100 images from the rest of disease categories.

\textbf{CIFAR-10}~\cite{Krizhevsky2009} dataset consists of \num{60000} $32 \times 32$ color images in $10$ classes, with \num{6000} images per class. There are \num{50000} training images and \num{10000} test images, divided equally across the classes.  For the experiments, we use a single class for training, and the rest classes as outliers (similar to the previous protocols).  We report averaged results overall classes with $5$-fold cross-validation.

\begin{table}[t]
\centering
\caption{Amount of images on the ISIC 2018 Challenge (Task~3) Dataset.}
\label{tab:dataset4}
\sisetup{table-format=4, table-text-alignment=right}
\setlength\tabcolsep{2pt} %
\scriptsize
\begin{tabular}{lS}
 \toprule
 {Disease} &  {No.\ of Images} \\
 \midrule
   MEL & 1113 \\
   NV  & 6705 \\
   BCC & 514 \\
   AKIEC & 327 \\
   BKL & 1099 \\
   DF & 115 \\
   VASC & 142 \\
  \bottomrule
\end{tabular}
\end{table}

\section{Extended Outlier Synthesis' Parameters}

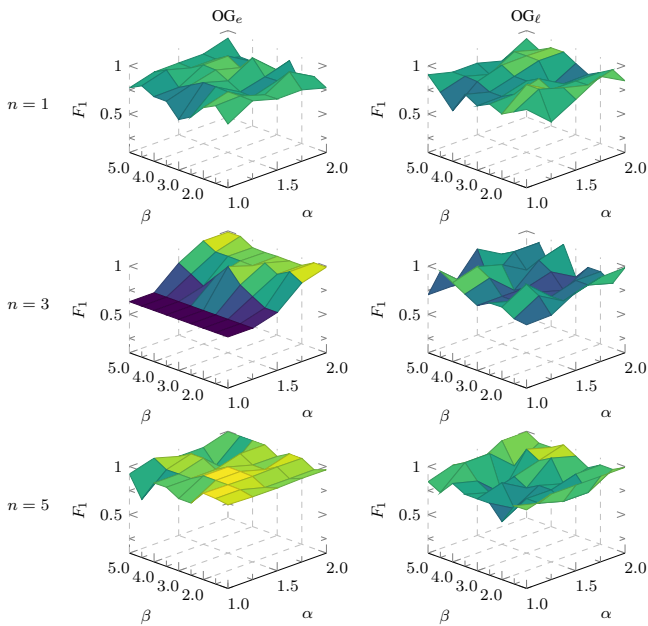
\begin{figure}[t]
  \centering
  \resizebox{\linewidth}{!}
  {\pgfplotstableread{

1	1	0.71819	0.76306	0.59755	0.63199	0.91151	0.9699	0.89764	0.91367	0.67541	0.75836	0.90124	0.91668
1	1.5	0.89168	0.90675	0.59723	0.63127	0.97932	0.98874	0.83156	0.84471	0.75153	0.77503	0.90415	0.91267
1	2	0.83009	0.84175	0.59723	0.63127	0.99776	0.99776	0.75815	0.77544	0.83727	0.85059	0.67838	0.70259
1	2.5	0.67321	0.69883	0.59723	0.63127	0.83651	0.84968	0.8188	0.83212	0.82258	0.84687	0.83546	0.84803
1	3	0.59348	0.62557	0.59723	0.63127	0.83687	0.85028	0.7475	0.76827	0.75433	0.77487	0.83455	0.84748
1	3.5	0.74895	0.7695	0.59723	0.63127	0.91591	0.9226	0.82817	0.8413	0.82945	0.84698	0.83638	0.84862
1	4	0.75505	0.77181	0.59723	0.63127	0.91144	0.92056	0.59351	0.6256	0.91257	0.91993	0.90697	0.91555
1	4.5	0.8272	0.83833	0.59723	0.63127	0.67592	0.70354	0.75569	0.77211	0.99034	0.99359	0.75632	0.77474
1	5	0.75158	0.77218	0.59723	0.63127	0.91657	0.92303	0.89963	0.90771	0.67193	0.70275	0.8273	0.84269
1.25	1	0.8984	0.91213	0.5973	0.63134	0.9257	0.96086	0.65846	0.69279	0.70027	0.76575	0.98732	0.98966
1.25	1.5	0.90981	0.91604	0.59714	0.63132	0.9819	0.98862	0.98718	0.98699	0.90342	0.92139	0.83573	0.84896
1.25	2	0.90802	0.91346	0.59714	0.63132	0.99125	0.99314	0.98856	0.98842	0.99555	0.99553	0.83648	0.84971
1.25	2.5	0.91347	0.91981	0.59714	0.63132	0.99022	0.99345	0.82686	0.84212	0.75662	0.77677	0.67456	0.70177
1.25	3	0.91164	0.91754	0.59714	0.63132	0.99295	0.9945	0.75482	0.77285	0.75729	0.77743	0.91657	0.92319
1.25	3.5	0.83437	0.84637	0.59714	0.63132	0.91441	0.921	0.75518	0.77387	0.9136	0.92228	0.75447	0.77538
1.25	4	0.75757	0.77591	0.59714	0.63132	0.91513	0.92235	0.82216	0.83592	0.99556	0.99554	0.83573	0.84896
1.25	4.5	0.90569	0.91181	0.59714	0.63132	0.91433	0.92094	0.82978	0.84511	0.75724	0.77711	0.99553	0.99551
1.25	5	0.83081	0.84343	0.59714	0.63132	0.99591	0.99589	0.83156	0.84348	0.67205	0.70259	0.91468	0.92126
1.5	1	0.82644	0.83951	0.67573	0.70255	0.92648	0.95952	0.81336	0.83449	0.85879	0.91293	0.98434	0.99079
1.5	1.5	0.82224	0.83685	0.67587	0.70313	0.91185	0.91989	0.89345	0.91007	0.75508	0.77649	0.75623	0.7767
1.5	2	0.75034	0.7693	0.75404	0.77503	0.98668	0.98831	0.83203	0.84497	0.75508	0.77606	0.83446	0.84863
1.5	2.5	0.75342	0.7752	0.83699	0.85006	0.99206	0.99461	0.90521	0.91207	0.75465	0.77571	0.83435	0.84878
1.5	3	0.83135	0.84398	0.99562	0.9956	0.99242	0.99548	0.8295	0.84237	0.7559	0.77671	0.75626	0.77721
1.5	3.5	0.9038	0.91061	0.8328	0.84727	0.91558	0.92335	0.83205	0.84403	0.67628	0.70337	0.83379	0.84848
1.5	4	0.91179	0.9174	0.67356	0.70127	0.99439	0.99435	0.90707	0.91504	0.91758	0.92384	0.59588	0.62994
1.5	4.5	0.82685	0.84148	0.75784	0.77734	0.99407	0.99404	0.75462	0.77333	0.67697	0.70409	0.99755	0.99754
1.5	5	0.83456	0.8466	0.8342	0.8479	0.91538	0.92199	0.83176	0.84448	0.99376	0.99576	0.83607	0.84976
1.75	1	0.83069	0.84227	0.91196	0.91804	0.95175	0.96887	0.97086	0.98221	0.91485	0.97701	0.98664	0.98985
1.75	1.5	0.98429	0.98402	0.98785	0.9877	0.90958	0.91912	0.83147	0.84532	0.83441	0.85049	0.75611	0.77609
1.75	2	0.83287	0.84497	0.90933	0.91562	0.99459	0.9958	0.83037	0.84158	0.83772	0.85106	0.91335	0.9206
1.75	2.5	0.82596	0.83941	0.98725	0.98707	0.91308	0.922	0.74554	0.76715	0.67662	0.70433	0.75454	0.77545
1.75	3	0.9909	0.9908	0.90589	0.91313	0.91737	0.92396	0.67583	0.70187	0.83004	0.84852	0.99572	0.9957
1.75	3.5	0.82991	0.84353	0.91123	0.91684	0.9951	0.99631	0.90866	0.91599	0.91478	0.92356	0.9153	0.92257
1.75	4	0.91154	0.91761	0.91315	0.91908	0.75565	0.77647	0.91361	0.91974	0.75308	0.77597	0.99534	0.99531
1.75	4.5	0.82757	0.84145	0.91163	0.91739	0.75634	0.77712	0.91225	0.91767	0.91301	0.92241	0.75586	0.77625
1.75	5	0.83406	0.84701	0.98725	0.98707	0.91511	0.92251	0.75389	0.77466	0.83704	0.85066	0.99452	0.99449
2	1	0.75199	0.77151	0.99013	0.99002	0.9512	0.96615	0.8266	0.84478	0.9442	0.98546	0.99173	0.99166
2	1.5	0.82958	0.84224	0.98959	0.98947	0.9149	0.92152	0.90772	0.91461	0.75484	0.77713	0.91731	0.9239
2	2	0.67442	0.70113	0.91105	0.91756	0.99055	0.99347	0.82697	0.84053	0.83578	0.85129	0.91847	0.92464
2	2.5	0.90977	0.917	0.91009	0.91679	0.99428	0.99574	0.75368	0.77335	0.83387	0.8497	0.83567	0.84969
2	3	0.75062	0.77148	0.91221	0.91895	0.91389	0.92178	0.82797	0.84113	0.67694	0.70412	0.91718	0.92384
2	3.5	0.75164	0.7721	0.90664	0.91353	0.99676	0.99675	0.90689	0.91395	0.99717	0.99717	0.9969	0.99689
2	4	0.74905	0.76894	0.91138	0.91727	0.996	0.99599	0.90624	0.91202	0.67719	0.70447	0.91746	0.92382
2	4.5	0.75012	0.77124	0.98903	0.9889	0.99525	0.99523	0.83202	0.84556	0.91415	0.92297	0.91458	0.92216
2	5	0.91462	0.92092	0.98646	0.98627	0.99602	0.996	0.90846	0.91564	0.67735	0.70442	0.99721	0.99721

}\mydata

\begin{tikzpicture}[
  font=\footnotesize,
]

\newcommand{\plot}[3][AUC]{%
  \begin{axis}[
  alpha beta 3d=#1,
  ]
  \addplot3[surf] table [x index=0, y index=1, z index=#2] {#3};
  \end{axis}}

\matrix[%
matrix of nodes, 
nodes in empty cells,
column sep=5pt,
] {%
  & $\text{OG}_e$ & $\text{OG}_\ell$ \\
  $n=1$ & \node{};\plot[$F_1$]{3}{\mydata} & \node{};\plot[$F_1$]{9}{\mydata} \\
  $n=3$ & \node{};\plot[$F_1$]{5}{\mydata} & \node{};\plot[$F_1$]{11}{\mydata} \\
  $n=5$ & \node{};\plot[$F_1$]{7}{\mydata} & \node{};\plot[$F_1$]{13}{\mydata} \\
};
\end{tikzpicture}}
  \caption{Comparisons of $F_1$ scores on Caltech-256 dataset for the proposed methods by varying $\alpha$ and $\beta$ parameters when inliers are taken to be images from $n \in \{1,3,5\}$ randomly chosen categories with $50\%$ outlier samples randomly selected from category clutter.}
  \label{fig:alpha-beta-caltech}
\end{figure}

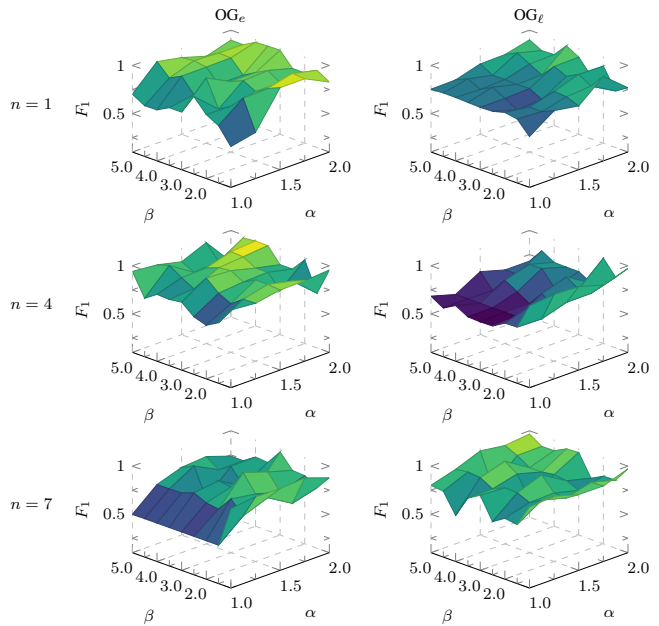
\begin{figure}[t]
  \centering
  \resizebox{\linewidth}{!}
  {\pgfplotstableread{

1	1	0.52827	0.52436	0.9165	0.90772	0.92802	0.92728	0.69451	0.62876	0.94355	0.9519	0.93791	0.94321
1	1.5	0.6728	0.69539	0.73392	0.72982	0.50014	0.49925	0.76838	0.76454	0.68187	0.68445	0.72326	0.71383
1	2	0.75026	0.76554	0.66013	0.65729	0.50014	0.49925	0.78312	0.78698	0.67997	0.67875	0.74995	0.73301
1	2.5	0.90256	0.90643	0.70946	0.69873	0.50014	0.49925	0.75786	0.7528	0.59971	0.60422	0.66589	0.65226
1	3	0.6753	0.69887	0.80626	0.80576	0.50014	0.49925	0.75838	0.74787	0.59714	0.60062	0.88749	0.88071
1	3.5	0.6753	0.69887	0.79478	0.78771	0.48903	0.49925	0.75786	0.75325	0.60183	0.60724	0.81777	0.80594
1	4	0.59784	0.63068	0.79913	0.79444	0.50014	0.49925	0.76312	0.75883	0.67232	0.67865	0.59673	0.57942
1	4.5	0.59546	0.62735	0.71444	0.7085	0.50014	0.49925	0.76286	0.76038	0.60183	0.60724	0.79258	0.78512
1	5	0.6768	0.7016	0.94524	0.94174	0.50014	0.49925	0.75786	0.7528	0.68187	0.68445	0.80795	0.80071
1.25	1	0.58184	0.57991	0.87921	0.88593	0.95348	0.95645	0.7799	0.80603	0.97942	0.97898	0.9571	0.96192
1.25	1.5	0.88564	0.89012	0.88402	0.88098	0.73088	0.71285	0.73468	0.74945	0.67022	0.6783	0.88543	0.87733
1.25	2	0.74103	0.75907	0.79	0.78931	0.84729	0.84089	0.73484	0.74946	0.7598	0.76081	0.81181	0.80434
1.25	2.5	0.66219	0.68399	0.85823	0.85186	0.7149	0.68987	0.65625	0.67653	0.59981	0.60903	0.86764	0.85957
1.25	3	0.96239	0.96077	0.96492	0.96352	0.73202	0.70852	0.73165	0.7479	0.60142	0.60997	0.87269	0.86484
1.25	3.5	0.72521	0.74071	0.81032	0.80565	0.78555	0.76695	0.81094	0.81906	0.60014	0.60815	0.81913	0.808
1.25	4	0.88436	0.88813	0.8775	0.87624	0.73857	0.72151	0.73484	0.74946	0.60142	0.60997	0.88618	0.88006
1.25	4.5	0.8114	0.82323	0.96003	0.95753	0.74118	0.71717	0.80711	0.81405	0.60142	0.60997	0.89881	0.89256
1.25	5	0.951	0.94821	0.92461	0.91798	0.73979	0.71653	0.74237	0.75811	0.60142	0.60997	0.87822	0.87477
1.5	1	0.92886	0.95274	0.93624	0.93172	0.9494	0.95033	0.74426	0.76328	0.96128	0.96772	0.8951	0.88799
1.5	1.5	0.89381	0.89801	0.9468	0.94344	0.81883	0.80948	0.8028	0.81127	0.75363	0.75798	0.81604	0.80822
1.5	2	0.80827	0.82174	0.94784	0.9441	0.80921	0.80176	0.81394	0.82175	0.66514	0.67449	0.81812	0.81235
1.5	2.5	0.88177	0.88802	0.88093	0.87624	0.97869	0.97822	0.6667	0.69124	0.75123	0.75547	0.95704	0.9549
1.5	3	0.82009	0.83294	0.95214	0.94948	0.7955	0.78202	0.74359	0.75967	0.59867	0.60575	0.66552	0.65433
1.5	3.5	0.74043	0.75957	0.93709	0.93247	0.80425	0.80022	0.73957	0.75562	0.67771	0.6826	0.87934	0.87222
1.5	4	0.88611	0.89036	0.88649	0.88595	0.59048	0.57158	0.74411	0.76144	0.75367	0.75734	0.95954	0.95763
1.5	4.5	0.90643	0.91135	0.79465	0.79371	0.86991	0.86395	0.74467	0.76018	0.59739	0.60395	0.6601	0.64621
1.5	5	0.89472	0.89825	0.85812	0.85789	0.80967	0.8035	0.82275	0.83224	0.75131	0.75557	0.97534	0.97468
1.75	1	0.93295	0.95347	0.92701	0.93689	0.97064	0.96954	0.97914	0.97864	0.89466	0.89152	0.89296	0.88611
1.75	1.5	0.96412	0.96249	0.89046	0.88795	0.67037	0.65917	0.82135	0.83145	0.75779	0.76011	0.88086	0.87607
1.75	2	0.80951	0.82034	0.74212	0.74272	0.73804	0.72911	0.90265	0.90699	0.75063	0.7574	0.75566	0.74689
1.75	2.5	0.89858	0.90126	0.88726	0.88596	0.90067	0.89354	0.8292	0.83959	0.83082	0.83626	0.8081	0.8015
1.75	3	0.82242	0.83343	0.88211	0.88387	0.677	0.6631	0.89365	0.89726	0.67532	0.68265	0.8248	0.8199
1.75	3.5	0.74298	0.76155	0.95708	0.95498	0.66816	0.65819	0.74679	0.76353	0.82729	0.83311	0.80681	0.80079
1.75	4	0.89174	0.89472	0.96474	0.96329	0.81903	0.81388	0.82041	0.83073	0.82577	0.83062	0.80726	0.79941
1.75	4.5	0.89198	0.89901	0.81656	0.81377	0.74041	0.73167	0.89677	0.90055	0.67686	0.68374	0.8847	0.88045
1.75	5	0.80819	0.82056	0.96678	0.96534	0.83192	0.82406	0.82444	0.83684	0.67398	0.68128	0.94226	0.93849
2	1	0.81524	0.82778	0.95834	0.95637	0.88513	0.87856	0.74937	0.76816	0.9741	0.97328	0.97239	0.97242
2	1.5	0.89827	0.90358	0.66384	0.6681	0.73937	0.72959	0.66718	0.68902	0.83289	0.83492	0.81691	0.80872
2	2	0.81773	0.82979	0.97044	0.96939	0.81389	0.80837	0.89575	0.90131	0.98759	0.98739	0.81231	0.80325
2	2.5	0.81921	0.83162	0.8257	0.82824	0.82648	0.81747	0.7378	0.75344	0.67903	0.68521	0.74581	0.73622
2	3	0.9636	0.97453	0.82801	0.82774	0.96538	0.96409	0.89849	0.90236	0.75115	0.75586	0.96036	0.95829
2	3.5	0.97241	0.97155	0.98071	0.98019	0.81326	0.80782	0.97519	0.97453	0.75121	0.75596	0.9526	0.94969
2	4	0.96699	0.96583	0.98232	0.98199	0.82329	0.81841	0.82705	0.83769	0.75507	0.75986	0.89051	0.88528
2	4.5	0.8897	0.89435	0.96839	0.96729	0.74753	0.73505	0.82176	0.83209	0.83079	0.83425	0.88735	0.88379
2	5	0.89256	0.89614	0.67302	0.67809	0.7462	0.73662	0.89497	0.89915	0.67452	0.68114	0.96768	0.96649

}\mydata

\begin{tikzpicture}[
  font=\footnotesize,
]

\newcommand{\plot}[3][AUC]{%
  \begin{axis}[
  alpha beta 3d=#1,
  ]
  \addplot3[surf] table [x index=0, y index=1, z index=#2] {#3};
  \end{axis}}

\matrix[%
matrix of nodes, 
nodes in empty cells,
column sep=5pt,
] {%
   & $\text{OG}_e$ & $\text{OG}_\ell$ \\
   $n=1$ & \node{};\plot[$F_1$]{3}{\mydata} & \node{};\plot[$F_1$]{9}{\mydata}  \\ 
   $n=4$ & \node{};\plot[$F_1$]{5}{\mydata} & \node{};\plot[$F_1$]{11}{\mydata}  \\ 
   $n=7$ & \node{};\plot[$F_1$]{7}{\mydata} & \node{};\plot[$F_1$]{13}{\mydata}  \\ 
};
\end{tikzpicture}}
  \caption{Comparisons of $F_1$ scores on Coil-100 dataset for the proposed methods by varying $\alpha$ and $\beta$ parameters when inliers are taken to be images from $n \in \{1,4,7\}$ randomly chosen categories with $50\%$, $25\%$, and $15\%$ outlier samples, respectively, randomly selected from the remaining categories.}
  \label{fig:alpha-beta-coil}
\end{figure}

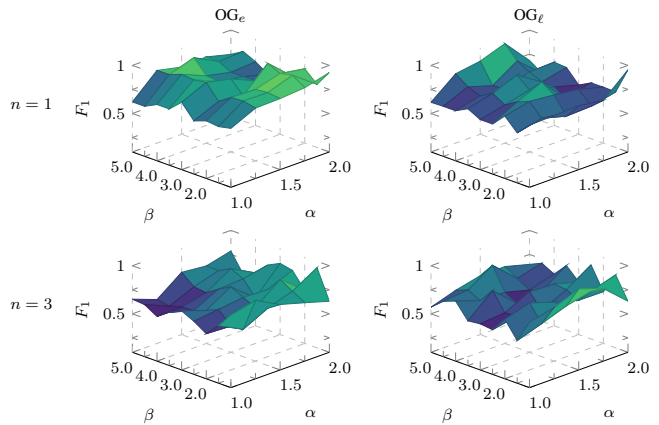
\begin{figure}[t]
  \centering
  \resizebox{\linewidth}{!}
  {\pgfplotstableread{
1	1	0.70734	0.7128	0.86306	0.85624	0.73869	0.74693	0.79261	0.78114
1	1.5	0.66863	0.68527	0.71048	0.69871	0.59893	0.62059	0.57614	0.54362
1	2	0.69502	0.69899	0.59328	0.56894	0.7385	0.7489	0.73387	0.71985
1	2.5	0.64023	0.65304	0.67364	0.6519	0.59893	0.62059	0.59042	0.56479
1	3	0.79652	0.79986	0.73399	0.7137	0.59893	0.62059	0.60042	0.57952
1	3.5	0.66863	0.68527	0.68024	0.65377	0.67098	0.68706	0.59042	0.56479
1	4	0.66863	0.68527	0.59042	0.56479	0.59893	0.62059	0.74399	0.72429
1	4.5	0.60113	0.62079	0.65452	0.64134	0.59893	0.62059	0.6969	0.67558
1	5	0.60113	0.62079	0.67078	0.6519	0.59599	0.61646	0.59328	0.56894
1.25	1	0.8584	0.83173	0.92078	0.91604	0.74844	0.75443	0.84777	0.8418
1.25	1.5	0.87665	0.85889	0.59328	0.56894	0.66642	0.67901	0.64923	0.62952
1.25	2	0.85277	0.84873	0.58495	0.55662	0.80418	0.80859	0.81435	0.79667
1.25	2.5	0.76707	0.7621	0.70959	0.68395	0.75163	0.75796	0.64685	0.6155
1.25	3	0.92225	0.91471	0.65114	0.62608	0.66938	0.68001	0.63852	0.60237
1.25	3.5	0.90742	0.89506	0.60042	0.57952	0.74734	0.75532	0.79524	0.7861
1.25	4	0.91511	0.9061	0.59328	0.56894	0.66637	0.68172	0.65828	0.63667
1.25	4.5	0.86101	0.85884	0.56899	0.53304	0.72398	0.73017	0.72852	0.70138
1.25	5	0.85639	0.85604	0.58495	0.55662	0.81717	0.82084	0.67364	0.6519
1.5	1	0.86165	0.83901	0.80185	0.7807	0.76672	0.76408	0.94316	0.94274
1.5	1.5	0.86007	0.85511	0.71724	0.69679	0.59509	0.61068	0.73471	0.72209
1.5	2	0.84835	0.84176	0.67414	0.65509	0.78571	0.7846	0.72533	0.71206
1.5	2.5	0.71912	0.72925	0.72238	0.71455	0.87113	0.87148	0.59602	0.58004
1.5	3	0.86411	0.85993	0.79819	0.77943	0.735	0.74386	0.7476	0.72994
1.5	3.5	0.782	0.78588	0.72156	0.71045	0.85912	0.85337	0.58347	0.56183
1.5	4	0.79554	0.79979	0.65243	0.62688	0.65565	0.66494	0.6457	0.63006
1.5	4.5	0.94539	0.94196	0.6524	0.64272	0.64818	0.65136	0.874	0.86641
1.5	5	0.79419	0.79906	0.76613	0.7477	0.86438	0.86077	0.72608	0.71243
1.75	1	0.86422	0.85984	0.73052	0.70726	0.72328	0.72905	0.86458	0.85392
1.75	1.5	0.94643	0.94299	0.79787	0.78018	0.60152	0.62115	0.79787	0.78018
1.75	2	0.92262	0.91518	0.86644	0.85606	0.74151	0.75047	0.6682	0.64203
1.75	2.5	0.97571	0.97511	0.72179	0.7057	0.58633	0.5997	0.71809	0.70518
1.75	3	0.73394	0.74588	0.5956	0.57273	0.66383	0.68315	0.73347	0.71573
1.75	3.5	0.73556	0.74478	0.74354	0.72539	0.67841	0.68939	0.6239	0.59615
1.75	4	0.73427	0.74393	0.79875	0.78594	0.74062	0.74806	0.58893	0.56308
1.75	4.5	0.80234	0.81032	0.74071	0.71869	0.80742	0.8163	0.65853	0.63825
1.75	5	0.80574	0.81262	0.73782	0.7221	0.95364	0.95094	0.72718	0.71565
2	1	0.93178	0.92607	0.66421	0.63072	0.95686	0.95449	0.66866	0.63111
2	1.5	0.7462	0.75712	0.93853	0.93318	0.66207	0.67859	0.86609	0.85908
2	2	0.87216	0.87324	0.66333	0.65388	0.66634	0.68166	0.65783	0.63895
2	2.5	0.73579	0.74932	0.82128	0.81352	0.66884	0.68002	0.87702	0.87089
2	3	0.72281	0.73053	0.84676	0.84074	0.59357	0.61247	0.66623	0.64716
2	3.5	0.6644	0.68013	0.79192	0.7778	0.7412	0.74866	0.78631	0.77189
2	4	0.75428	0.76175	0.66381	0.6499	0.74727	0.75797	0.65202	0.63727
2	4.5	0.82159	0.82306	0.60143	0.58913	0.58284	0.59746	0.71832	0.70484
2	5	0.73728	0.74709	0.7912	0.78549	0.6674	0.68432	0.65833	0.64671
}\mydata

\begin{tikzpicture}[
  font=\footnotesize,
]

\newcommand{\plot}[3][AUC]{%
  \begin{axis}[
  alpha beta 3d=#1,
  ]
  \addplot3[surf] table [x index=0, y index=1, z index=#2] {#3};
  \end{axis}}

\matrix[%
matrix of nodes, 
nodes in empty cells,
column sep=5pt,
] {%
  & $\text{OG}_e$ & $\text{OG}_\ell$  \\
  $n=1$ & \node{};\plot[$F_1$]{3}{\mydata} & \node{};\plot[$F_1$]{7}{\mydata} \\ 
  $n=3$ & \node{};\plot[$F_1$]{5}{\mydata} & \node{};\plot[$F_1$]{9}{\mydata} \\ 
};
\end{tikzpicture}}
  \caption{Comparisons of $F_1$ scores on Extended Yale~B dataset for the proposed methods by varying $\alpha$ and $\beta$ parameters when inliers are taken from $n \in \{1,3\}$ randomly chosen subjects, respectively, and outliers are randomly chosen from the other subjects (at most one from each subject).}
  \label{fig:alpha-beta-yale}
\end{figure}

\begin{figure}[t]
  \centering
  \resizebox{\linewidth}{!}
  {\input{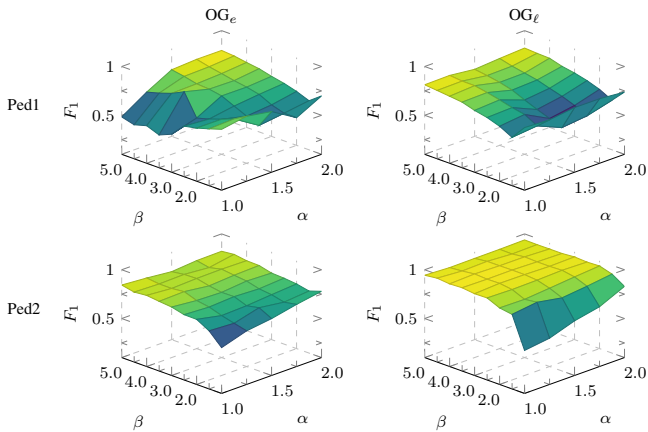}}
  \caption{Comparisons of $F_1$ scores on UCSD Pedestrian dataset for the proposed methods by varying $\alpha$ and $\beta$ parameters when inliers come from each class, and outliers are randomly selected from the remaining classes. We report the average values among all categories of video scenes.}
  \label{fig:alpha-beta-ucsd}
\end{figure}

\begin{figure*}[t]
  \centering
  {\input{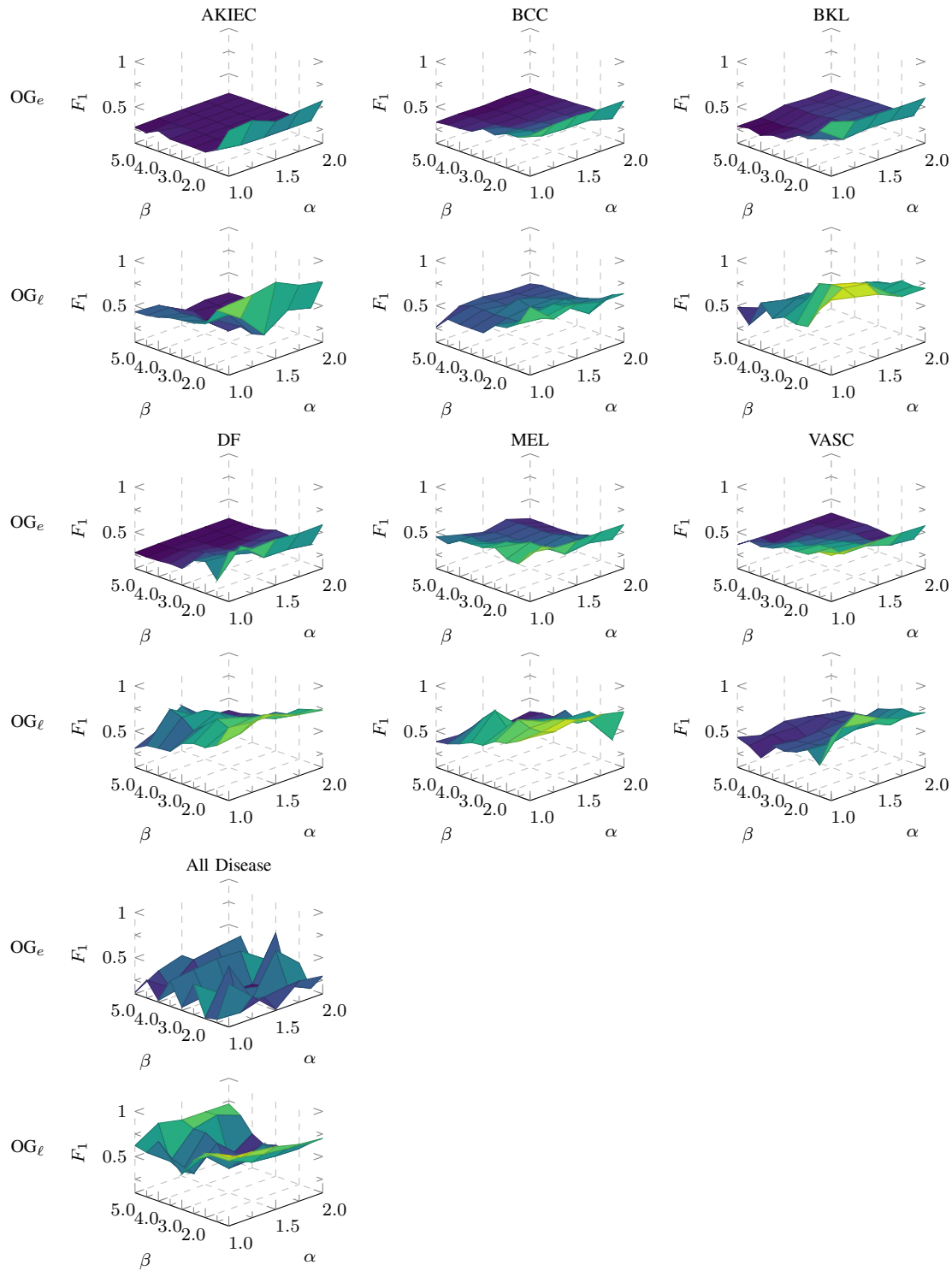}}
  \caption{Comparisons of $F_1$ scores on ISIC 2018 Challenge (Task~3) dataset for the proposed methods by varying $\alpha$ and $\beta$ parameters when inlier are taken images from NV disease category and outliers are randomly chosen from each individual disease category (show column wise, every two rows).  Last rows show the experimental results when we consider random outliers from all disease categories together.}
  \label{fig:alpha-beta-isic}
\end{figure*}

We present high resolution versions for Fig.~\ref{fig:alpha-beta}, and all configurations for all experiments in the six datasets (according to the setup shown in Appendix~\ref{sec:setup}) in Figs.~\ref{fig:alpha-beta-mnist}, \ref{fig:alpha-beta-caltech}, \ref{fig:alpha-beta-coil}, \ref{fig:alpha-beta-yale}, \ref{fig:alpha-beta-ucsd}, and~\ref{fig:alpha-beta-isic}.

\begin{figure}[t]
  \centering
  \resizebox{\linewidth}{!}
  {\pgfplotstableread{
1	1	0.91583	0.92033	0.91051	0.9146	0.90968	0.91382	0.90907	0.91321	0.90879	0.91295	0.92059	0.9236	0.91491	0.91735	0.91418	0.91666	0.91317	0.91557	0.91242	0.91472
1	1.5	0.96582	0.96681	0.96392	0.96489	0.9635	0.96448	0.96318	0.96416	0.9635	0.96452	0.96719	0.96781	0.96559	0.96619	0.96498	0.96557	0.96503	0.96564	0.96452	0.96511
1	2	0.96772	0.96813	0.96698	0.96742	0.96679	0.96725	0.96582	0.96623	0.96585	0.96627	0.97097	0.97101	0.96863	0.96858	0.96814	0.96807	0.9679	0.96782	0.96756	0.96747
1	2.5	0.97048	0.97028	0.96958	0.96937	0.96964	0.96945	0.96899	0.96876	0.9686	0.96835	0.97178	0.97162	0.96985	0.9696	0.96993	0.96971	0.96911	0.96885	0.96931	0.96907
1	3	0.97323	0.97312	0.97068	0.97046	0.97048	0.97026	0.97015	0.96992	0.97004	0.96981	0.97286	0.97282	0.97135	0.97126	0.97073	0.97061	0.97051	0.97038	0.97026	0.97012
1	3.5	0.97348	0.97307	0.97171	0.97121	0.97178	0.97129	0.97163	0.97113	0.9712	0.97068	0.97545	0.9749	0.97292	0.97223	0.97224	0.97152	0.97242	0.9717	0.97204	0.97131
1	4	0.97236	0.97218	0.97183	0.97165	0.97138	0.97119	0.97071	0.97049	0.97058	0.97034	0.97401	0.97361	0.97291	0.97246	0.972	0.97151	0.97178	0.97128	0.97195	0.97146
1	4.5	0.97303	0.97289	0.97225	0.9721	0.97155	0.97138	0.97098	0.97077	0.97102	0.97083	0.97444	0.97413	0.97314	0.97279	0.97229	0.9719	0.97233	0.97194	0.97166	0.97124
1	5	0.97436	0.97382	0.9729	0.9723	0.97311	0.97252	0.97245	0.97183	0.97271	0.97211	0.9756	0.97521	0.9733	0.9728	0.97334	0.97285	0.97306	0.97256	0.97268	0.97216
1.25	1	0.96912	0.97034	0.96723	0.96845	0.96703	0.96828	0.96658	0.96783	0.96663	0.96789	0.96941	0.9703	0.96855	0.96948	0.96829	0.96924	0.96812	0.96908	0.96747	0.9684
1.25	1.5	0.97402	0.97492	0.97182	0.97269	0.97102	0.97187	0.97109	0.97196	0.97099	0.97187	0.97337	0.97402	0.97231	0.97298	0.97181	0.97248	0.97197	0.97266	0.97151	0.97219
1.25	2	0.97605	0.97653	0.97481	0.97529	0.97441	0.9749	0.97442	0.97491	0.97404	0.97452	0.97678	0.97699	0.97498	0.97515	0.97472	0.97489	0.97476	0.97494	0.9746	0.97478
1.25	2.5	0.97756	0.97789	0.97632	0.97663	0.97589	0.9762	0.97559	0.97589	0.97555	0.97585	0.9771	0.97732	0.97605	0.97627	0.97593	0.97616	0.97545	0.97566	0.97555	0.97577
1.25	3	0.9788	0.97897	0.97784	0.97801	0.97705	0.97719	0.97693	0.97707	0.97713	0.97729	0.97839	0.97836	0.97829	0.97829	0.97739	0.97736	0.97721	0.97717	0.97727	0.97723
1.25	3.5	0.97932	0.9792	0.97868	0.97855	0.97806	0.97791	0.97788	0.97772	0.9777	0.97754	0.97955	0.97945	0.97874	0.97862	0.97759	0.97742	0.97732	0.97715	0.97723	0.97706
1.25	4	0.97966	0.97962	0.97879	0.97872	0.97819	0.9781	0.97838	0.97831	0.97792	0.97783	0.9807	0.98048	0.97935	0.97909	0.97912	0.97885	0.97888	0.97861	0.97879	0.97851
1.25	4.5	0.98013	0.98017	0.97871	0.97872	0.97864	0.97864	0.97822	0.97822	0.97817	0.97816	0.98027	0.98018	0.97901	0.97889	0.97883	0.97871	0.97814	0.978	0.97836	0.97823
1.25	5	0.98043	0.98026	0.97913	0.97892	0.97895	0.97875	0.97879	0.97858	0.97861	0.9784	0.98074	0.9804	0.97945	0.97907	0.97914	0.97875	0.97884	0.97844	0.9791	0.97871
1.5	1	0.975	0.9765	0.97313	0.97465	0.97266	0.9742	0.97272	0.97428	0.97249	0.97405	0.97573	0.97716	0.9739	0.97536	0.97393	0.97543	0.97359	0.97509	0.97349	0.97499
1.5	1.5	0.97903	0.97998	0.97748	0.97844	0.97711	0.97808	0.9768	0.97777	0.97687	0.97785	0.97899	0.97955	0.97782	0.97839	0.978	0.9786	0.97744	0.97802	0.97755	0.97814
1.5	2	0.98094	0.98111	0.97972	0.97988	0.97979	0.97996	0.97942	0.97958	0.97952	0.97969	0.981	0.98177	0.97967	0.98044	0.97901	0.97978	0.97903	0.97981	0.97885	0.97963
1.5	2.5	0.98136	0.98157	0.98135	0.98159	0.98057	0.98079	0.98081	0.98105	0.98053	0.98076	0.98158	0.98156	0.98126	0.98125	0.98073	0.9807	0.98062	0.98059	0.98052	0.98049
1.5	3	0.98309	0.98301	0.98205	0.98195	0.98201	0.98191	0.98164	0.98153	0.98183	0.98173	0.9833	0.98335	0.9823	0.98234	0.98193	0.98196	0.98189	0.98192	0.98186	0.9819
1.5	3.5	0.98292	0.9829	0.98183	0.98178	0.98169	0.98165	0.98148	0.98143	0.98174	0.9817	0.98366	0.98355	0.98257	0.98243	0.982	0.98184	0.98207	0.98192	0.982	0.98185
1.5	4	0.98423	0.98424	0.98272	0.9827	0.98253	0.9825	0.98228	0.98225	0.98211	0.98207	0.98356	0.98365	0.98231	0.98238	0.9823	0.98238	0.98204	0.98211	0.98198	0.98205
1.5	4.5	0.98399	0.9839	0.98305	0.98294	0.98246	0.98233	0.98221	0.98208	0.98244	0.98231	0.98352	0.98336	0.98297	0.9828	0.98263	0.98245	0.98256	0.98237	0.98246	0.98227
1.5	5	0.98399	0.98401	0.98294	0.98293	0.98243	0.98241	0.98233	0.98231	0.98213	0.98211	0.98408	0.98387	0.98311	0.98287	0.98296	0.98271	0.98257	0.98232	0.98242	0.98216
1.75	1	0.97951	0.9806	0.97851	0.97964	0.97792	0.97906	0.97768	0.97881	0.9776	0.97874	0.97991	0.9809	0.97924	0.98029	0.97868	0.97973	0.97837	0.97942	0.97835	0.9794
1.75	1.5	0.98262	0.98309	0.98156	0.98204	0.98159	0.98208	0.98153	0.98203	0.98133	0.98182	0.98219	0.98334	0.98146	0.98266	0.98122	0.98244	0.98064	0.98185	0.98087	0.98209
1.75	2	0.98434	0.98456	0.98342	0.98363	0.98309	0.98331	0.983	0.98322	0.98267	0.98288	0.98379	0.98425	0.98317	0.98366	0.98287	0.98336	0.98282	0.98331	0.98283	0.98333
1.75	2.5	0.98441	0.98473	0.98391	0.98425	0.98359	0.98393	0.98328	0.98362	0.98321	0.98355	0.98508	0.98537	0.98402	0.98431	0.98371	0.98399	0.9839	0.9842	0.98368	0.98398
1.75	3	0.98541	0.98558	0.98466	0.98484	0.98434	0.98451	0.98428	0.98446	0.98407	0.98424	0.98597	0.98581	0.98497	0.98479	0.98518	0.985	0.9851	0.98493	0.98489	0.98471
1.75	3.5	0.98526	0.98534	0.9853	0.9854	0.98483	0.98491	0.9848	0.98489	0.9848	0.9849	0.98639	0.98646	0.98544	0.9855	0.98517	0.98523	0.98496	0.98501	0.9851	0.98515
1.75	4	0.98636	0.98637	0.9855	0.98549	0.98522	0.98522	0.98484	0.98482	0.98517	0.98516	0.98605	0.9861	0.98569	0.98575	0.98549	0.98555	0.985	0.98505	0.98509	0.98514
1.75	4.5	0.98659	0.98671	0.98572	0.98584	0.98533	0.98545	0.98509	0.98521	0.98513	0.98524	0.98628	0.98635	0.98577	0.98584	0.98551	0.98557	0.98533	0.98539	0.98524	0.98531
1.75	5	0.98664	0.98661	0.9859	0.98586	0.98554	0.9855	0.98562	0.98557	0.9854	0.98535	0.98708	0.98706	0.98642	0.98639	0.98598	0.98594	0.98568	0.98563	0.98579	0.98574
2	1	0.98283	0.98365	0.98225	0.98311	0.98184	0.98271	0.98168	0.98255	0.98173	0.98261	0.98372	0.98415	0.98328	0.98374	0.98223	0.98266	0.98244	0.98288	0.98241	0.98286
2	1.5	0.9848	0.98548	0.98403	0.98473	0.98404	0.98475	0.98389	0.98461	0.98384	0.98456	0.9859	0.98628	0.98461	0.98499	0.98466	0.98505	0.98455	0.98495	0.98445	0.98485
2	2	0.98637	0.98683	0.98561	0.98609	0.98533	0.98582	0.98536	0.98585	0.98524	0.98574	0.98655	0.98677	0.98593	0.98615	0.98555	0.98577	0.98569	0.98591	0.98545	0.98567
2	2.5	0.98625	0.9868	0.98638	0.98698	0.98576	0.98636	0.98586	0.98646	0.98547	0.98607	0.98692	0.98728	0.98651	0.98689	0.98639	0.98678	0.98592	0.9863	0.98587	0.98625
2	3	0.98677	0.9868	0.98695	0.98699	0.98666	0.9867	0.98669	0.98674	0.98661	0.98666	0.98808	0.98808	0.98716	0.98714	0.98724	0.98723	0.98682	0.9868	0.98675	0.98674
2	3.5	0.988	0.98809	0.98743	0.98752	0.9871	0.98719	0.98701	0.9871	0.98699	0.98708	0.98835	0.98841	0.98758	0.98763	0.98748	0.98754	0.98723	0.98728	0.98719	0.98724
2	4	0.98802	0.98813	0.98774	0.98785	0.9872	0.9873	0.98728	0.98738	0.98716	0.98727	0.98778	0.98788	0.98715	0.98724	0.98699	0.98708	0.98695	0.98704	0.98702	0.98712
2	4.5	0.98822	0.98833	0.98748	0.98759	0.98732	0.98743	0.98708	0.98718	0.98711	0.98722	0.98822	0.98848	0.98764	0.98791	0.98719	0.98746	0.98722	0.9875	0.98709	0.98736
2	5	0.98851	0.98857	0.98811	0.98817	0.98789	0.98795	0.98765	0.98771	0.98757	0.98762	0.98864	0.98865	0.9876	0.9876	0.98749	0.98749	0.98731	0.98731	0.98747	0.98747

}\mydata

\begin{tikzpicture}[
]

\newcommand{\plot}[3][AUC]{%
  \begin{axis}[
    alpha beta 3d=#1,
  ]
  \addplot3[surf] table [x index=0, y index=1, z index=#2] {#3};
  \end{axis}}

\matrix[%
matrix of nodes, 
nodes in empty cells,
column sep=5pt,
] {%
  & $\text{OG}_e$ & $\text{OG}_\ell$ \\
  $10\%$ & \node{};\plot[$F_1$]{3}{\mydata} & \node{};\plot[$F_1$]{13}{\mydata} \\
  $30\%$ & \node{};\plot[$F_1$]{7}{\mydata} & \node{};\plot[$F_1$]{17}{\mydata} \\
  $50\%$ & \node{};\plot[$F_1$]{11}{\mydata} & \node{};\plot[$F_1$]{21}{\mydata} \\
};
\end{tikzpicture}}
  \caption{Comparisons of $F_1$ scores on MNIST dataset for the proposed methods by varying $\alpha$ and $\beta$ parameters when inliers come from each class, and outliers are taken from three different percentages ($\%$). We report average values among all ten classes of the digits.}
  \label{fig:alpha-beta-mnist}
\end{figure}
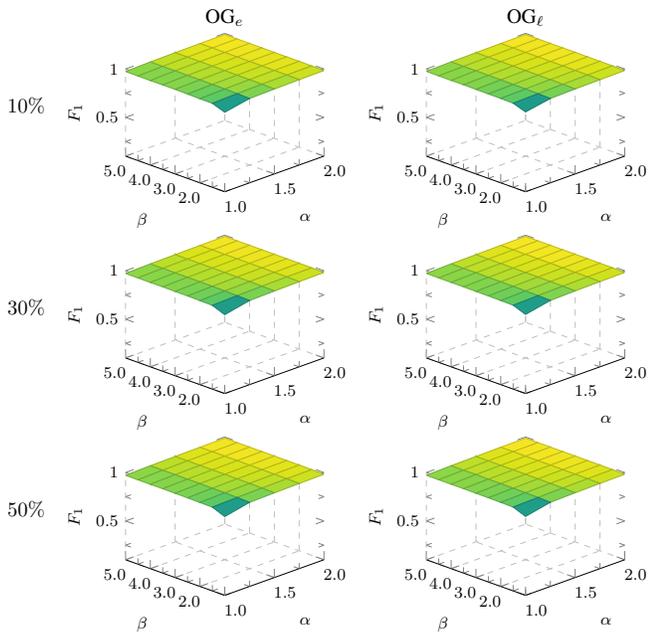

\section{Extended Results}
We show the details of the true positive rate (TPR) and false positive rate (FPR) for the MNIST, Caltech-256, Coil-100, Extended Yale-B, UCSD Pedestrian, and ISIC 2018 Challenge (Task 3) datasets experiments in Tables~\ref{tab:tpr-fpr}, \ref{tab:tpr-fpr2}, \ref{tab:tpr-fpr3}, \ref{tab:tpr-fpr4}, \ref{tab:tpr-fpr5}, and~\ref{tab:tpr-fpr6}, respectively. 

\begin{table}[t]
\centering
\caption{Results of different percentage ($\%$) of outliers on the MNIST dataset for the proposed methods: (a)~{$\text{OG}_e$}, and (b)~{$\text{OG}_\ell$}. The top row show the zero to nine digit inlier categories an d average values among all ten classes of the digits.}
\label{tab:tpr-fpr}
\sisetup{table-format=1.3,}%
\setlength\tabcolsep{2pt} %
\footnotesize%
\begin{subtable}[t]{\linewidth}%
\caption{}%
\resizebox{\columnwidth}{!}{%
  \begin{tabular}{llSSSSSSSSSSS}
    \toprule
    & $\%$ & 0 & 1 & 2 & 3 & 4 & 5 & 6 & 7 & 8 & 9 & {Avg.} \\
    \midrule
    {TPR} & 10 &  0.98444	& 0.98381 &	0.98424	& 0.98279 &	0.98537	& 0.98182 &	0.98254 & 0.98151 &	0.98631 & 0.98310 &	0.98359 \\
    {FPR} &  & 0.00000	& 0.00000 &	0.00000	& 0.00000 &	0.00000	& 0.00000 &	0.00000	& 0.00102 &	0.00000 & 0.00423 & 0.00052 \\
    \midrule
    {TPR} & 20 & 0.98588 & 0.98242 & 0.98514 & 0.98189 & 0.98601 & 0.97945 & 0.98551 & 0.98157 & 0.98581 & 0.98341 & 0.98370 \\
    {FPR} &  & 0.00000 & 0.00000 & 0.00000 & 0.00000 & 0.00000 & 0.00000 & 0.00000 & 0.00102 & 0.00000 & 0.00423 & 0.00052 \\
    
    \midrule
    {TPR} & 30 & 0.98453 & 0.98420 & 0.98370 & 0.98316 & 0.98444 & 0.98198 & 0.98405 & 0.98107 & 0.98611 & 0.98194 & 0.98351 \\ 
    {FPR} &  & 0.00000 & 0.00000 & 0.00000 & 0.00000 & 0.00000 & 0.00000 & 0.00000 & 0.00102 & 0.00000 & 0.00423 & 0.00052 \\
    \midrule
    {TPR} & 40 & 0.98321 & 0.98325 & 0.98360 & 0.98255 & 0.98388 & 0.98158 & 0.98417 & 0.98035 & 0.98646 & 0.98302 & 0.98320 \\
    {FPR} &  & 0.00000 & 0.00000 & 0.00000 & 0.00000 & 0.00000 & 0.00000 & 0.00000 & 0.00102 & 0.00000 & 0.00423 & 0.00052 \\
    \midrule
    {TPR} & 50 & 0.98351 & 0.98382 & 0.98366 & 0.98211 & 0.98410 & 0.98122 & 0.98345 & 0.98036 & 0.98593 & 0.98377 & 0.98319 \\
    {FPR} &  & 0.00000 & 0.00000 & 0.00000 & 0.00000 & 0.00000 & 0.00000 & 0.00000 & 0.00102 & 0.00000 & 0.00423 & 0.00052 \\ 
    \bottomrule
  \end{tabular}%
}
\end{subtable}
\begin{subtable}[t]{\linewidth}
\caption{}
\resizebox{\columnwidth}{!}{%
  \begin{tabular}{llSSSSSSSSSSS}
    \toprule
    & $\%$ & 0 & 1 & 2 & 3 & 4 & 5 & 6 & 7 & 8 & 9 & {Avg.} \\
    \midrule
    {TPR} & 10 & 0.98567&0.98422&0.98402&0.98559&0.98557&0.98369&0.98514&0.98131&0.98611&0.98320&0.98445 \\
    {FPR} &  & 0.00196&0.00000&0.00000&0.00000&0.00000&0.00000&0.00000&0.00000&0.00000&0.00206&0.00040 \\
    \midrule
    {TPR} & 20 & 0.98502&0.98280&0.98384&0.98282&0.98433&0.98257&0.98300&0.98011&0.98668&0.98330&0.98344 \\
    {FPR} &  & 0.00196&0.00000&0.00000&0.00000&0.00000&0.00000&0.00000&0.00000&0.00000&0.00206&0.00040 \\
    \midrule
    {TPR} & 30 & 0.98377&0.98372&0.98451&0.98316&0.98337&0.98164&0.98364&0.98093&0.98523&0.98234&0.98323 \\		
    {FPR} &  & 0.00196&0.00000&0.00000&0.00000&0.00000&0.00000&0.00000&0.00000&0.00000&0.00206&0.00040 \\
    \midrule
    {TPR} & 40 & 0.98268&0.98337&0.98399&0.98258&0.98396&0.98206&0.98333&0.98082&0.98462&0.98224&0.98296 \\
    {FPR} &  & 0.00196&0.00000&0.00000&0.00000&0.00000&0.00000&0.00000&0.00000&0.00000&0.00206&0.00040 \\
    \midrule
    {TPR} & 50 & 0.98451&0.98306&0.98390&0.98267&0.98406&0.98137&0.98426&0.98132&0.98571&0.98144&0.98322 \\
    {FPR} &  & 0.00196&0.00000&0.00000&0.00000&0.00000&0.00000&0.00000&0.00000&0.00000&0.00206&0.00040 \\
    \bottomrule
  \end{tabular}%
}
\end{subtable}
\end{table}

\begin{table}[t]
\centering
\caption{Results of $50\%$ randomly chosen outliers from category 257-clutter of the Caltech-256 dataset for the proposed methods: (a)~{$\text{OG}_e$}, and (b)~{$\text{OG}_\ell$}.  The top row shows the 2, 4, 6, 8, and 10 inlier categories and average values among all five classes of the objects.}
\label{tab:tpr-fpr2}
\sisetup{table-format=1.3,}
\setlength\tabcolsep{2pt} %
\footnotesize
\begin{subtable}[t]{0.3\textwidth}
\caption{}
\resizebox{\columnwidth}{!}{%
  \begin{tabular}{llSSSSSSS}
    \toprule
    & $\%$ & 2 & 4 & 6 & 8 & 10 & {Avg.} \\
    \midrule
    {TPR} & 50 &  0.976 & 0.894 & 0.901 & 0.899 & 0.971 & 0.928 \\
    {FPR} &  & 0.000 & 0.092 & 0.084 & 0.086 & 0.000 & 0.052 \\
    \bottomrule
  \end{tabular}%
}
\end{subtable}
\begin{subtable}[t]{0.3\textwidth}
\caption{}
\resizebox{\columnwidth}{!}{%
  \begin{tabular}{llSSSSSSS}
    \toprule
    & $\%$ & 2 & 4 & 6 & 8 & 10 & {Avg.} \\
    \midrule
    {TPR} & 50 &  0.979 & 0.894 & 0.896 & 0.898 & 0.971 & 0.927 \\
    {FPR} &  & 0.000 & 0.087 & 0.084 & 0.087 & 0.094 & 0.070\\
    \bottomrule
  \end{tabular}%
}
\end{subtable}
\end{table}

\begin{table}[t]
\centering
\caption{Results of $50\%$, $25\%$, and $15\%$ outliers of the Coil-100 dataset for the proposed methods: (a)~{$\text{OG}_e$}, and (b)~{$\text{OG}_\ell$}. The top row shows the 2, 4, 6, 8, and 10 inlier categories and average values among all five classes of the objects.}
\label{tab:tpr-fpr3}
\sisetup{table-format=1.3,}
\setlength\tabcolsep{2pt} %
\footnotesize
\begin{subtable}[t]{0.3\textwidth}
\caption{}
\resizebox{\columnwidth}{!}{%
  \begin{tabular}{llSSSSSSS}
    \toprule
    & $\%$ & 2 & 4 & 6 & 8 & 10 & {Avg.} \\
    \midrule
    {TPR} & 50 & 0.933 & 0.928 & 0.934 & 0.922 & 0.949 & 0.933\\
    {FPR} &  & 0.069 & 0.075 & 0.068 & 0.080 & 0.040 & 0.066 \\
    \midrule
    {TPR} & 25 & 0.959 & 0.972 & 0.955 & 0.940 & 0.964 & 0.958\\
    {FPR} &  & 0.041 & 0.021 & 0.040 & 0.059 & 0.029 & 0.038 \\
    \midrule
    {TPR} & 15 & 0.877 & 0.860 & 0.870 & 0.884 & 0.856 & 0.869\\
    {FPR} &  & 0.118 & 0.126 & 0.109 & 0.114 & 0.1392 & 0.121\\
    \bottomrule
  \end{tabular}%
}
\end{subtable}
\begin{subtable}[t]{0.3\textwidth}
\caption{}
\resizebox{\columnwidth}{!}{%
  \begin{tabular}{llSSSSSSS}
    \toprule
    & $\%$ & 2 & 4 & 6 & 8 & 10 & {Avg.} \\
    \midrule
    {TPR} & 50 & 0.903 & 0.893 & 0.888 & 0.892 & 0.924 & 0.900\\
    {FPR} &  & 0.086 & 0.093 & 0.102 & 0.090 & 0.074 & 0.089\\
    \midrule
    {TPR} & 25 & 0.900 & 0.903 & 0.968 & 0.890 & 0.924 & 0.917\\
    {FPR} &  & 0.069 & 0.072 & 0.039 & 0.092 & 0.072 & 0.068\\
    \midrule
    {TPR} & 15 & 0.837 & 0.850 & 0.917 & 0.876 & 0.860 & 0.868\\
    {FPR} &  & 0.154 & 0.102 & 0.073 & 0.118 & 0.129 & 0.115 \\
    \bottomrule
  \end{tabular}%
}
\end{subtable}
\end{table}

\begin{table}[t]
\centering
\caption{Results of $35\%$ and $15\%$ outliers of the Extended Yale B dataset for the proposed methods: (a)~{$\text{OG}_e$}, and (b)~{$\text{OG}_\ell$}. The top row shows the 2, 4, 6, 8, and 10 inlier categories and average values among all five classes of the objects.}
\label{tab:tpr-fpr4}
\sisetup{table-format=1.3,}
\setlength\tabcolsep{2pt} %
\footnotesize
\begin{subtable}[t]{0.3\textwidth}
\caption{}
\resizebox{\columnwidth}{!}{%
  \begin{tabular}{llSSSSSSS}
    \toprule
    & $\%$ & 2 & 4 & 6 & 8 & 10 & {Avg.} \\
    \midrule
    {TPR} & 35 &  0.941 & 0.870 & 0.844 & 0.921 & 0.913 & 0.897\\
    {FPR} &  & 0.049 & 0.075 & 0.083 & 0.076 & 0.074 & 0.061 \\
    \midrule
    {TPR} & 15 &  0.882 & 0.860 & 0.862 & 0.910 & 0.876 & 0.869 \\
    {FPR} &  &  0.048 & 0.079 & 0.088 & 0.059 & 0.083 & 0.070 \\    
    \bottomrule
  \end{tabular}%
}
\end{subtable}
\begin{subtable}[t]{0.3\textwidth}
\caption{}
\resizebox{\columnwidth}{!}{%
  \begin{tabular}{llSSSSSSS}
    \toprule
    & $\%$ & 2 & 4 & 6 & 8 & 10 & {Avg.} \\
    \midrule
    {TPR} & 35 &  0.932 & 0.852 & 0.837 & 0.901 & 0.923 & 0.889 \\
    {FPR} &  &  0.037 & 0.092 & 0.088 & 0.081 & 0.074 & 0.067 \\
    \midrule
    {TPR} & 15 &   0.899 & 0.907 & 0.872 & 0.908 & 0.891 & 0.882 \\
    {FPR} &  & 0.053 & 0.049 & 0.052 & 0.069 & 0.068 & 0.062 \\    
    \bottomrule
  \end{tabular}%
}
\end{subtable}
\end{table}

\begin{table}[t]
\centering
\caption{Results of randomly chosen outliers of the UCSD Pedestrian dataset (on both sets Ped1 and Ped2) for the proposed methods: (a)~{$\text{OG}_e$}, and (b)~{$\text{OG}_\ell$}.  The top row shows the 1, 3, 5, 7, and 9 inlier categories and average values among all five classes of the scenes.}
\label{tab:tpr-fpr5}
\sisetup{table-format=1.3,}
\setlength\tabcolsep{2pt} %
\footnotesize
\begin{subtable}[t]{0.35\textwidth}
\caption{}
\resizebox{\columnwidth}{!}{%
  \begin{tabular}{lllSSSSSSS}
    \toprule
    & Subset & & & 1 & 3 & 5 & 7 & 9 & {Avg.}\\
    \midrule
    & Ped-1 & {TPR} & & 0.9358 & 0.9463 & 0.9512 & 0.9390 & 0.9721 & 0.9488\\
    & & {FPR} &  & 0.0186 & 0.0145 & 0.0109 & 0.0498 & 0.0059 & 0.0199\\
    \midrule
    & Ped-2 & {TPR} & & 0.9693 & 0.9631 & 0.9345 & 0.8859 & 0.9741 & 0.9453\\
    & & {FPR} &  & 0.0063 & 0.0174 & 0.0063 & 0.0971 & 0.0154 & 0.0285\\
    \bottomrule
    \end{tabular}%
}
\end{subtable}
\begin{subtable}[t]{0.35\textwidth}
\caption{}
\resizebox{\columnwidth}{!}{%
  \begin{tabular}{lllSSSSSSS}
    \toprule
    & Scene & & & 1 & 3 & 5 & 7 & 9 & {Avg.}\\
    \midrule
    & Ped-1 & {TPR} & & 0.9431 & 0.9275 & 0.9470 & 0.9161 & 0.9384 & 0.9344\\
    & & {FPR} &  & 0.0220 & 0.0208 & 0.0491 & 0.0126 & 0.0048 & 0.0218\\
    \midrule
    & Ped-2 & {TPR} & & 0.9610 & 0.9851 & 0.9619 & 0.9878 & 0.9228 & 0.9637\\
    & & {FPR} &  & 0.0026 & 0.0491 & 0.0067 & 0.0065 & 0.0018 & 0.0133\\
    \bottomrule
  \end{tabular}%
}
\end{subtable}
\end{table}

\begin{table}[t]
\centering
\caption{Results of randomly chosen outliers of the ISIC 2018 Challenge (Task~3) dataset for the proposed methods: (a)~{$\text{OG}_e$}, and (b)~{$\text{OG}_\ell$}.  The top row shows the disease categories for each outlier category, and average values among all.}
\label{tab:tpr-fpr6}
\sisetup{table-format=1.3,}
\setlength\tabcolsep{2pt} %
\footnotesize
\begin{subtable}[t]{0.35\textwidth}
\caption{}
\resizebox{\columnwidth}{!}{%
  \begin{tabular}{lllSSSSSSSS}
    \toprule
    & & & {AKIEC} & {BCC} & {BKL} & {DF} & {MEL} & {VASC} & {All} & {Avg.}\\
    \midrule
    & & {TPR} & 0.4508 & 0.4146 & 0.4590 & 0.4824 & 0.4811 & 0.43778 & 0.41510 & 0.4486\\
    & & {FPR} & 0.0074 & 0.0018 & 0.0170 & 0.0204 & 0.0067 & 0.0080 & 0.0240 & 0.0091\\
    \bottomrule
  \end{tabular}%
}
\end{subtable}
\begin{subtable}[t]{0.35\textwidth}
\caption{}
\resizebox{\columnwidth}{!}{%
  \begin{tabular}{lllSSSSSSSS}
    \toprule
    & & & {AKIEC} & {BCC} & {BKL} & {DF} & {MEL} & {VASC} & {All} & {Avg.}\\
    \midrule
    & & {TPR} & 0.6357 & 0.6757 & 0.8789 & 0.8451 & 0.6661 & 0.7028 & 0.7445 & 0.7355\\
    & & {FPR} & 0.0125 & 0.0201 & 0.0105 & 0.0173 & 0.0157 & 0.0181 & 0.0201 & 0.0164\\
    \bottomrule
  \end{tabular}%
}
\end{subtable}
\end{table}

\section{Implementation Details}

Our proposed framework comprises three main modules: deterministic and variational autoencoders (as shown in Fig.~\ref{fig:encoding}), and the classifier. 

For the feature encoding, we use denoising autoencoders~\cite{Vincent2008}. There are four hyperparameters that we used for each denoising autoencoder: number of layers, neurons per layer, type of activation function, and type of reconstruction loss. We evaluated an amount of layers among $1$, $3$, $5$, and $7$. For the number of neurons, we used three settings `same,' `double,' and `half' number of neurons of the input per layer.  The decoder is a mirror of the encoder in terms of layer structure. We evaluated four types of activation functions: ReLU, sigmoid, tanh, and linear. And for the reconstruction loss function~(\ref{eq:loss-f}), we used mean squared error (MSE) or binary cross entropy.  If the input values are in $[0, 1]$, then we use cross entropy, otherwise we use MSE\@.

For the fusion, we use a VAE with a Gaussian distribution. There are three hyperparameters that we tuned for our VAE: number of layers, neurons per layer, and dimensionality of the Gaussian. Similarly to the AEs, we used between $1$ and $2$ layers, with `same' and `double' amount of neurons of the input per layer. And the sizes of the Gaussians vary from $2$ to $6$ (with step one)---we selected small dimensions due to resource constraints for executing the experiments.

There are three hyper-parameters that we tuned for our SVMs: kernel type, kernel coefficient ($\gamma$), and regularization. We evaluated four types of kernels: linear, polynomial, RBF, and sigmoid.  For the kernel coefficient, we used values $1$, $\SI{e-1}{}$, $\SI{e-2}{}$, $\SI{e-3}{}$, $\SI{e-4}{}$, and $\SI{e-5}{}$.  And for the regularization parameter ($C$) we used $\SI{e-3}{}$, $\SI{e-2}{}$, $\SI{e-1}{}$, $1$, and $10$.

\section{Hyperparameters}

\subsection{Features}

We show the hyperparameters we used for extracting the features, HOG~\cite{Dalal2005} and LBP~\cite{Ahonen2006}, in Tables~\ref{tab:HOGhyperparameters} and~\ref{tab:LBPhyperparameters}.

\subsection{Autoencoders (AEs), Variational Autoencoders (VAEs) and SVM}

We show the best set of hyperparameters found on experimental results for MNIST, Caltech-256, Coil-100, Extended Yale-B, UCSD Pedestrian, and ISIC 2018 Challenge (Task~3) on Tables~\ref{tab:MNISThyperparameters}, \ref{tab:Caltechhyperparameters}, \ref{tab:Coilhyperparameters}, \ref{tab:YaleBhyperparameters}, \ref{tab:UCSDhyperparameters}, and~\ref{tab:Diseasehyperparameters}, respectively.  The corresponding original results (on the paper) are in Tables~\ref{tab:mnist}, \ref{tab:caltech}, \ref{tab:coil}, \ref{tab:yale-b}, \ref{tab:UCSD}, and~\ref{tab:ISIC2018}.

\begin{table}[t]
  \centering
  \caption{HOG's hyperparameters. The best cased used in our proposed experiment is typeset in bold.  Abbreviations:  cells per block (CpB), orientations (O).} 
  \label{tab:HOGhyperparameters}
  \sisetup{table-format=1.3}
  \setlength\tabcolsep{3pt} %
  \footnotesize
  \begin{tabular}{lllll}
    \toprule
    {CpB} & {O} & \multicolumn{3}{c}{Pixels per Cell} 	\\
    \midrule
    & & {(12 , 12)} & {(14 , 14)} & {(16 , 16)} \\
    \midrule
    {(1 , 1)} & {9} & {(1,1) - 9 - (12,12)} & \textbf{{(1,1) - 9 - (14,14)}} & {(1,1) - 9 - (16,16)} \\
    & {12} & {(1,1) - 12 - (12-12)} & {(1,1) - 12 - (14,14)} & {(1,1) - 12 - (16,16)} \\
    & {15} & {(1,1) - 15 - (12,12)} & {(1,1) - 15 - (14,14)} & {(1,1) - 15 - (16,16)} \\
    \midrule
    {(2 , 2)} & {9} & {(2,2) - 9 - (12,12)} & {(2,2) - 9 - (14,14)} & {(2,2) - 9 - (16,16)} \\
    & {12} & {(2,2) - 12 - (12,12)} & {(2,2) - 12 - (14,14)} & {(2,2) - 12 - (16,16)} \\
    & {15} & {(2,2) - 15 - (12,12)} & {(2,2) - 15 - (14,14)} & {(2,2) - 15 - (16,16)} \\
    \midrule
    {(4 , 4)} & {9} & {(4,4) - 9 - (12,12)} & {(4,4) - 9 - (14,14)} & {(4,4) - 9 - (16,16)} \\
    & {12} & {(4,4) - 12 - (12,12)} & {(4,4) - 12 - (14,14)} & {(4,4) - 12 - (16,16)} \\
    & {15} & {(4,4) - 15 - (12,12)} & {(4,4) - 15 - (14,14)} & {(4,4) - 15 - (16,16)} \\
    \bottomrule
  \end{tabular}%
\end{table}

\begin{table}[t]
  \centering
  \caption{LBP's hyperparameters. The best result is typeset in bold.  Abbreviations: number of patterns (NP).}
  \label{tab:LBPhyperparameters}
  \setlength\tabcolsep{3pt} %
  \footnotesize
  \begin{tabular}{lllll}
    \toprule
    {Method} & {\# NP} & \multicolumn{3}{c}{Radius of circle} \\
    \midrule
    & & {8} & {16} & {24}\\
    \midrule	
    Default	& 4 & {Default-4-8} & {Default-4-16} & {Default-4-24} \\
    & 6 & {Default-6-8} & {Default-6-16} & {Default-6-24} \\
    & 8 & {Default-8-8} & {Default-8-16} & {Default-8-24} \\
    \midrule
    ROR	& 4 & {ROR-4-8} & {ROR-4-16} & {ROR-4-24} \\
    & 6 & {ROR-6-8} & {ROR-6-16} & {ROR-6-24} \\
    & 8 & {ROR-8-8} & {ROR-8-16} & {ROR-8-24} \\
    \midrule
    Uniform	& 4 & \textbf{{Uniform-4-8}} & {Uniform-4-16} & {Uniform-4-24} \\ 
    & 6 & {Uniform-6-8} & {Uniform-6-16} & {Uniform-6-24} \\
    & 8 & {Uniform-8-8} & {Uniform-8-16} & {Uniform-8-24} \\
    \midrule
    VAR	& 4 & {VAR-4-8} & {VAR-4-16} & {VAR-4-24} \\ & 6 & {VAR-6-8} & {VAR-6-16} & {VAR-6-24} \\
    & 8 & {VAR-8-8} & {VAR-8-16} & {VAR-8-24} \\
    \bottomrule
  \end{tabular}%
\end{table}

\begin{table}[t]
  \centering
  \caption{Best set of hyperparameters for MNIST dataset in proposed experiments. Abbreviations: Amount of Layers (AL), Number of neurons per layer (NN), Activation Function (AF), Latent Dimension (LD), Kernel coefficient ($\gamma$), and regularization parameter ($C$).}
  \label{tab:MNISThyperparameters}
  \setlength\tabcolsep{3pt} %
  \footnotesize
  \begin{tabular}{llllllllll}
    \toprule
    {Method} & \multicolumn{3}{c}{AutoEncoder} & \multicolumn{3}{c}{Variational AutoEncoder} & \multicolumn{3}{c}{SVM}\\
    \midrule
    & {AL} & {NN} & {AF} & {LD} & {Alpha} & {Beta} & {Kernel} & {$\gamma$} & {$C$}\\
    \midrule	
    {$\text{OG}_e$}	& {1}& {Same}& {ReLU} & {2} & {3.0} & {4.5} & {rbf} & {1} & {0.1}\\
    {$\text{OG}_\ell$} & {1}& {Same}& {ReLU} & {2} & {3.0} & {5.0} & {rbf} & {1} & {0.1}\\
    \bottomrule
  \end{tabular}%
\end{table}

\begin{table}[t]
  \centering
  \caption{Best set of hyperparameters for Caltech-256 dataset in proposed experiments.  Abbreviations: Amount of Layers (AL), Number of neurons per layer (NN), Activation Function (AF), Latent Dimension (LD), Kernel coefficient ($\gamma$) and regularization parameter ($C$).}
  \label{tab:Caltechhyperparameters}
  \setlength\tabcolsep{3pt} %
  \footnotesize
  \begin{tabular}{llllllllll}
    \toprule
    {Method} & \multicolumn{3}{c}{AutoEncoder} & \multicolumn{3}{c}{Variational AutoEncoder} & \multicolumn{3}{c}{SVM}\\
    \midrule
    & {AL} & {NN} & {AF} & {LD} & {Alpha} & {Beta} & {Kernel} & {$\gamma$} & {$C$}\\
    \midrule	
    {$\text{OG}_e$}	& {5}& {Half}& {ReLU} & {2} & {1.75} & {3.0} & {rbf} & {1} & {0.1}\\
    {$\text{OG}_\ell$} & {5}& {Half}& {ReLU} & {2} & {1.25} & {2.0} & {rbf} & {1} & {0.1}\\

    \midrule	
    {$\text{OG}_e$}	& {5}& {Same}& {ReLU} & {4} & {1.5} & {3.0} & {rbf} & {1} & {0.1}\\
    {$\text{OG}_\ell$} & {7}& {Same}& {ReLU} & {2} & {2.0} & {3.5} & {rbf} & {1} & {0.1}\\
    
    \midrule	
    {$\text{OG}_e$}	& {5}& {Same}& {ReLU} & {2} & {1.0} & {2.0} & {rbf} & {1} & {0.1}\\
    {$\text{OG}_\ell$} & {7}& {Same}& {ReLU} & {4} & {1.5} & {4.5} & {rbf} & {1} & {0.1}\\
    \bottomrule
  \end{tabular}%
\end{table}

\begin{table}[t]
  \centering
  \caption{Best set of hyperparameters for Coil-100 dataset in proposed experiments. Abbreviations: Amount of Layers(AL), Number of neurons per layer (NN), Activation Function (AF), Latent Dimension (LD), Kernel coefficient ($\gamma$) and regularization parameter ($C$).}
  \label{tab:Coilhyperparameters}
  \setlength\tabcolsep{3pt} %
  \footnotesize
  \begin{tabular}{llllllllll}
    \toprule
    {Method} & \multicolumn{3}{c}{AutoEncoder} & \multicolumn{3}{c}{Variational AutoEncoder} & \multicolumn{3}{c}{SVM}\\
    \midrule
   & {AL} & {NN} & {AF} & {LD} & {Alpha} & {Beta} & {Kernel} & {$\gamma$} & {$C$}\\    \midrule	
    {$\text{OG}_e$}	& {5}& {Half}& {ReLU} & {2} & {2.0} & {3.5} & {rbf} & {1} & {0.1}\\
    {$\text{OG}_\ell$} & {1}& {Same}& {ReLU} & {2} & {1.75} & {1.0} & {rbf} & {1} & {0.1}\\

    \midrule	
    {$\text{OG}_e$}	& {1}& {Double}& {ReLU} & {3} & {2.0} & {4.0} & {rbf} & {1} & {0.1}\\
    {$\text{OG}_\ell$} & {3}& {Same}& {ReLU} & {2} & {2.0} & {2.0} & {rbf} & {1} & {0.1}\\
    
    \midrule	
    {$\text{OG}_e$}	& {1}& {Double}& {ReLU} & {3} & {1.5} & {2.5} & {rbf} & {1} & {0.1}\\
    {$\text{OG}_\ell$} & {3}& {Half}& {ReLU} & {2} & {1.5} & {5.0} & {rbf} & {1} & {0.1}\\
    \bottomrule
  \end{tabular}%
\end{table}

\begin{table}[t]
  \centering
  \caption{Best set of hyperparameters for Extended Yale B dataset in proposed experiments. Abbreviations: Amount of Layers(AL), Number of neurons per layer(NN), Activation Function (AF), Latent Dimension (LD), Kernel coefficient ($\gamma$) and regularization parameter ($C$).}
  \label{tab:YaleBhyperparameters}
  \setlength\tabcolsep{3pt} %
  \footnotesize
  \begin{tabular}{llllllllll}
    \toprule
    {Method} & \multicolumn{3}{c}{AutoEncoder} & \multicolumn{3}{c}{Variational AutoEncoder} & \multicolumn{3}{c}{SVM}\\
    \midrule
    & {AL} & {NN} & {AF} & {LD} & {Alpha} & {Beta} & {Kernel} & {$\gamma$} & {$C$}\\
    \midrule	
    {$\text{OG}_e$}	& {3}& {Half}& {ReLU} & {3} & {1.75} & {2.5} & {rbf} & {1} & {25}\\
    {$\text{OG}_\ell$} & {1}& {Half}& {ReLU} & {4} & {2.0} & {1.0} & {rbf} & {1} & {25}\\

    \midrule	
    {$\text{OG}_e$}	& {5}& {Half}& {ReLU} & {3} & {2.0} & {1.5} & {rbf} & {1} & {0.1}\\
    {$\text{OG}_\ell$} & {5}& {Half}& {ReLU} & {3} & {1.5} & {1.0} & {rbf} & {1} & {0.1}\\
    \bottomrule
  \end{tabular}%
\end{table}

\begin{table}[t]
  \centering
  \caption{Best set of hyperparameters for UCSD Pedestrian dataset in proposed experiments. Abbreviations: Amount of Layers(AL), Number of neurons per layer (NN), Activation Function (AF), Latent Dimension (LD), Kernel coefficient ($\gamma$) and regularization parameter ($C$).}
  \label{tab:UCSDhyperparameters}
  \setlength\tabcolsep{3pt} %
  \footnotesize
  \begin{tabular}{llllllllll}
    \toprule
    {Method} & \multicolumn{3}{c}{AutoEncoder} & \multicolumn{3}{c}{Variational AutoEncoder} & \multicolumn{3}{c}{SVM}\\
\midrule	
& {AL} & {NN} & {AF} & {LD} & {Alpha} & {Beta} & {Kernel} & {$\gamma$} & {$C$}\\
\midrule	
{$\text{OG}_e$}	& {5}& {Half} & {1} & {2} & {2} & {2.5} & {rbf} & {1} & {10}\\
{$\text{OG}_\ell$}	& {3}& {Half} & {1} & {2} & {2} & {2.5} & {rbf} & {1} & {10}\\
\midrule
{$\text{OG}_e$}	& {1}& {Half} & {1} & {3} & {1.25} & {2.0} & {rbf} & {1} & {10}\\
{$\text{OG}_\ell$}	& {5} & {Half}& {1} & {3} & {1.75} & {4.0} & {rbf} & {1} & {10}\\    
    \bottomrule
  \end{tabular}%
\end{table}

\begin{table}[t]
  \centering
  \caption{Best set of hyperparameters for ISIC 2018 Challenge (Task~3) dataset in proposed experiments. Abbreviations: Amount of Layers (AL), Number of neurons per layer (NN), Activation Function (AF), Latent Dimension (LD), Kernel coefficient ($\gamma$) and regularization parameter ($C$).}
  \label{tab:Diseasehyperparameters}
  \setlength\tabcolsep{3pt} %
  \footnotesize
  \begin{tabular}{llllllllll}
    \toprule
    {Method} & \multicolumn{3}{c}{AutoEncoder} & \multicolumn{3}{c}{Variational AutoEncoder} & \multicolumn{3}{c}{SVM}\\
\midrule
& {AL} & {NN} & {AF} & {LD} & {Alpha} & {Beta} & {Kernel} & {$\gamma$} & {$C$}\\
\midrule	
{$\text{OG}_e$}	& {5}& {Half}& {1} & {12} & {1.25} & {1.0} & {rbf} & {1} & {1}\\
{$\text{OG}_\ell$}	& {3}& {Half}& {1} & {16} & {1.5} & {1.0} & {rbf} & {1} & {0.1}\\

\midrule
{$\text{OG}_e$}	& {5}& {Half}& {1} & {11} & {1.5} & {1.0} & {rbf} & {1} & {1}\\
{$\text{OG}_\ell$}	& {3}& {Half}& {1} & {14} & {1.0} & {1.0} & {rbf} & {1} & {1}\\

\midrule
{$\text{OG}_e$}	& {3}& {Half}& {1} & {10} & {1.0} & {1.0} & {rbf} & {1} & {0.1}\\
{$\text{OG}_\ell$}	& {3}& {Half}& {1} & {15} & {1.0} & {1.0} & {rbf} & {1} & {1}\\

\midrule
{$\text{OG}_e$}	& {5}& {Half}& {1} & {16} & {1.0} & {1.0} & {rbf} & {1} & {1}\\
{$\text{OG}_\ell$}	& {3}& {Half}& {1} & {14} & {1.0} & {1.0} & {rbf} & {1} & {1}\\

\midrule
{$\text{OG}_e$}	& {5}& {Half}& {1} & {17} & {1.0} & {1.0} & {rbf} & {1} & {1}\\
{$\text{OG}_\ell$}	& {3}& {Half}& {1} & {17} & {1.0} & {1.5} & {rbf} & {1} & {1}\\

\midrule
{$\text{OG}_e$}	& {3}& {Half}& {1} & {12} & {1.0} & {1.0} & {rbf} & {1} & {1}\\
{$\text{OG}_\ell$}	& {3}& {Half}& {1} & {18} & {1.5} & {1.0} & {rbf} & {0.1} & {0.1}\\

\midrule
{$\text{OG}_e$}	& {5}& {Half}& {1} & {17} & {1.0} & {1.0} & {rbf} & {1} & {0.1}\\
{$\text{OG}_\ell$}	& {3}& {Half}& {1} & {16} & {1.0} & {1.0} & {rbf} & {1} & {1}\\

    \bottomrule
  \end{tabular}%
\end{table}

\bibliographystyle{ieeetr}
\bibliography{abrv,references}

\begin{thebibliography}{10}

\bibitem{Ahmed2016}
M.~Ahmed, A.~N. Mahmood, and M.~R. Islam, ``A survey of anomaly detection
  techniques in financial domain,'' {\em Fut. Gen. Comput. Syst.}, vol.~55,
  pp.~278--288, 2016.

\bibitem{Chandola2009}
V.~Chandola, A.~Banerjee, and V.~Kumar, ``Anomaly detection: A survey,'' {\em
  {ACM} Comput. Surv.}, vol.~41, no.~3, p.~15, 2009.

\bibitem{Hodge2004}
V.~Hodge and J.~Austin, ``A survey of outlier detection methodologies,'' {\em
  Artif. Intell. Rev.}, vol.~22, no.~2, pp.~85--126, 2004.

\bibitem{Pimentel2014}
M.~A. Pimentel, D.~A. Clifton, L.~Clifton, and L.~Tarassenko, ``A review of
  novelty detection,'' {\em Signal Process.}, vol.~99, pp.~215--249, 2014.

\bibitem{Buades2005}
A.~Buades, B.~Coll, and J.-M. Morel, ``A non-local algorithm for image
  denoising,'' in {\em {IEEE}/{CVF} Inter. Conf. Comput. Vis. Pattern Recog.
  ({CVPR})}, vol.~2, pp.~60--65, IEEE, 2005.

\bibitem{Cong2011}
Y.~Cong, J.~Yuan, and J.~Liu, ``Sparse reconstruction cost for abnormal event
  detection,'' in {\em {IEEE}/{CVF} Inter. Conf. Comput. Vis. Pattern Recog.
  ({CVPR})}, pp.~3449--3456, IEEE, 2011.

\bibitem{Li2014}
W.~Li, V.~Mahadevan, and N.~Vasconcelos, ``Anomaly detection and localization
  in crowded scenes,'' {\em {IEEE} Trans. Pattern Anal. Mach. Intell.},
  vol.~36, no.~1, pp.~18--32, 2014.

\bibitem{Sabokrou2017}
M.~Sabokrou, M.~Fayyaz, M.~Fathy, and R.~Klette, ``Deep-cascade: cascading {3D}
  deep neural networks for fast anomaly detection and localization in crowded
  scenes,'' {\em {IEEE} Trans. Image Process.}, vol.~26, no.~4, pp.~1992--2004,
  2017.

\bibitem{Sabokrou2018}
M.~Sabokrou, M.~Khalooei, M.~Fathy, and E.~Adeli, ``Adversarially learned
  one-class classifier for novelty detection,'' in {\em {IEEE}/{CVF} Inter.
  Conf. Comput. Vis. Pattern Recog. ({CVPR})}, pp.~3379--3388, 2018.

\bibitem{Xia2015}
Y.~Xia, X.~Cao, F.~Wen, G.~Hua, and J.~Sun, ``Learning discriminative
  reconstructions for unsupervised outlier removal,'' in {\em {IEEE}/{CVF}
  Inter. Conf. Comput. Vis. Pattern Recog. ({CVPR})}, pp.~1511--1519, 2015.

\bibitem{You2017}
C.~You, D.~P. Robinson, and R.~Vidal, ``Provable selfrepresentation based
  outlier detection in a union of subspaces,'' in {\em {IEEE}/{CVF} Inter.
  Conf. Comput. Vis. Pattern Recog. ({CVPR})}, pp.~1--10, 2017.

\bibitem{Eskin2000}
E.~Eskin, ``Anomaly detection over noisy data using learned probability
  distributions,'' in {\em Inter. Conf. Mach. Learn. ({ICML})}, 2000.

\bibitem{Kim2012}
J.~Kim and C.~D. Scott, ``Robust kernel density estimation,'' {\em J. Mach.
  Learn. Res.}, vol.~13, no.~Sep, pp.~2529--2565, 2012.

\bibitem{Lerman2015}
G.~Lerman, M.~B. McCoy, J.~A. Tropp, and T.~Zhang, ``Robust computation of
  linear models by convex relaxation,'' {\em Found. Comput. Math.}, vol.~15,
  no.~2, pp.~363--410, 2015.

\bibitem{Markou2003}
M.~Markou and S.~Singh, ``Novelty detection: a review---part 1: statistical
  approaches,'' {\em Signal Process.}, vol.~83, no.~12, pp.~2481--2497, 2003.

\bibitem{Yamanishi2004}
K.~Yamanishi, J.-I. Takeuchi, G.~Williams, and P.~Milne, ``On-line unsupervised
  outlier detection using finite mixtures with discounting learning
  algorithms,'' {\em Data Min. Knowl. Discov.}, vol.~8, no.~3, pp.~275--300,
  2004.

\bibitem{Bodesheim2013}
P.~Bodesheim, A.~Freytag, E.~Rodner, M.~Kemmler, and J.~Denzler, ``Kernel null
  space methods for novelty detection,'' in {\em {IEEE}/{CVF} Inter. Conf.
  Comput. Vis. Pattern Recog. ({CVPR})}, pp.~3374--3381, 2013.

\bibitem{Breunig2000}
M.~M. Breunig, H.-P. Kriegel, R.~T. Ng, and J.~Sander, ``{LOF}: Identifying
  density-based local outliers,'' in {\em {ACM} Conf. Manag. Data ({ACM
  SIGMOD})}, vol.~29, pp.~93--104, ACM, 2000.

\bibitem{Liu2017}
J.~Liu, Z.~Lian, Y.~Wang, and J.~Xiao, ``Incremental kernel null space
  discriminant analysis for novelty detection,'' in {\em {IEEE}/{CVF} Inter.
  Conf. Comput. Vis. Pattern Recog. ({CVPR})}, pp.~792--800, 2017.

\bibitem{Rahmani2017}
M.~Rahmani and G.~K. Atia, ``Coherence pursuit: Fast, simple, and robust
  principal component analysis,'' {\em {IEEE} Trans. Signal Process.}, vol.~65,
  no.~23, pp.~6260--6275, 2017.

\bibitem{Sabokrou2016}
M.~Sabokrou, M.~Fathy, and M.~Hoseini, ``Video anomaly detection and
  localisation based on the sparsity and reconstruction error of
  auto-encoder,'' {\em Electron. Lett.}, vol.~52, no.~13, pp.~1122--1124, 2016.

\bibitem{Hasan2016}
M.~Hasan, J.~Choi, J.~Neumann, A.~K. Roy-Chowdhury, and L.~S. Davis, ``Learning
  temporal regularity in video sequences,'' in {\em {IEEE}/{CVF} Inter. Conf.
  Comput. Vis. Pattern Recog. ({CVPR})}, pp.~733--742, IEEE, 2016.

\bibitem{Kimura2018}
M.~Kimura and T.~Yanagihara, ``Semi-supervised anomaly detection using {GANs}
  for visual inspection in noisy training data,'' {\em arXiv:1807.01136}, 2018.

\bibitem{Pidhorskyi2018}
S.~Pidhorskyi, R.~Almohsen, D.~A. Adjeroh, and G.~Doretto, ``Generative
  probabilistic novelty detection with adversarial autoencoders,'' in {\em Adv.
  Neural Inf. Process. Sys. ({NeurIPS})}, 2018.

\bibitem{Ravanbakhsh2017}
M.~Ravanbakhsh, M.~Nabi, E.~Sangineto, L.~Marcenaro, C.~Regazzoni, and N.~Sebe,
  ``Abnormal event detection in videos using generative adversarial nets,'' in
  {\em {IEEE} Inter. Conf. Image Process. ({ICIP})}, pp.~1577--1581, IEEE,
  2017.

\bibitem{Sabokrou2015}
M.~Sabokrou, M.~Fathy, M.~Hoseini, and R.~Klette, ``Real-time anomaly detection
  and localization in crowded scenes,'' in {\em {IEEE} Inter. Conf. Comput.
  Vis., Pattern Recog. Wksps. ({CVPRW})}, pp.~56--62, 2015.

\bibitem{Schlegl2017}
T.~Schlegl, P.~Seeb{\"o}ck, S.~M. Waldstein, U.~Schmidt-Erfurth, and G.~Langs,
  ``Unsupervised anomaly detection with generative adversarial networks to
  guide marker discovery,'' in {\em Inter. Conf. Inf. Process. Medical Imag.
  ({IPMI})}, pp.~146--157, Springer, 2017.

\bibitem{Wang2018}
H.-g. Wang, X.~Li, and T.~Zhang, ``Generative adversarial network based novelty
  detection usingminimized reconstruction error,'' {\em Front. Inf. Technol.
  Electron. Eng.}, vol.~19, no.~1, pp.~116--125, 2018.

\bibitem{Dalal2005}
N.~Dalal and B.~Triggs, ``Histograms of oriented gradients for human
  detection,'' in {\em {IEEE}/{CVF} Inter. Conf. Comput. Vis. Pattern Recog.
  ({CVPR})}, vol.~1, pp.~886--893, IEEE, 2005.

\bibitem{Ojala1996}
T.~Ojala, M.~Pietik{\"a}inen, and D.~Harwood, ``A comparative study of texture
  measures with classification based on featured distributions,'' {\em Pattern
  Recogn.}, vol.~29, no.~1, pp.~51--59, 1996.

\bibitem{Huang2011}
D.~Huang, C.~Shan, M.~Ardabilian, Y.~Wang, and L.~Chen, ``Local binary patterns
  and its application to facial image analysis: a survey,'' {\em {IEEE} Trans.
  Syst., Man, Cybern. {C}}, vol.~41, no.~6, pp.~765--781, 2011.

\bibitem{RamirezRivera2013}
A.~Ram\'irez~Rivera, J.~Rojas~Castillo, and O.~Chae, ``Local directional number
  pattern for face analysis: Face and expression recognition,'' {\em {IEEE}
  Trans. Image Process.}, vol.~22, no.~5, pp.~1740--1752, 2013.

\bibitem{RamirezRivera2015a}
A.~Ram\'irez~Rivera and O.~Chae, ``Spatiotemporal directional number
  transitional graph for dynamic texture recognition,'' {\em {IEEE} Trans.
  Pattern Anal. Mach. Intell.}, vol.~37, no.~10, pp.~2146--2152, 2015.

\bibitem{RamirezRivera2015}
A.~Ram\'irez~Rivera, J.~Rojas~Castillo, and O.~Chae, ``Local directional
  texture pattern image descriptor,'' {\em Pattern Recogn. Lett.}, vol.~51,
  no.~0, pp.~94--100, 2015.

\bibitem{Venhuizen2015}
F.~G. Venhuizen, B.~van Ginneken, B.~Bloemen, M.~J. van Grinsven, R.~Philipsen,
  C.~Hoyng, T.~Theelen, and C.~I. S{\'a}nchez, ``Automated age-related macular
  degeneration classification in oct using unsupervised feature learning,'' in
  {\em Med. Imag. {CAD}}, vol.~9414, p.~94141I, International Society for
  Optics and Photonics, 2015.

\bibitem{Morris2011}
B.~T. Morris and M.~M. Trivedi, ``Trajectory learning for activity
  understanding: Unsupervised, multilevel, and long-term adaptive approach,''
  {\em {IEEE} Trans. Pattern Anal. Mach. Intell.}, vol.~33, no.~11,
  pp.~2287--2301, 2011.

\bibitem{Erfani2016}
S.~M. Erfani, S.~Rajasegarar, S.~Karunasekera, and C.~Leckie,
  ``High-dimensional and large-scale anomaly detection using a linear one-class
  svm with deep learning,'' {\em Pattern Recogn.}, vol.~58, pp.~121--134, 2016.

\bibitem{Xu2015}
D.~Xu, E.~Ricci, Y.~Yan, J.~Song, and N.~Sebe, ``Learning deep representations
  of appearance and motion for anomalous event detection,'' {\em
  arXiv:1510.01553}, 2015.

\bibitem{Xu2010}
H.~Xu, C.~Caramanis, and S.~Sanghavi, ``Robust {PCA} via outlier pursuit,'' in
  {\em Adv. Neural Inf. Process. Sys. ({NeurIPS})}, pp.~2496--2504, 2010.

\bibitem{Park2018}
J.~Park, D.~H. Choi, Y.-B. Jeon, Y.~Nam, M.~Hong, and D.-S. Park, ``Network
  anomaly detection based on probabilistic analysis,'' {\em Soft Computing},
  vol.~22, no.~20, pp.~6621--6627, 2018.

\bibitem{Atli2018}
B.~G. Atli, Y.~Miche, A.~Kalliola, I.~Oliver, S.~Holtmanns, and A.~Lendasse,
  ``Anomaly-based intrusion detection using extreme learning machine and
  aggregation of network traffic statistics in probability space,'' {\em Cog.
  Comput.}, vol.~10, no.~5, pp.~848--863, 2018.

\bibitem{Cao2016}
V.~L. Cao, M.~Nicolau, and J.~McDermott, ``One-class classification for anomaly
  detection with kernel density estimation and genetic programming,'' in {\em
  European Conference on Genetic Programming} (M.~I. Heywood, J.~McDermott,
  M.~Castelli, E.~Costa, and K.~Sim, eds.), (Cham), pp.~3--18, Springer
  International Publishing, 2016.

\bibitem{Scott2006}
C.~D. Scott and R.~D. Nowak, ``Learning minimum volume sets,'' {\em J. Mach.
  Learn. Res.}, vol.~7, no.~Apr., pp.~665--704, 2006.

\bibitem{Hero2007}
A.~O. Hero, ``Geometric entropy minimization ({GEM}) for anomaly detection and
  localization,'' in {\em Adv. Neural Inf. Process. Sys. ({NeurIPS})},
  pp.~585--592, 2007.

\bibitem{Sricharan2011}
K.~Sricharan and A.~O. Hero, ``Efficient anomaly detection using bipartite
  k-{NN} graphs,'' in {\em Adv. Neural Inf. Process. Sys. ({NeurIPS})}
  (J.~Shawe-Taylor, R.~S. Zemel, P.~L. Bartlett, F.~Pereira, and K.~Q.
  Weinberger, eds.), pp.~478--486, Curran Associates, Inc., 2011.

\bibitem{Hou2018}
E.~Hou, K.~Sricharan, and A.~O. Hero, ``Latent laplacian maximum entropy
  discrimination for detection of high-utility anomalies,'' {\em {IEEE} Trans.
  Inf. Forensics Security}, vol.~13, no.~6, pp.~1446--1459, 2018.

\bibitem{Schoelkopf2000}
B.~Sch{\"o}lkopf, R.~C. Williamson, A.~J. Smola, J.~Shawe-Taylor, and J.~C.
  Platt, ``Support vector method for novelty detection,'' in {\em Adv. Neural
  Inf. Process. Sys. ({NeurIPS})}, pp.~582--588, 2000.

\bibitem{Lim2018}
S.~K. Lim, Y.~Loo, N.-T. Tran, N.-M. Cheung, G.~Roig, and Y.~Elovici, ``Doping:
  Generative data augmentation for unsupervised anomaly detection with gan,''
  in {\em {IEEE} Inter. Conf. Data Min. ({ICDM})}, pp.~1122--1127, IEEE, 2018.

\bibitem{Abdulhammed2019}
R.~Abdulhammed, M.~Faezipour, A.~Abuzneid, and A.~AbuMallouh, ``Deep and
  machine learning approaches for anomaly-based intrusion detection of
  imbalanced network traffic,'' {\em {IEEE} Sensors Lett.}, vol.~3, no.~1,
  pp.~1--4, 2019.

\bibitem{Goodfellow2014}
I.~Goodfellow, J.~Pouget-Abadie, M.~Mirza, B.~Xu, D.~Warde-Farley, S.~Ozair,
  A.~Courville, and Y.~Bengio, ``Generative adversarial nets,'' in {\em Adv.
  Neural Inf. Process. Sys. ({NeurIPS})}, pp.~2672--2680, 2014.

\bibitem{AriasFigueroa2017}
J.~Arias~Figueroa and A.~Ram\'irez~Rivera, ``Learning to cluster with auxiliary
  tasks: A semi-supervised approach,'' in {\em Conf. Graphics Patterns Images
  ({SIBGRAPI})}, pp.~1--8, Oct. 2017.

\bibitem{AriasFigueroa2017a}
J.~Arias~Figueroa and A.~Ram\'irez~Rivera, ``Is simple better?: Revisiting
  simple generative models for unsupervised clustering,'' in {\em Wksp.
  Bayesian Deep Learn. ({NeurIPS})}, Dec. 2017.

\bibitem{Kingma2014}
D.~P. Kingma and M.~Welling, ``Auto-encoding variational bayes,'' in {\em
  Inter. Conf. Learn. Represent. ({ICLR})}, 2014.

\bibitem{Baur2018}
C.~Baur, B.~Wiestler, S.~Albarqouni, and N.~Navab, ``Deep autoencoding models
  for unsupervised anomaly segmentation in brain {MR} images,'' in {\em Inter.
  {MICCAI} Brainlesion Wksp.}, pp.~161--169, Springer, 2018.

\bibitem{Xu2018a}
H.~Xu, W.~Chen, N.~Zhao, Z.~Li, J.~Bu, Z.~Li, Y.~Liu, Y.~Zhao, D.~Pei, Y.~Feng,
  {\em et~al.}, ``Unsupervised anomaly detection via variational auto-encoder
  for seasonal kpis in web applications,'' in {\em {WWW} Conference on World
  Wide Web}, pp.~187--196, International World Wide Web Conferences Steering
  Committee, 2018.

\bibitem{Zhai2016}
S.~Zhai, Y.~Cheng, W.~Lu, and Z.~Zhang, ``Deep structured energy based models
  for anomaly detection,'' {\em arXiv:1605.07717}, 2016.

\bibitem{Nguyen2018}
M.-N. Nguyen and N.~A. Vien, ``Scalable and interpretable one-class {SVMs} with
  deep learning and random fourier features,'' in {\em European Conf. Princ.
  Data Min. Knowl. Discov. ({ECML-PKDD})}, pp.~157--172, Springer, 2018.

\bibitem{Landgrebe2006}
T.~C. Landgrebe, D.~M. Tax, P.~Pacl{\'i}k, and R.~P. Duin, ``The interaction
  between classification and reject performance for distance-based
  reject-option classifiers,'' {\em Pattern Recogn. Lett.}, vol.~27, no.~8,
  pp.~908--917, 2006.

\bibitem{Ahonen2006}
T.~Ahonen, A.~Hadid, and M.~Pietikainen, ``Face description with local binary
  patterns: Application to face recognition,'' {\em {IEEE} Trans. Pattern Anal.
  Mach. Intell.}, no.~12, pp.~2037--2041, 2006.

\bibitem{Lecun1998}
Y.~Lecun, L.~Bottou, Y.~Bengio, and P.~Haffner, ``Gradient-based learning
  applied to document recognition,'' in {\em Proc. {IEEE}}, vol.~86,
  pp.~2278--2324, 1998.

\bibitem{Lecun2010}
Y.~LeCun, C.~Cortes, and C.~Burges, ``{MNIST} handwritten digit database.''
  AT\&T Labs, 2010.

\bibitem{Griffin2007}
G.~Griffin, A.~Holub, and P.~Perona, ``Caltech-256 object category dataset,''
  Tech. Rep. CaltechAUTHORS:CNS-TR-2007-001, Caltech, 2007.

\bibitem{Nene1996}
S.~A. Nene, S.~K. Nayar, and H.~Murase, ``Columbia object image library
  ({COIL100}),'' Tech. Rep. CUCS-006-96,, Columbia University, 1996.

\bibitem{Georghiades2001}
A.~S. Georghiades, P.~N. Belhumeur, and D.~J. Kriegman, ``From few to many:
  Illumination cone models for face recognition under variable lighting and
  pose,'' {\em {IEEE} Trans. Pattern Anal. Mach. Intell.}, vol.~23, no.~6,
  pp.~643--660, 2001.

\bibitem{Kclee05}
K.~Lee, J.~Ho, and D.~Kriegman, ``Acquiring linear subspaces for face
  recognition under variable lighting,'' {\em {IEEE} Trans. Pattern Anal. Mach.
  Intell.}, vol.~27, no.~5, pp.~684--698, 2005.

\bibitem{Chan2008}
A.~Chan and N.~Vasconcelos, ``{UCSD} pedestrian dataset,'' {\em {IEEE} Trans.
  Pattern Anal. Mach. Intell.}, vol.~30, no.~5, pp.~909--926, 2008.

\bibitem{Tschandl2018}
P.~Tschandl, C.~Rosendahl, and H.~Kittler, ``The {HAM10000} dataset, a large
  collection of multi-source dermatoscopic images of common pigmented skin
  lesions,'' {\em Scientific data}, vol.~5, p.~180161, 2018.

\bibitem{Codella2018}
N.~C. Codella, D.~Gutman, M.~E. Celebi, B.~Helba, M.~A. Marchetti, S.~W. Dusza,
  A.~Kalloo, K.~Liopyris, N.~Mishra, H.~Kittler, {\em et~al.}, ``Skin lesion
  analysis toward melanoma detection: A challenge at the 2017 international
  symposium on biomedical imaging ({ISBI}), hosted by the international skin
  imaging collaboration ({ISIC}),'' in {\em {IEEE} Inter. Symp. Biomed. Imag.
  ({ISBI})}, pp.~168--172, IEEE, 2018.

\bibitem{Krizhevsky2009}
A.~Krizhevsky, ``Learning multiple layers of features from tiny images,''
  Master's thesis, University of Toronto, 2009.

\bibitem{Chaker2017}
R.~Chaker, Z.~Al~Aghbari, and I.~N. Junejo, ``Social network model for crowd
  anomaly detection and localization,'' {\em Pattern Recogn.}, vol.~61,
  pp.~266--281, 2017.

\bibitem{Lu2018}
Y.~Lu and P.~Xu, ``Anomaly detection for skin disease images using variational
  autoencoder,'' {\em arXiv:1807.01349}, 2018.

\bibitem{Golan2018}
I.~Golan and R.~El-Yaniv, ``Deep anomaly detection using geometric
  transformations,'' in {\em Adv. Neural Inf. Process. Sys. ({NeurIPS})},
  pp.~9758--9769, 2018.

\bibitem{Hofer2019}
C.~Hofer, R.~Kwitt, M.~Dixit, and M.~Niethammer, ``Connectivity-optimized
  representation learning via persistent homology,'' in {\em Inter. Conf. Mach.
  Learn. ({ICML})}, 2019.

\bibitem{Ruff2018}
L.~Ruff, R.~Vandermeulen, N.~Goernitz, L.~Deecke, S.~A. Siddiqui, A.~Binder,
  E.~M{\"u}ller, and M.~Kloft, ``Deep one-class classification,'' in {\em
  Inter. Conf. Mach. Learn. ({ICML})}, pp.~4393--4402, 2018.

\bibitem{Liu2010}
G.~Liu, Z.~Lin, and Y.~Yu, ``Robust subspace segmentation by low-rank
  representation,'' in {\em Inter. Conf. Mach. Learn. ({ICML})}, pp.~663--670,
  2010.

\bibitem{Tsakiris2015}
M.~C. Tsakiris and R.~Vidal, ``Dual principal component pursuit,'' in {\em
  {IEEE} Inter. Conf. Comput. Vis. Wksps. ({ICCVW})}, pp.~10--18, 2015.

\bibitem{Soltanolkotabi2012}
M.~Soltanolkotabi, E.~J. Candes, {\em et~al.}, ``A geometric analysis of
  subspace clustering with outliers,'' {\em Anals Stats.}, vol.~40, no.~4,
  pp.~2195--2238, 2012.

\bibitem{Adam2008}
A.~Adam, E.~Rivlin, I.~Shimshoni, and D.~Reinitz, ``Robust real-time unusual
  event detection using multiple fixed-location monitors,'' {\em {IEEE} Trans.
  Pattern Anal. Mach. Intell.}, vol.~30, no.~3, pp.~555--560, 2008.

\bibitem{Mehran2009}
R.~Mehran, A.~Oyama, and M.~Shah, ``Abnormal crowd behavior detection using
  social force model,'' in {\em {IEEE}/{CVF} Inter. Conf. Comput. Vis. Pattern
  Recog. ({CVPR})}, pp.~935--942, IEEE, 2009.

\bibitem{Kim2009}
J.~Kim and K.~Grauman, ``Observe locally, infer globally: a space-time mrf for
  detecting abnormal activities with incremental updates,'' in {\em
  {IEEE}/{CVF} Inter. Conf. Comput. Vis. Pattern Recog. ({CVPR})},
  pp.~2921--2928, IEEE, 2009.

\bibitem{Mahadevan2010}
V.~Mahadevan, W.~Li, V.~Bhalodia, and N.~Vasconcelos, ``Anomaly detection in
  crowded scenes,'' in {\em {IEEE}/{CVF} Inter. Conf. Comput. Vis. Pattern
  Recog. ({CVPR})}, pp.~1975--1981, IEEE, 2010.

\bibitem{Sun2018}
J.~Sun, X.~Wang, N.~Xiong, and J.~Shao, ``Learning sparse representation with
  variational auto-encoder for anomaly detection,'' {\em {IEEE} Access},
  vol.~6, pp.~33353--33361, 2018.

\bibitem{Fan2018}
Y.~Fan, G.~Wen, D.~Li, S.~Qiu, and M.~D. Levine, ``Video anomaly detection and
  localization via gaussian mixture fully convolutional variational
  autoencoder,'' {\em arXiv:1805.11223}, 2018.

\bibitem{Vincent2008}
P.~Vincent, H.~Larochelle, Y.~Bengio, and P.-A. Manzagol, ``Extracting and
  composing robust features with denoising autoencoders,'' in {\em Inter. Conf.
  Mach. Learn. ({ICML})}, pp.~1096--1103, ACM, 2008.

\end{thebibliography}

\vspace*{-30pt}
\begin{IEEEbiography}[{\includegraphics[width=1in,height=1.1in,clip,keepaspectratio]{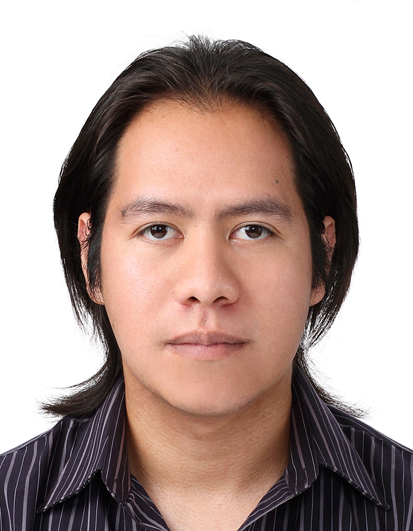}}]{Ad\'{i}n Ram\'{i}rez Rivera} (S'12, M'14) received his B.Eng.\ degree in Computer Engineering from Universidad de San Carlos de Guatemala (USAC), Guatemala in 2009.  He completed his M.Sc.\ and Ph.D.\ degrees in Computer Engineering from Kyung Hee University, South Korea in 2013.  He is currently Assistant Professor at the Institute of Computing, University of Campinas, Brazil.  His research interests are video understanding (including video classification, semantic segmentation, spatiotemporal feature modeling, and generation), and understanding and creating complex feature spaces.
\end{IEEEbiography}

\vspace*{-20pt}

\begin{IEEEbiography}[{\includegraphics[width=1in,height=1.1in,clip,keepaspectratio]{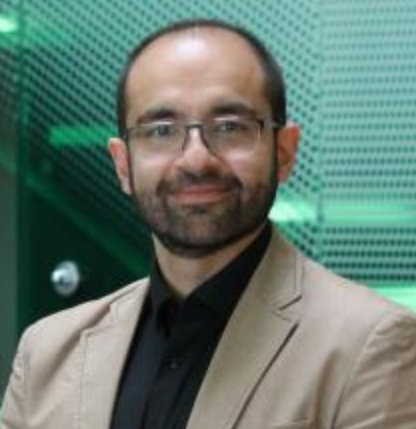}}]{Adil Khan} received his B.S.\ degree in Information Technology from National University of Sciences and Technology (NUST), Pakistan in 2005. He completed his M.Sc.\ and Ph.D.\ degrees in Computer Engineering from Kyung Hee University, South Korea in 2011. He is currently Professor at the Institute of Artificial Intelligence and Data Science, Innopolis University, Russia. His research interests are machine learning and deep learning.
\end{IEEEbiography}

\vspace*{-20pt}

\begin{IEEEbiography}[{\includegraphics[width=1in,height=1.1in,clip,keepaspectratio]{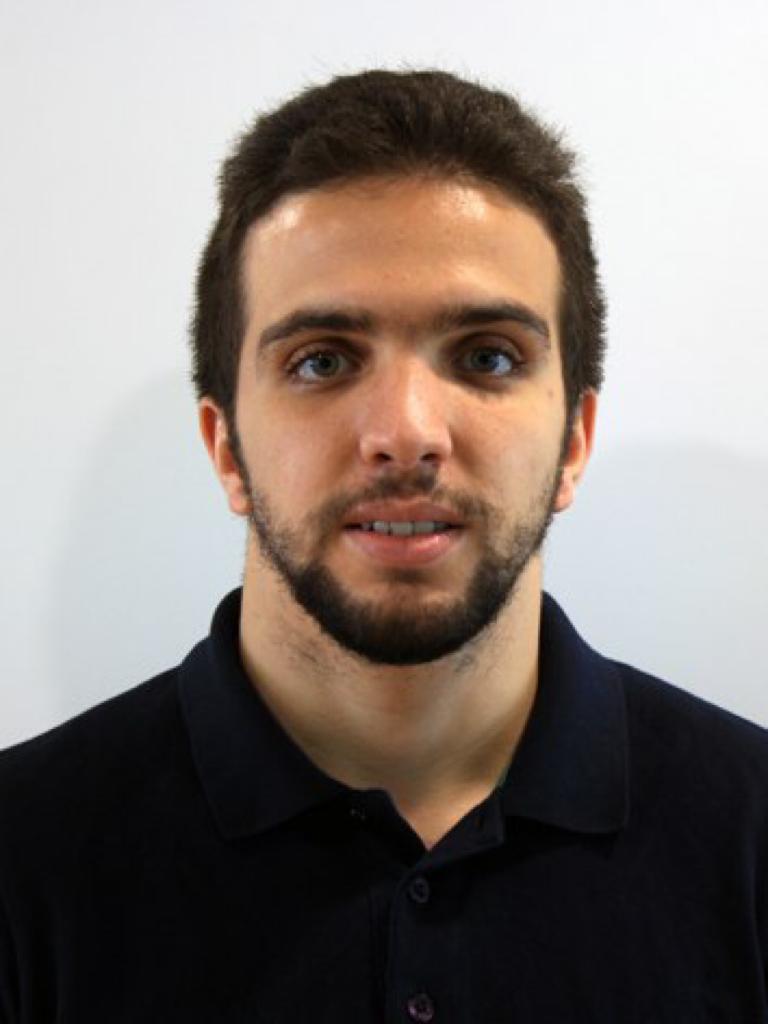}}]{Imad Eddine  Ibrahim Bekkouch} received his B.S.\ degree in Computer science from Abdel Hamid Mehri Constantine 2 University, Algeria in 2018.  He got his M.Sc.\ in data science at Innopolis University, Russia.  Currently, he is pursuing his PhD in Sorbonne Center for Artificial Intelligence, Paris, France, and working as a research assistant at the Institute  of Artificial Intelligence and Data Science, Innopolis  University, Russia.  His research  interests are Domain adaptation, Computer Vision, Machine Learning and Deep Learning.
\end{IEEEbiography}

\vspace*{-20pt}

\begin{IEEEbiography}[{\includegraphics[width=1in,height=1.1in,clip,keepaspectratio]{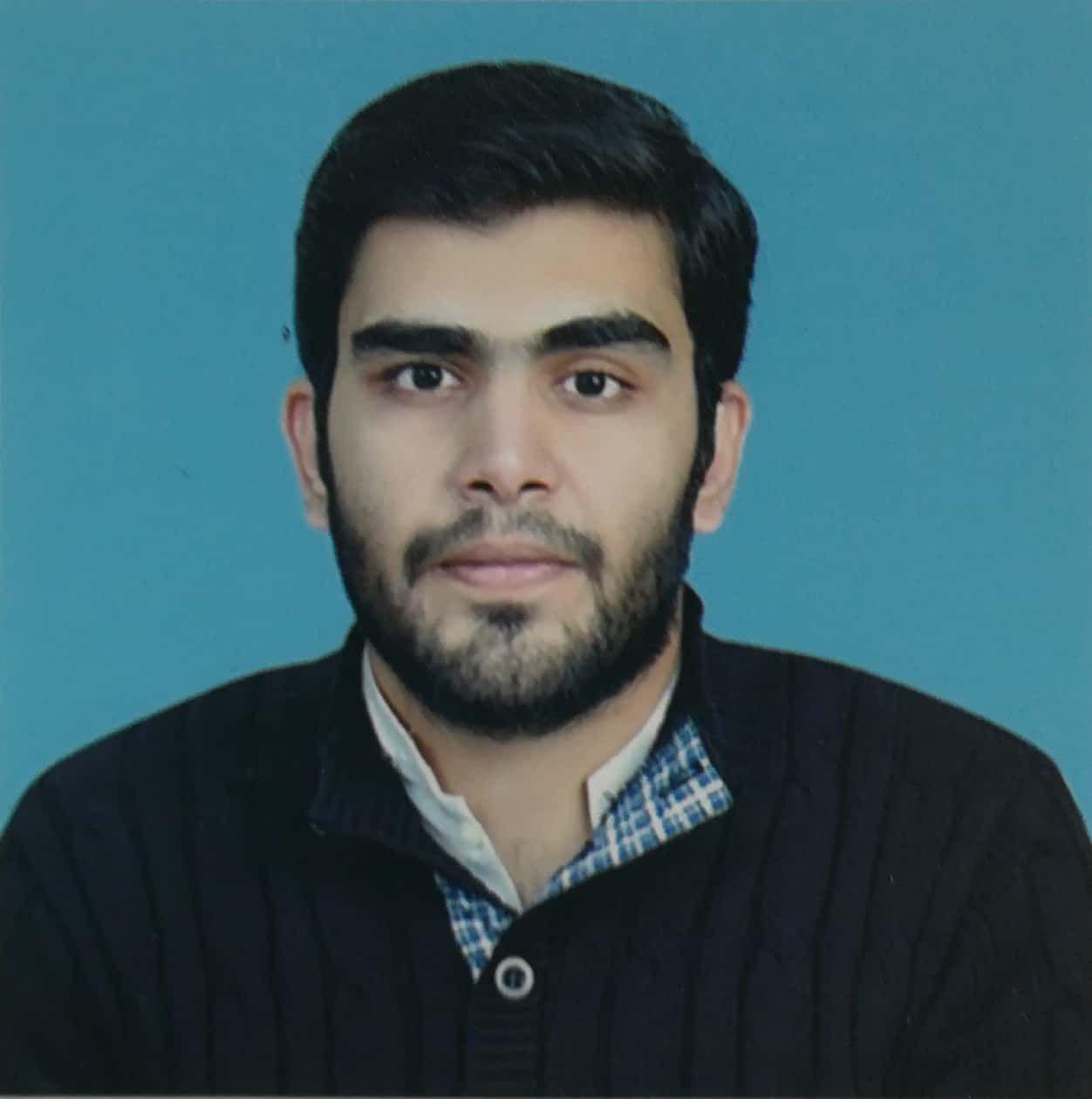}}]{Taimoor Shakeel Sheikh} received the B.Eng. (Telecommunication)\ and M.Eng.\ (Electrical) degrees from National University of Computer and Emerging Sciences (NUCES), Pakistan in 2012 and 2015 respectively. He completed his another M.Sc.\ degree in Informatics and Computer Science from Innopolis University,Russia, in 2019. He is currently an researcher at the Institute of Artificial Intelligence and Data Science, Innopolis University, Russia. His research interests are computer vision related tasks, and deep learning.
\end{IEEEbiography}

\vfill

\end{document}


\pgfplotstableread{
1	1	0.89649	0.88097	0.71985	0.6042	0.87283	0.85411	0.68933	0.54912	0.93438	0.92959	0.7717	0.70408	0.91615	0.90841	0.74139	0.65018	0.83306	0.79613	0.73641	0.6418	0.8035	0.7982	0.71649	0.61029	0.87226	0.85355	0.70303	0.57745
1	1.5	0.79286	0.77101	0.60669	0.34573	0.56111	0.62	0.66005	0.48458	0.70583	0.7798	0.69371	0.5549	0.74975	0.66607	0.58706	0.2907	0.77923	0.70952	0.65583	0.4724	0.38	0.44714	0.71655	0.60416	0.87095	0.85182	0.53738	0.13885
1	2	0.70347	0.57823	0.58138	0.2753	0.6	0.53381	0.65802	0.47934	0.4925	0.54545	0.63743	0.42542	0.71976	0.59384	0.64816	0.45211	0.73603	0.6413	0.66265	0.48122	0.58	0.57	0.68128	0.532	0.84204	0.81129	0.52999	0.11296
1	2.5	0.68449	0.53841	0.5823	0.28052	0.6	0.51238	0.63279	0.41931	0.43333	0.52583	0.63622	0.42744	0.71111	0.5928	0.6313	0.41515	0.72201	0.61475	0.69572	0.55887	0.48667	0.49921	0.67593	0.52055	0.77177	0.68813	0.63927	0.43447
1	3	0.67751	0.52338	0.5823	0.28052	0.48333	0.38571	0.64194	0.44074	0.50833	0.55758	0.59784	0.3266	0.68998	0.5478	0.58405	0.28364	0.67984	0.52828	0.67642	0.51455	0.46667	0.46349	0.6757	0.51983	0.67322	0.49284	0.53625	0.13505
1	3.5	0.66886	0.50466	0.5823	0.28052	0.53333	0.38571	0.61315	0.36791	0.50333	0.49704	0.59726	0.32384	0.63144	0.41501	0.5855	0.28997	0.67198	0.51169	0.67	0.50732	0.48667	0.47778	0.65085	0.462	0.74194	0.62543	0.61701	0.38066
1	4	0.65209	0.46571	0.55208	0.32654	0.53333	0.38571	0.60819	0.35437	0.55667	0.6226	0.58165	0.2773	0.62668	0.40304	0.58169	0.27703	0.6464	0.45187	0.66612	0.49763	0.4075	0.29786	0.63888	0.4343	0.80646	0.7599	0.53192	0.11988
1	4.5	0.64126	0.44004	0.57829	0.2598	0.43889	0.39725	0.6054	0.34746	0.335	0.33667	0.59177	0.30924	0.62091	0.3874	0.58054	0.27128	0.6275	0.40562	0.63423	0.4192	0.44417	0.38437	0.64558	0.4501	0.72454	0.59196	0.58664	0.29645
1	5	0.63834	0.43255	0.58138	0.2753	0.39444	0.26593	0.59917	0.33064	0.4475	0.48231	0.58295	0.28374	0.59286	0.31194	0.58126	0.2749	0.61921	0.38457	0.6458	0.44943	0.50333	0.43175	0.611	0.36317	0.73994	0.62267	0.53058	0.11518
1.25	1	0.90556	0.89476	0.72702	0.62436	0.73214	0.63045	0.70334	0.57794	0.88612	0.87145	0.72292	0.61664	0.89086	0.87732	0.71065	0.59249	0.81574	0.76733	0.71492	0.60109	0.96589	0.96446	0.71659	0.60425	0.85473	0.82974	0.6051	0.34685
1.25	1.5	0.62667	0.59286	0.59223	0.30856	0.71027	0.59103	0.6294	0.40974	0.93218	0.92702	0.61411	0.37147	0.65	0.70617	0.70144	0.5736	0.70786	0.7181	0.70663	0.57209	0.81545	0.77104	0.6633	0.49133	0.78657	0.72533	0.53321	0.12453
1.25	2	0.58556	0.58773	0.58297	0.28388	0.70456	0.57995	0.63314	0.41976	0.92418	0.91795	0.59774	0.32494	0.65773	0.7166	0.63862	0.43376	0.70786	0.7181	0.70079	0.57133	0.69258	0.55011	0.67111	0.50989	0.73881	0.63905	0.70787	0.58713
1.25	2.5	0.61444	0.47922	0.58255	0.28179	0.65321	0.46801	0.62084	0.38899	0.89383	0.88094	0.59068	0.30628	0.81545	0.77104	0.61985	0.38586	0.73619	0.76537	0.67117	0.50953	0.61578	0.48249	0.66506	0.49614	0.75844	0.67157	0.57411	0.25663
1.25	3	0.4754	0.42253	0.58297	0.28388	0.64708	0.45414	0.61163	0.36465	0.7155	0.67569	0.59072	0.30653	0.62424	0.65954	0.61435	0.37169	0.62786	0.6381	0.6599	0.48442	0.64818	0.45612	0.64236	0.44264	0.72323	0.59636	0.54748	0.17156
1.25	3.5	0.50766	0.43387	0.58292	0.28367	0.64535	0.4477	0.60613	0.35014	0.72901	0.62727	0.59054	0.30567	0.59494	0.54896	0.60091	0.33425	0.81964	0.77385	0.6346	0.42358	0.55441	0.4405	0.64252	0.44328	0.61	0.35937	0.63784	0.41377
1.25	4	0.50022	0.41463	0.58255	0.28179	0.64708	0.45414	0.60601	0.34957	0.69725	0.56312	0.58239	0.28068	0.66249	0.65388	0.58879	0.30127	0.65429	0.64189	0.62331	0.3953	0.53062	0.38775	0.61876	0.38339	0.60223	0.3381	0.54838	0.17644
1.25	4.5	0.59414	0.43534	0.58175	0.27744	0.64654	0.45222	0.60488	0.34656	0.65165	0.46324	0.58305	0.2843	0.76742	0.69639	0.58281	0.28305	0.62286	0.49212	0.62662	0.40383	0.54378	0.41876	0.60082	0.33505	0.63507	0.40277	0.53317	0.12435
1.25	5	0.58635	0.41395	0.58242	0.28123	0.5821	0.40484	0.60023	0.33355	0.61996	0.38492	0.58297	0.28389	0.54896	0.42191	0.58218	0.27988	0.50798	0.36687	0.6175	0.37899	0.51894	0.36225	0.60685	0.35168	0.81897	0.77731	0.57807	0.26207
1.5	1	0.94646	0.94337	0.67685	0.52104	0.77299	0.70377	0.70771	0.58693	0.84129	0.81125	0.71901	0.60916	0.88492	0.86992	0.71726	0.60576	0.84453	0.81052	0.70788	0.58988	0.84	0.8264	0.71002	0.59509	0.82863	0.79295	0.52985	0.11258
1.5	1.5	0.365	0.31984	0.58874	0.30107	0.72996	0.62975	0.62382	0.39671	0.89579	0.88366	0.62317	0.39453	0.8462	0.81785	0.60988	0.3596	0.80962	0.76451	0.61575	0.39651	0.78818	0.72941	0.66202	0.48931	0.75201	0.66632	0.55545	0.19963
1.5	2	0.39333	0.28095	0.58329	0.28548	0.72031	0.61142	0.62292	0.39457	0.86422	0.84281	0.61543	0.37451	0.794	0.73987	0.61139	0.36411	0.77967	0.71662	0.62191	0.39169	0.72	0.60952	0.65374	0.4696	0.68089	0.53113	0.53569	0.13323
1.5	2.5	0.36833	0.31667	0.58312	0.2846	0.67935	0.52592	0.61626	0.37576	0.7703	0.70013	0.61554	0.37498	0.77045	0.69806	0.60505	0.34651	0.70121	0.6525	0.61751	0.38047	0.5836	0.50696	0.64181	0.44177	0.67003	0.49929	0.55619	0.202
1.5	3	0.22833	0.22381	0.58315	0.2848	0.66136	0.48729	0.60411	0.34353	0.75639	0.67475	0.61463	0.37285	0.73614	0.64108	0.5915	0.30814	0.72692	0.62187	0.61635	0.3774	0.6003	0.44972	0.63048	0.41355	0.61346	0.36984	0.52977	0.11239
1.5	3.5	0.32833	0.23333	0.58308	0.28445	0.68058	0.52988	0.60319	0.34136	0.70774	0.58666	0.60475	0.346	0.74901	0.66343	0.58322	0.28511	0.6728	0.51049	0.61276	0.36758	0.64788	0.45604	0.62036	0.38773	0.73164	0.60194	0.64601	0.44358
1.5	4	0.42667	0.25	0.58296	0.28381	0.63144	0.41501	0.60336	0.34209	0.64449	0.54463	0.60369	0.34318	0.73528	0.63887	0.58306	0.28433	0.62692	0.40126	0.61182	0.36544	0.45668	0.32674	0.61683	0.37847	0.76437	0.67068	0.53425	0.12776
1.5	4.5	0.44833	0.3	0.58325	0.2853	0.63144	0.41501	0.59805	0.32688	0.60877	0.35627	0.60435	0.34427	0.68479	0.539	0.58307	0.28438	0.62665	0.40103	0.60939	0.3586	0.62788	0.40476	0.611	0.3625	0.76658	0.67022	0.60072	0.33799
1.5	5	0.42667	0.25	0.58262	0.2821	0.63144	0.41501	0.59795	0.32639	0.60796	0.35274	0.60034	0.3336	0.71374	0.59652	0.58271	0.28258	0.62051	0.38574	0.59349	0.31454	0.59061	0.42511	0.60529	0.34709	0.7828	0.72233	0.60233	0.33884
1.75	1	0.84799	0.82046	0.68866	0.55094	0.72728	0.63311	0.69759	0.56643	0.76569	0.69393	0.7138	0.5989	0.81292	0.77554	0.70403	0.57917	0.8159	0.77363	0.7098	0.59107	0.78824	0.72991	0.70665	0.58458	0.78616	0.7279	0.60011	0.33336
1.75	1.5	0.66573	0.49683	0.58902	0.30133	0.65079	0.4599	0.61255	0.39675	0.77169	0.67608	0.6203	0.38738	0.77923	0.70952	0.60863	0.3563	0.74975	0.66607	0.63597	0.4273	0.72988	0.61551	0.62706	0.40511	0.73404	0.63566	0.56688	0.23349
1.75	2	0.66573	0.49683	0.58322	0.28515	0.68231	0.52865	0.61516	0.37436	0.75044	0.64587	0.6184	0.38282	0.73603	0.6413	0.61786	0.38106	0.71976	0.59384	0.63546	0.42551	0.69924	0.58846	0.6187	0.38286	0.66937	0.50578	0.70303	0.57745
1.75	2.5	0.59806	0.44233	0.58295	0.28376	0.67888	0.52385	0.61132	0.36395	0.78127	0.71497	0.61305	0.36815	0.72201	0.61475	0.61042	0.36109	0.71111	0.5928	0.63385	0.4219	0.68593	0.52866	0.61484	0.37322	0.71944	0.59468	0.53738	0.13885
1.75	3	0.54949	0.31641	0.58327	0.28538	0.65934	0.48206	0.61008	0.3601	0.74009	0.64086	0.61219	0.36606	0.67984	0.52828	0.59378	0.31541	0.68998	0.5478	0.61965	0.38601	0.6632	0.48719	0.62067	0.38836	0.61776	0.38118	0.52999	0.11296
1.75	3.5	0.581	0.27381	0.58312	0.28462	0.64579	0.44976	0.60756	0.35385	0.68224	0.51837	0.60707	0.35198	0.67198	0.51169	0.58298	0.28393	0.63144	0.41501	0.61584	0.3761	0.65	0.46129	0.61879	0.38371	0.61556	0.37541	0.63927	0.43447
1.75	4	0.58333	0.28571	0.58289	0.28346	0.63252	0.41781	0.60573	0.34895	0.62409	0.38561	0.60337	0.34257	0.6464	0.45187	0.5832	0.28501	0.62668	0.40304	0.61668	0.37813	0.64329	0.44208	0.60959	0.35924	0.58923	0.3028	0.53625	0.13505
1.75	4.5	0.581	0.27381	0.58306	0.28433	0.62608	0.40147	0.60375	0.34334	0.60112	0.33587	0.59596	0.32108	0.6275	0.40562	0.58312	0.28461	0.62091	0.3874	0.61292	0.36824	0.64099	0.43845	0.60836	0.35557	0.78709	0.72946	0.61701	0.38066
1.75	5	0.49328	0.28855	0.58279	0.28288	0.61849	0.3805	0.602	0.33876	0.59924	0.32711	0.59425	0.31698	0.61921	0.38457	0.58286	0.28336	0.59286	0.31194	0.61192	0.36555	0.62268	0.38506	0.60493	0.34516	0.78063	0.71869	0.53192	0.11988
2	1	0.81015	0.76507	0.69905	0.56862	0.73325	0.63404	0.69716	0.56548	0.76528	0.69321	0.71293	0.59722	0.79082	0.73294	0.70514	0.58502	0.77973	0.71728	0.70728	0.58591	0.77422	0.70568	0.69915	0.57314	0.77035	0.70188	0.58664	0.29645
2	1.5	0.63887	0.43301	0.58901	0.30216	0.7014	0.57364	0.59709	0.3249	0.77015	0.67352	0.60835	0.35584	0.76463	0.69129	0.59477	0.31824	0.61148	0.3603	0.6159	0.37615	0.74879	0.65256	0.61763	0.37963	0.71772	0.60664	0.53058	0.11518
2	2	0.61808	0.38113	0.58301	0.28402	0.66141	0.48698	0.60841	0.35608	0.76047	0.68491	0.60752	0.35369	0.74225	0.6525	0.58288	0.28335	0.70384	0.5593	0.61391	0.36993	0.74737	0.65752	0.60665	0.35095	0.66527	0.49659	0.6051	0.34685
2	2.5	0.59927	0.32728	0.58311	0.28457	0.64215	0.44237	0.60911	0.35769	0.71872	0.60841	0.60804	0.35408	0.69103	0.55234	0.60021	0.33187	0.70585	0.57937	0.59823	0.32532	0.71644	0.60298	0.59894	0.32911	0.63609	0.42772	0.53321	0.12453
2	3	0.58688	0.2937	0.58328	0.28543	0.62891	0.40903	0.60513	0.34606	0.70551	0.58226	0.60722	0.35254	0.67847	0.52567	0.59707	0.33843	0.64669	0.43955	0.58567	0.29173	0.69631	0.56373	0.60413	0.34448	0.64655	0.44122	0.70787	0.58713
2	3.5	0.5823	0.28068	0.58297	0.28386	0.62921	0.41048	0.60387	0.34345	0.67434	0.5169	0.60543	0.34799	0.64164	0.44022	0.58792	0.29874	0.66882	0.50122	0.58292	0.28362	0.64167	0.44002	0.60746	0.35283	0.63566	0.40436	0.57411	0.25663
2	4	0.58222	0.27997	0.58323	0.28522	0.62673	0.40371	0.60105	0.33529	0.60639	0.35064	0.60256	0.34015	0.63074	0.41205	0.5832	0.28505	0.61932	0.38397	0.58242	0.2811	0.63379	0.42205	0.6041	0.3444	0.66943	0.47411	0.54748	0.17156
2	4.5	0.58289	0.28348	0.58323	0.28522	0.62708	0.40514	0.60051	0.33441	0.59562	0.31962	0.60063	0.33479	0.61503	0.37282	0.58308	0.28441	0.6165	0.37765	0.58298	0.28399	0.62644	0.40142	0.6041	0.34443	0.74163	0.63263	0.63784	0.41377
2	5	0.58127	0.27521	0.58265	0.2822	0.61582	0.37548	0.60053	0.33449	0.59584	0.32064	0.59747	0.32545	0.61156	0.36251	0.58325	0.28529	0.6062	0.34906	0.58281	0.28305	0.59928	0.32743	0.60398	0.34399	0.77821	0.71492	0.54838	0.17644
}\mydata

\begin{tikzpicture}[
  font=\footnotesize,
]

\newcommand{\plot}[3][AUC]{%
  \begin{axis}[
  alpha beta 3d=#1,
  width=4.5cm,
  height=3.88125cm, %
  ]
  \addplot3[surf] table [x index=0, y index=1, z index=#2] {#3};
  \end{axis}}

\matrix[%
matrix of nodes, 
nodes in empty cells,
column sep=5pt,
] {%
  & AKIEC & BCC & BKL  \\
  $\text{OG}_e$ & 
  \node{};\plot[$F_1$]{5}{\mydata} & 
  \node{};\plot[$F_1$]{9}{\mydata} & \node{};\plot[$F_1$]{13}{\mydata} \\

  $\text{OG}_\ell$ & 
  \node{};\plot[$F_1$]{3}{\mydata} & 
  \node{};\plot[$F_1$]{7}{\mydata} & \node{};\plot[$F_1$]{11}{\mydata}\\

  & DF & MEL & VASC \\
  $\text{OG}_e$ &
  \node{};\plot[$F_1$]{17}{\mydata} &
  \node{};\plot[$F_1$]{21}{\mydata} & \node{};\plot[$F_1$]{25}{\mydata}\\

  $\text{OG}_\ell$ &
  \node{};\plot[$F_1$]{15}{\mydata} &
  \node{};\plot[$F_1$]{19}{\mydata} & \node{};\plot[$F_1$]{23}{\mydata}\\

  & All Disease & & \\
  $\text{OG}_e$ &  \node{};\plot[$F_1$]{29}{\mydata}  & \\ 
  $\text{OG}_\ell$ & \node{};\plot[$F_1$]{27}{\mydata} & \\ 
};
\end{tikzpicture}


\pgfplotstableread{
1	1	0.87712	0.74161	0.8874	0.72339	0.9008	0.54053	0.89186	0.56923
1	1.5	0.89993	0.59567	0.88746	0.68244	0.95556	0.86406	0.87581	0.69521
1	2	0.90639	0.66289	0.88742	0.64451	0.94062	0.88437	0.89315	0.76718
1	2.5	0.905	0.70651	0.88749	0.61793	0.94528	0.94074	0.88307	0.76923
1	3	0.91066	0.73363	0.88073	0.48949	0.94108	0.93793	0.89606	0.8014
1	3.5	0.91248	0.77249	0.8805	0.43601	0.94529	0.95321	0.8967	0.81378
1	4	0.9111	0.7868	0.88001	0.48488	0.93226	0.94701	0.89903	0.82871
1	4.5	0.90908	0.80024	0.8801	0.44871	0.94395	0.95937	0.88976	0.81684
1	5	0.90756	0.81267	0.87425	0.48319	0.95186	0.94379	0.90288	0.84345
1.25	1	0.89161	0.72931	0.86643	0.76565	0.92063	0.57421	0.92724	0.64906
1.25	1.5	0.90287	0.62126	0.82638	0.69262	0.93288	0.89621	0.87798	0.6806
1.25	2	0.88216	0.69144	0.85643	0.64302	0.9555	0.88023	0.87892	0.71373
1.25	2.5	0.8903	0.71588	0.80633	0.59883	0.94329	0.93469	0.88161	0.75504
1.25	3	0.88137	0.74933	0.7963	0.56319	0.93342	0.95217	0.88942	0.77626
1.25	3.5	0.88247	0.77832	0.87668	0.78659	0.9482	0.949	0.8972	0.82127
1.25	4	0.90163	0.80312	0.82659	0.7244	0.94931	0.95656	0.8933	0.80795
1.25	4.5	0.88188	0.80425	0.84666	0.68069	0.95711	0.94068	0.88938	0.80808
1.25	5	0.8966	0.81665	0.80668	0.64573	0.93483	0.93026	0.89026	0.82356
1.5	1	0.88491	0.76282	0.90777	0.72179	0.93891	0.66	0.93697	0.68289
1.5	1.5	0.8806	0.51255	0.8988	0.53421	0.95991	0.84964	0.90371	0.74823
1.5	2	0.87942	0.55849	0.90036	0.58255	0.94635	0.9022	0.8788	0.72259
1.5	2.5	0.88011	0.60477	0.90874	0.63964	0.93402	0.9389	0.8924	0.77011
1.5	3	0.88332	0.65636	0.90705	0.69835	0.94891	0.94966	0.88463	0.76783
1.5	3.5	0.88931	0.7117	0.90777	0.72179	0.94898	0.95059	0.88502	0.78841
1.5	4	0.89439	0.74509	0.90421	0.73041	0.93574	0.94874	0.8986	0.81813
1.5	4.5	0.88781	0.76863	0.90617	0.7604	0.93263	0.94776	0.89108	0.81393
1.5	5	0.89136	0.7873	0.9076	0.78269	0.93564	0.93264	0.88344	0.79667
1.75	1	0.87512	0.76131	0.90339	0.56674	0.95232	0.71285	0.95365	0.73667
1.75	1.5	0.88398	0.54081	0.90339	0.56674	0.94259	0.87659	0.88603	0.71325
1.75	2	0.89154	0.6002	0.90472	0.60622	0.93581	0.92612	0.88297	0.74383
1.75	2.5	0.89106	0.6477	0.91081	0.66762	0.9344	0.9482	0.88443	0.75832
1.75	3	0.88964	0.68427	0.91013	0.69649	0.93816	0.94264	0.89216	0.7919
1.75	3.5	0.8963	0.72352	0.91017	0.71695	0.93496	0.94272	0.88977	0.80558
1.75	4	0.88666	0.74909	0.91062	0.74327	0.93233	0.94789	0.90065	0.82166
1.75	4.5	0.89356	0.77368	0.91278	0.76714	0.93617	0.93128	0.88079	0.7921
1.75	5	0.89919	0.80655	0.91409	0.798	0.93764	0.93386	0.8919	0.81802
2	1	0.88914	0.73821	0.90705	0.69835	0.9353	0.8295	0.96339	0.77912
2	1.5	0.89895	0.60867	0.90982	0.60336	0.93475	0.9097	0.88922	0.73602
2	2	0.90448	0.67349	0.91462	0.65247	0.93537	0.94701	0.88839	0.76256
2	2.5	0.90645	0.71753	0.91247	0.68914	0.93879	0.94092	0.88765	0.77573
2	3	0.90816	0.74955	0.91452	0.71543	0.93761	0.94787	0.8897	0.78229
2	3.5	0.90748	0.77212	0.91672	0.74765	0.94247	0.93716	0.8967	0.81375
2	4	0.90724	0.80878	0.91643	0.76565	0.94045	0.93703	0.8994	0.817
2	4.5	0.90328	0.81353	0.91668	0.78659	0.93836	0.93595	0.9008	0.83579
2	5	0.90377	0.82174	0.91379	0.79711	0.94033	0.93821	0.89291	0.82193
}\mydata

\begin{tikzpicture}[
  font=\footnotesize,
]

\newcommand{\plot}[3][AUC]{%
  \begin{axis}[
  alpha beta 3d=#1,
  ]
  \addplot3[surf] table [x index=0, y index=1, z index=#2] {#3};
  \end{axis}}

\matrix[%
matrix of nodes, 
nodes in empty cells,
column sep=5pt,
] {%
  &  $\text{OG}_e$ &  $\text{OG}_\ell$ \\
  Ped1 & 
  \node{};\plot[$F_1$]{5}{\mydata} & 
  \node{};\plot[$F_1$]{3}{\mydata} \\
  
  Ped2 & 
  \node{};\plot[$F_1$]{9}{\mydata} & 
  \node{};\plot[$F_1$]{7}{\mydata} \\
};
\end{tikzpicture}